# Completeness Guarantees for Incomplete Ontology Reasoners: Theory and Practice

**Bernardo Cuenca Grau**                    BERNARDO.CUENCA.GRAU@CS.OX.AC.UK
**Boris Motik**                                     BORIS.MOTIK@CS.OX.AC.UK
**Giorgos Stoilos**                           GIORGOS.STOILOS@CS.OX.AC.UK
**Ian Horrocks**                                  IAN.HORROCKS@CS.OX.AC.UK
*Department of Computer Science, University of Oxford*
*Wolfson Building, Parks Road, OX1 3QD, Oxford*

## Abstract

To achieve scalability of query answering, the developers of Semantic Web applications are often forced to use *incomplete* OWL 2 reasoners, which fail to derive all answers for at least one query, ontology, and data set. The lack of completeness guarantees, however, may be unacceptable for applications in areas such as health care and defence, where missing answers can adversely affect the application's functionality. Furthermore, even if an application can tolerate some level of incompleteness, it is often advantageous to estimate how many and what kind of answers are being lost.

In this paper, we present a novel logic-based framework that allows one to check whether a reasoner is complete for a given query $\mathcal{Q}$ and ontology $\mathcal{T}$—that is, whether the reasoner is guaranteed to compute all answers to $\mathcal{Q}$ w.r.t. $\mathcal{T}$ and an arbitrary data set $\mathcal{A}$. Since ontologies and typical queries are often fixed at application design time, our approach allows application developers to check whether a reasoner known to be incomplete in general is actually complete for the kinds of input relevant for the application.

We also present a technique that, given a query $\mathcal{Q}$, an ontology $\mathcal{T}$, and reasoners $R_1$ and $R_2$ that satisfy certain assumptions, can be used to determine whether, for each data set $\mathcal{A}$, reasoner $R_1$ computes more answers to $\mathcal{Q}$ w.r.t. $\mathcal{T}$ and $\mathcal{A}$ than reasoner $R_2$. This allows application developers to select the reasoner that provides the highest degree of completeness for $\mathcal{Q}$ and $\mathcal{T}$ that is compatible with the application's scalability requirements.

Our results thus provide a theoretical and practical foundation for the design of future ontology-based information systems that maximise scalability while minimising or even eliminating incompleteness of query answers.

## 1. Introduction

Efficient management and querying of large amounts of data is a core problem for a growing range of applications in fields as diverse as biology (Sidhu, Dillon, Chang, & Sidhu, 2005), medicine (Golbreich, Zhang, & Bodenreider, 2006), geography (Goodwin, 2005), astronomy (Derriere, Richard, & Preite-Martinez, 2006), agriculture (Soergel, Lauser, Liang, Fisseha, Keizer, & Katz, 2004), and defence (Lacy, Aviles, Fraser, Gerber, Mulvehill, & Gaskill, 2005). In order to facilitate interoperability, such applications often use standard data models and query languages. In particular, RDF (Hayes, 2004) provides a standard model for semistructured data, SPARQL (Prud'hommeaux & Seaborne, 2008) is a standard query language for RDF, and ontology languages such as OWL (Horrocks, Patel-Schneider, & van Harmelen, 2003) and OWL 2 (Cuenca Grau, Horrocks, Motik, Parsia, Patel-Schneider,

& Sattler, 2008b) can be used to describe background knowledge about the application domain. Thus, answering SPARQL queries over RDF data sets structured using an OWL ontology is a key service in ontology-based information systems.

An important question in the design of such systems is the selection of an appropriate reasoner. Systems such as Pellet (Sirin, Parsia, Cuenca Grau, Kalyanpur, & Katz, 2007), HermiT (Motik, Shearer, & Horrocks, 2009b), and RACER (Haarslev & Möller, 2001) are based on (hyper)tableau algorithms that are provably *complete*—that is, they are guaranteed to compute *all* answers for *each* query, ontology, and data set. Completeness, however, comes at the cost of scalability, as answering queries over OWL 2 ontologies is of high computational complexity (Glimm, Horrocks, Lutz, & Sattler, 2007; Ortiz, Calvanese, & Eiter, 2008; Calvanese, De Giacomo, Lembo, Lenzerini, & Rosati, 2007; Lutz, Toman, & Wolter, 2009). Thus, complete systems often fail to meet the scalability demands of applications that manage data sets consisting of hundreds of millions or even billions of assertions.

Scalability of query answering can be ensured by restricting the expressive power of the ontology language to the level that makes provably complete reasoning tractable. This has led to the development of three *profiles* of OWL 2 (Motik, Cuenca Grau, Horrocks, Wu, Fokoue, & Lutz, 2009a): OWL 2 EL, OWL 2 RL, and OWL 2 QL. Query answering in all three profiles can be implemented in polynomial time w.r.t. the size of data (and even in logarithmic space in the case of OWL 2 QL). Such appealing theoretical properties have spurred the development of specialised reasoners such as QuONTO (Acciarri, Calvanese, De Giacomo, Lembo, Lenzerini, Palmieri, & Rosati, 2005) that target specific profiles and typically reject ontologies that fall outside the target profile.

A different solution to the scalability problem is adopted in reasoners such as Oracle's Semantic Data Store (Wu, Eadon, Das, Chong, Kolovski, Annamalai, & Srinivasan, 2008), Sesame (Broekstra, Kampman, & van Harmelen, 2002), Jena (McBride, Brian, 2001), OWLim (Kiryakov, Ognyanov, & Manov, 2005), Minerva (Ma, Yang, Qiu, Xie, Pan, & Liu, 2006), DLE-Jena (Meditskos & Bassiliades, 2008), and Virtuoso (Erling & Mikhailov, 2009). These reasoners accept all OWL 2 ontologies as input—that is, they never reject inputs. Furthermore, to the best of our knowledge, all of these systems are intended to be *sound*, which means that all results of a query are indeed correct answers. Finally, these reasoners typically use scalable reasoning techniques, such as various (deductive) database algorithms. As a consequence, the reasoners are *incomplete*: for each reasoner, at least one query, ontology, and data set exist for which the reasoner does not return all answers to the query. Some of these reasoners were actually designed to be complete for a particular profile of OWL 2 (typically this is OWL 2 RL due to its close connection with datalog), and they can often additionally handle certain kinds of axiom that fall outside the target profile.

Since incomplete reasoners can handle large data sets, they often provide the best practical choice for developers of ontology-based applications. For example, OWLim was used for reasoning in the backend of the BBC's 2010 World Cup website, and Oracle's reasoner is being used by the University of Texas Health Science Center to improve large-scale public health surveillance. In order to verify that the selected reasoner meets the application's requirements, developers typically resort to empirical testing, in which they check the reasoner's answers w.r.t. the application ontology and queries for representative data sets. Although primarily intended for testing performance, benchmark suites such as the Lehigh

University Benchmark (LUBM) (Guo, Pan, & Heflin, 2005) and the University Ontology Benchmark (UOBM) (Ma et al., 2006) have been used for such completeness testing.

Empirical completeness testing, however, has several important limitations. First, tests are not *generic*, as data sets used for testing typically have a fixed and/or repetitive structure, which can skew test results. Second, test data is not *exhaustive*, as completeness is tested only w.r.t. a limited number of data sets. Finally, query answers may not be *verifiable*: since complete reasoners fail to handle large data sets, they often cannot compute the control answers needed to check the answers produced by an incomplete reasoner. As a consequence, empirical completeness tests provide only limited assurance of a reasoner's ability to meet the requirements of a given application.

In this paper, we present a radically different approach to solving these problems. We observed that, given a query $\mathcal{Q}$ and ontology $\mathcal{T}$, even if a reasoner is not complete for the language of $\mathcal{T}$, the reasoner may be able to correctly answer $\mathcal{Q}$ w.r.t. $\mathcal{T}$ and an arbitrary data set $\mathcal{A}$; in such a case, we say that the reasoner is $(\mathcal{Q}, \mathcal{T})$-*complete*. Given that ontology-based applications often use a limited set of queries and a fixed ontology (or at least the queries and the ontology evolve relatively slowly), a scalable reasoner that is generally incomplete, but is $(\mathcal{Q}, \mathcal{T})$-complete for all relevant combinations of $\mathcal{Q}$ and $\mathcal{T}$, may provide a solid foundation for ontology-based applications, allowing them to enjoy 'the best of both worlds': regardless of the data set encountered, such applications will enjoy completeness guarantees normally available only with computationally-intensive complete reasoners, while at the same time exhibiting scalability levels normally available only by sacrificing completeness. To develop an approach for testing the $(\mathcal{Q}, \mathcal{T})$-completeness of a given reasoner, we proceed as follows.

In Section 3 we develop a logic-based framework that allows us to establish formally provable $(\mathcal{Q}, \mathcal{T})$-completeness guarantees. The following two notions are central to our framework. First, in order to abstract away from the implementation details of concrete reasoners, we introduce the notion of an *abstract reasoner*—an idealised reasoner that captures the intended behaviour and salient features (such as soundness and monotonicity) of a class of concrete reasoners. Second, we introduce the notion of a *test suite*—a finite set of data sets and queries. Intuitively, given $\mathcal{Q}$ and $\mathcal{T}$, our goal is to construct a test suite such that, if a reasoner correctly answers all queries on all data sets in the test suite, then the reasoner is guaranteed to be $(\mathcal{Q}, \mathcal{T})$-complete.

Unfortunately, as we show in Section 3.4, for certain $\mathcal{Q}$ and $\mathcal{T}$, it is impossible to construct a finite test suite that will provide the aforementioned completeness guarantees. Therefore, we investigate assumptions on $\mathcal{Q}$, $\mathcal{T}$, and the reasoner under which testing $(\mathcal{Q}, \mathcal{T})$-completeness becomes practically feasible.

In Section 3.5 we consider the case where $\mathcal{Q}$ and $\mathcal{T}$ can be rewritten into a union of conjunctive queries $\mathcal{R}$—that is, when answering $\mathcal{Q}$ w.r.t. $\mathcal{T}$ and a data set $\mathcal{A}$ is equivalent to evaluating $\mathcal{R}$ over $\mathcal{A}$. For $\mathcal{T}$ expressed in OWL 2 QL, a rewriting $\mathcal{R}$ can be computed using the algorithm by Calvanese et al. (2007); additionally, the algorithm by Pérez-Urbina, Motik, and Horrocks (2010) can sometimes compute $\mathcal{R}$ even if $\mathcal{T}$ is syntactically outside this fragment. We show that such $\mathcal{R}$ can be converted into a test suite $\mathbf{E}^{\mathcal{R}}$ that can be used for testing the $(\mathcal{Q}, \mathcal{T})$-completeness of any reasoner that satisfies some basic assumptions; roughly speaking, the reasoner's answers should not depend on the names of the individuals occurring in a data set, and its answers must increase monotonically when new data is added. The size of each test in $\mathbf{E}^{\mathcal{R}}$ is polynomial in the size of the longest conjunctive query in $\mathcal{R}$, so it should be feasible to compute correct answers to the tests using a complete reasoner. The number of tests in $\mathbf{E}^{\mathcal{R}}$, however, can be exponential in the size of $\mathcal{R}$, which may lead to problems in practice. As a remedy, in Section 3.6 we strengthen our assumptions and require the reasoner not to drop answers when 'merging' individuals—that is, if the reasoner returns $\vec{a}$ given inputs $Q$, $\mathcal{T}$, and $\mathcal{A}$, then for each (possibly noninjective) mapping $\mu$ the reasoner returns $\mu(\vec{a})$ given inputs $Q$, $\mathcal{T}$, and $\mu(\mathcal{A})$—and we show that $(\mathcal{Q}, \mathcal{T})$-completeness of such reasoners can be checked using a test suite $\mathbf{I}^{\mathcal{R}}$ obtained from $\mathcal{R}$ by a linear transformation.

That $\mathcal{Q}$ and $\mathcal{T}$ should be rewritable into a union of conjunctive queries effectively prevents $\mathcal{T}$ from stating recursive axioms. To overcome this restriction, in Section 3.7 we consider *first-order reproducible* reasoners—that is, reasoners whose behaviour on $\mathcal{Q}$, $\mathcal{T}$, and $\mathcal{A}$ can be seen as computing certain answers of $\mathcal{Q}$ w.r.t. some (possibly unknown) first-order theory $\mathcal{F}_{\mathcal{T}}$ and $\mathcal{A}$. Since $\mathcal{F}_{\mathcal{T}}$ can be a datalog program, most reasoners based on deductive databases are first-order reproducible. In addition, we require $\mathcal{Q}$ and $\mathcal{T}$ to be rewritable into datalog$^{\pm,\vee}$—an extension of datalog that allows for existential quantifiers and disjunction in rule heads. In many cases, $\mathcal{T}$ can be transformed into a datalog$^{\pm,\vee}$ program using equivalence-preserving transformations; furthermore, the algorithm by Pérez-Urbina et al. (2010) can in many cases produce a plain datalog rewriting. We then show how to transform a datalog$^{\pm,\vee}$ rewriting of $\mathcal{Q}$ and $\mathcal{T}$ into a test suite that can be used to test $(\mathcal{Q}, \mathcal{T})$-completeness of first-order reproducible reasoners.

In Section 4 we turn our attention to comparing incomplete reasoners. Roughly speaking, given $\mathcal{Q}$ and $\mathcal{T}$, reasoner $R_1$ is 'more complete' than reasoner $R_2$ if, for each data set $\mathcal{A}$, reasoner $R_1$ computes all the answers to $\mathcal{Q}$ w.r.t. $\mathcal{T}$ and $\mathcal{A}$ that are computed by $R_2$. We show that comparing incomplete reasoners is infeasible in general. Therefore, we introduce the notion of *compact* reasoners—that is, reasoners whose behaviour on $\mathcal{Q}$, $\mathcal{T}$, and $\mathcal{A}$ can be seen as first selecting some subset $\mathcal{T}'$ of $\mathcal{T}$ and then using a complete reasoner to evaluate $\mathcal{Q}$ w.r.t. $\mathcal{T}'$ and $\mathcal{A}$. Thus, the class of compact reasoners captures all reasoners that reduce the input ontology $\mathcal{T}$ to a set of axioms that match certain parameters, such as fitting into language fragments. For $\mathcal{Q}$ and $\mathcal{T}$ that can be rewritten into a union of conjunctive queries $\mathcal{R}$, we show that the test suite $\mathbf{I}^{\mathcal{R}}$ can be used to compare compact reasoners.

We have implemented our approaches for computing test suites, and have tested completeness of several well-known reasoners (see Section 5). These show that test suites can be efficiently computed for realistic ontologies. Furthermore, we were able to guarantee $(\mathcal{Q}, \mathcal{T})$-completeness of the evaluated reasoners for many queries and ontologies. Finally, when no $(\mathcal{Q}, \mathcal{T})$-completeness guarantee could be provided, we were able to compute a counter-example—a small data set for which the reasoner at hand was incomplete.

## 2. Preliminaries

In this section we briefly introduce Description Logics (DLs) (Baader, McGuinness, Nardi, & Patel-Schneider, 2002)—a family of knowledge representation formalisms which underpin the OWL and OWL 2 ontology languages. We describe description logics in the wider framework of first-order logic since many of our results hold for arbitrary first-order theories.

We then introduce the datalog$^{\pm,\vee}$ and datalog languages, and we define the syntax and semantics of unions of conjunctive queries (UCQs). Finally, we introduce the notions of UCQ, datalog, and datalog$^{\pm,\vee}$ rewritings, which underpin many of our techniques.

## 2.1 Description Logics and First-Order Logic

Most of the results in this paper hold for arbitrary first-order theories, rather than description logics. Our work, however, is motivated by description logics and ontologies, so we use the DL terminology throughout the paper; for example, we often talk about TBoxes and ABoxes instead of first-order theories and sets of facts.

All definitions in this paper are implicitly parameterised by a *signature* $\Sigma = \langle \Sigma_P, \Sigma_I \rangle$, which consists of countably infinite disjoint sets of *predicates* $\Sigma_P$ and *individuals* (commonly called *constants* in first-order logic) $\Sigma_I$. Each predicate is associated with a nonnegative *arity*; predicates of zero arity are commonly called *propositional symbols*. The notions of *variables*, *terms*, *atoms*, *first-order formulae*, and *sentences* are defined as usual (Fitting, 1996); we do not consider function symbols in this article and we assume all formulae to be function-free. The atom that is false (true) in all interpretations is written $\bot$ ($\top$). An atom is a *fact* if it does not contain variables. We use the standard first-order notions of *satisfiability*, *unsatisfiability*, and *entailment* (written $\models$) of sets of first-order sentences.

We assume that $\Sigma_P$ contains the special *equality* and *inequality* predicates $\approx$ and $\not\approx$, respectively; atoms of the form $\approx(t_1, t_2)$ and $\not\approx(t_1, t_2)$ are commonly written as $t_1 \approx t_2$ and $t_1 \not\approx t_2$, respectively. We make a technical assumption that $\approx$ and $\not\approx$ are distinct predicates rather than, as it is common in first-order logic, that $t_1 \not\approx t_2$ is an abbreviation for $\neg(t_1 \approx t_2)$; furthermore, we assume that each theory that uses $\approx$ and $\not\approx$ axiomatises their semantics as follows, where (5) is instantiated for each predicate $P$ of arity $n$ and each $i$ with $1 \leq i \leq n$.

$$\forall x, y.[x \not\approx y \wedge x \approx y \to \bot] \quad (1)$$

$$\forall x.[x \approx x] \quad (2)$$

$$\forall x, y.[x \approx y \to y \approx x] \quad (3)$$

$$\forall x, y, z.[x \approx y \wedge y \approx z \to x \approx z] \quad (4)$$

$$\forall x_1, \ldots, x_i, \ldots, x_n, y_i.[P(x_1, \ldots, x_i, \ldots, x_n) \wedge x_i \approx y_i \to P(x_1, \ldots, y_i, \ldots, x_n)] \quad (5)$$

Note that, according to this assumption, each set of facts is satisfiable. For example, the set of atoms $\{a \approx b, a \not\approx b\}$ is satisfiable since both $a \approx b$ and $a \not\approx b$ are positive variable-free atoms that are semantically independent from each other; moreover, axiom (1) is required to obtain the expected contradiction.

An *individual renaming* (often just *renaming*) is a partial function $\mu : \Sigma_I \to \Sigma_I$ that maps individuals to individuals. The domain and the range of $\mu$ are written $\mathsf{dom}(\mu)$ and $\mathsf{rng}(\mu)$; unless otherwise noted, we assume that $\mathsf{dom}(\mu)$ is finite. For $\alpha$ an object containing individuals (such as a formula, a set of formulae, or a tuple of individuals), $\mathsf{ind}(\alpha)$ is the set of individuals occurring in $\alpha$, and $\mu(\alpha)$ is obtained from $\alpha$ by simultaneously replacing each individual $a \in \mathsf{ind}(\alpha) \cap \mathsf{dom}(\mu)$ with $\mu(a)$.

We use the notion of substitutions from first-order logic; that is, a *substitution* $\sigma$ is a mapping of variables to terms. For $\alpha$ a term, an atom, or a formula, the result of applying a substitution $\sigma$ to $\alpha$ is written as $\sigma(\alpha)$.

A *TBox* $\mathcal{T}$ a is a finite set of first-order sentences that contains axioms (1)–(5) whenever $\approx$ and/or $\not\approx$ are used. An *ABox* $\mathcal{A}$ is a finite set of facts. Note that this definition allows for atoms of the form $a \approx b$ and $a \not\approx b$ in ABoxes; furthermore, since ABoxes can contain only positive atoms, each ABox (when considered without a TBox) is satisfiable.

| DL Name | Roles | Concepts | TBox Axioms |
|---------|-------|----------|-------------|
| $\mathcal{EL}$ | $R$ | $\top$, $A$, $C_1 \sqcap C_2$, $\exists R.C$ | $C_1 \sqsubseteq C_2$ |
| $\mathcal{FL}$ | $R$ | $\top$, $A$, $C_1 \sqcap C_2$, $\forall R.C$ | $C_1 \sqsubseteq C_2$ |
| $\mathcal{ALC}$ | $R$ | $\top$, $\bot$, $A$, $\neg C$, $C_1 \sqcap C_2$, $C_1 \sqcup C_2$, $\exists R.C$, $\forall R.C$ | $C_1 \sqsubseteq C_2$ |
| $+(\mathcal{H})$ | | | $R_1 \sqsubseteq R_2$ |
| $+(\mathcal{R})$ | | $\exists R.\mathsf{Self}$ | $R \circ S \sqsubseteq T$ |
| $+(\mathcal{S})$ | | | $\mathsf{Trans}(R)$ |
| $+(\mathcal{I})$ | $R^-$ | | |
| $+(\mathcal{Q})$ | | $\geq nS.C$, $\leq nS.C$ | |
| $+(\mathcal{O})$ | | $\{a\}$ | |

Table 1: Syntax of standard description logics. Typical extensions of $\mathcal{EL}$, $\mathcal{ALC}$, and $\mathcal{FL}$ are named by appending calligraphic letters ($\mathcal{H}$, $\mathcal{R}$, $\mathcal{S}$, $\mathcal{I}$, $\mathcal{Q}$, and/or $\mathcal{O}$).

A *description logic* $\mathcal{DL}$ is a (usually infinite) recursive set of TBoxes satisfying the following conditions:

- for each $\mathcal{T} \in \mathcal{DL}$ and each renaming $\mu$, we have $\mu(\mathcal{T}) \in \mathcal{DL}$, and

- for each $\mathcal{T} \in \mathcal{DL}$ and each $\mathcal{T}' \subseteq \mathcal{T}$, we have $\mathcal{T}' \in \mathcal{DL}$.

If $\mathcal{T} \in \mathcal{DL}$, we say that $\mathcal{T}$ is a $\mathcal{DL}$-TBox. Finally, $\mathcal{FOL}$ is the largest description logic that contains all finite sets of first-order sentences over the signature in question.

We next present an overview of the DLs commonly considered in the literature. Typically, the predicates in DL signatures are required to be unary or binary; the former are commonly called *atomic concepts* and the latter are commonly called *atomic roles*. DLs typically use a specialised syntax, summarised in Table 1, that provides a set of constructors for constructing complex concepts and roles from simpler ones, as well as different kinds of axioms. Using the translation from Table 2, concepts can be translated into first-order formulae with one free variable, roles can be translated into first-order formulae with two free variables, and axioms can be translated into first-order sentences. Note that the translation uses counting quantifiers $\exists^{\geq n}$ and $\exists^{\leq n}$, which can be expressed by using ordinary quantifiers and equality by well-known transformations.

In the rest of this paper, we commonly write TBoxes and ABoxes in DL syntax; however, to simplify the presentation, we identify $\mathcal{T}$ and $\mathcal{A}$ written in DL syntax with $\pi(\mathcal{T})$ and $\pi(\mathcal{A})$.

## 2.2 Datalog$^{\pm,\vee}$

We next introduce a fragment of first-order logic called datalog$^{\pm,\vee}$ as an extension of datalog$^{\pm}$ by Calì, Gottlob, Lukasiewicz, Marnette, and Pieris (2010). A datalog$^{\pm,\vee}$ *rule* (or commonly just a *rule*) $r$ is a formula of the form (6), where each $B_j$ is an atom different from $\bot$ whose free variables are contained in $\vec{x}$, and

- $m = 1$ and $\varphi_1(\vec{x}, \vec{y}_1) = \bot$, or

- $m \geq 1$ and, for each $1 \leq i \leq m$, formula $\varphi_i(\vec{x}, \vec{y}_i)$ is a conjunction of atoms different from $\bot$ whose free variables are contained in $\vec{x} \cup \vec{y}_i$.

| Mapping DL roles into first-order logic | |
|---|---|
| $\pi(R, x, y)$ | $= R(x, y)$ |
| $\pi(R^-, x, y)$ | $= R(y, x)$ |

| Mapping DL concepts into first-order logic | |
|---|---|
| $\pi(\top, x, y)$ | $= \top$ |
| $\pi(\bot, x, y)$ | $= \bot$ |
| $\pi(A, x, y)$ | $= A(x)$ |
| $\pi(\{a\}, x, y)$ | $= x \approx a$ |
| $\pi(\neg C, x, y)$ | $= \neg\pi(C, x, y)$ |
| $\pi(C \sqcap D, x, y)$ | $= \pi(C, x, y) \wedge \pi(D, x, y)$ |
| $\pi(C \sqcup D, x, y)$ | $= \pi(C, x, y) \vee \pi(D, x, y)$ |
| $\pi(\exists R.C, x, y)$ | $= \exists y.[\pi(R, x, y) \wedge \pi(C, y, x)]$ |
| $\pi(\exists R.\mathsf{Self}, x, y)$ | $= R(x, x)$ |
| $\pi(\forall R.C, x, y)$ | $= \forall y.[\pi(R, x, y) \rightarrow \pi(C, y, x)]$ |
| $\pi(\geq nS.C, x, y)$ | $= \exists^{\geq n} y.[\pi(S, x, y) \wedge \pi(C, y, x)]$ |
| $\pi(\leq nS.C, x, y)$ | $= \exists^{\leq n} y.[\pi(S, x, y) \wedge \pi(C, y, x)]$ |

| Mapping TBox axioms into first-order logic | |
|---|---|
| $\pi(C \sqsubseteq D)$ | $= \forall x.[\pi(C, x, y) \rightarrow \pi(D, x, y)]$ |
| $\pi(R \sqsubseteq S)$ | $= \forall x, y.[\pi(R, x, y) \rightarrow \pi(S, x, y)]$ |
| $\pi(\mathsf{Trans}(R))$ | $= \forall x, y, z.[\pi(R, x, y) \wedge \pi(R, y, z) \rightarrow \pi(R, x, z)]$ |
| $\pi(R \circ S \sqsubseteq T)$ | $= \forall x, y, z.[\pi(R, x, y) \wedge \pi(S, y, z) \rightarrow \pi(T, x, z)]$ |

| Mapping ABox axioms into first-order logic | |
|---|---|
| $\pi(C(a))$ | $= \pi(C, a, y)$ |
| $\pi(R(a, b))$ | $= R(a, b)$ |
| $\pi(a \approx b)$ | $= a \approx b$ |
| $\pi(a \not\approx b)$ | $= a \not\approx b$ |

Table 2: Translation of DL syntax into first-order logic

$$\forall \vec{x}.[B_1 \wedge \ldots \wedge B_n \rightarrow \bigvee_{i=1}^{m} \exists \vec{y}_i . \varphi_i(\vec{x}, \vec{y}_i)] \tag{6}$$

A rule is *safe* if each variable in $\vec{x}$ also occurs in some $B_j$; unless otherwise noted, all rules are assumed to be safe. For brevity, the outer quantifier $\forall \vec{x}$ is commonly left implicit. The *body* of $r$ is the set of atoms $\mathsf{body}(r) = \{B_1, \ldots, B_n\}$, and the head of $r$ is the formula $\mathsf{head}(r) = \bigvee_{i=1}^{m} \exists \vec{y}_i . \varphi_i(\vec{x}, \vec{y}_i)$. A datalog$^{\pm, \vee}$ program is a finite set of safe datalog$^{\pm, \vee}$ rules. Note that, since $\approx$ and $\not\approx$ are treated as ordinary predicates, they can occur in rules, provided that their semantics is appropriately axiomatised; furthermore, note that the latter can be achieved using datalog$^{\pm, \vee}$ rules.

Let $r$ be a datalog$^{\pm, \vee}$ rule. Then, $r$ is a datalog$^{\vee}$ rule if $\mathsf{head}(r)$ contains no existential quantifier. Also, $r$ is a datalog$^{\pm}$ rule if $m = 1$. Finally, $r$ is a datalog rule if $m = 1$ and the head of $r$ is a single atom without existential quantifiers (Ceri, Gottlob, & Tanca, 1989).

In several places in this paper, we check whether a set of first-order sentences entails a datalog$^{\pm, \vee}$ rule, which can be accomplished using the following simple result.

**Proposition 2.1.** *Let $\mathcal{F}$ be a set of first-order sentences, and let $r$ be a datalog$^{\pm,\vee}$ rule of the form (6). Then, for each substitution $\sigma$ mapping the free variables of $r$ to distinct individuals not occurring in $\mathcal{F}$ or $r$, we have $\mathcal{F} \models r$ if and only if*

$$\mathcal{F} \cup \{\sigma(B_1), \ldots, \sigma(B_n)\} \models \bigvee_{i=1}^{m} \exists \vec{y}_i.\varphi_i(\sigma(\vec{x}), \vec{y}_i)$$

*Proof.* Let $\vec{x}$ be the tuple of free variables in $r$ and let $\sigma$ be an arbitrary substitution mapping the variables in $\vec{x}$ to distinct individuals not occurring in $\mathcal{F}$ or $r$. The claim of this proposition follows from the following equivalences:

$$\mathcal{F} \models \forall \vec{x}.[B_1 \wedge \ldots \wedge B_n \rightarrow \bigvee_{i=1}^{m} \exists \vec{y}_i.\varphi_i(\vec{x}, \vec{y}_i)] \qquad\qquad \text{iff}$$

$$\mathcal{F} \cup \{\neg[\forall \vec{x}.B_1 \wedge \ldots \wedge B_n \rightarrow \bigvee_{i=1}^{m} \exists \vec{y}_i.\varphi_i(\vec{x}, \vec{y}_i)]\} \text{ is unsatisfiable} \qquad\qquad \text{iff}$$

$$\mathcal{F} \cup \{\exists \vec{x}.[B_1 \wedge \ldots \wedge B_n \wedge \neg \bigvee_{i=1}^{m} \exists \vec{y}_i.\varphi_i(\vec{x}, \vec{y}_i)]\} \text{ is unsatisfiable} \qquad \text{iff (skolem. of } \exists \vec{x})$$

$$\mathcal{F} \cup \{\sigma(B_1) \wedge \ldots \wedge \sigma(B_n) \wedge \neg \bigvee_{i=1}^{m} \exists \vec{y}_i.\varphi_i(\sigma(\vec{x}), \vec{y}_i)\} \text{ is unsatisfiable} \qquad\qquad \text{iff}$$

$$\mathcal{F} \cup \{\sigma(B_1), \ldots, \sigma(B_n), \neg \bigvee_{i=1}^{m} \exists \vec{y}_i.\varphi_i(\sigma(\vec{x}), \vec{y}_i)\} \text{ is unsatisfiable} \qquad\qquad \text{iff}$$

$$\mathcal{F} \cup \{\sigma(B_1), \ldots, \sigma(B_n)\} \models \bigvee_{i=1}^{m} \exists \vec{y}_i.\varphi_i(\sigma(\vec{x}), \vec{y}_i). \qquad\qquad\qquad \square$$

## 2.3 Queries

In order to achieve a high degree of generality, we define a *query* $\mathcal{Q}$ as a finite set of first-order sentences containing a distinct *query predicate* $Q$. Intuitively, the query predicate $Q$ determines the answers of $\mathcal{Q}$. In order to simplify the notation, we typically assume that the association between $\mathcal{Q}$ and the query predicate is implicit (e.g., we may require each query to contain precisely one such predicate), and we assume that no query predicate occurs in a TBox or an ABox.

A tuple of constants $\vec{a}$ is a *certain answer* to a query $\mathcal{Q}$ with query predicate $Q$ with respect to a TBox $\mathcal{T}$ and an ABox $\mathcal{A}$ if the arity of $\vec{a}$ agrees with the arity of $Q$ and $\mathcal{T} \cup \mathcal{A} \cup \mathcal{Q} \models Q(\vec{a})$. The set of all certain answers of $\mathcal{Q}$ w.r.t. $\mathcal{T}$ and $\mathcal{A}$ is denoted as $\mathsf{cert}(\mathcal{Q}, \mathcal{T}, \mathcal{A})$. If the query predicate of $\mathcal{Q}$ is propositional (i.e., if the query is *Boolean*), then $\mathsf{cert}(\mathcal{Q}, \mathcal{T}, \mathcal{A})$ is either empty or it contains the tuple of zero length; in such cases, we commonly write $\mathsf{cert}(\mathcal{Q}, \mathcal{T}, \mathcal{A}) = \mathsf{f}$ and $\mathsf{cert}(\mathcal{Q}, \mathcal{T}, \mathcal{A}) = \mathsf{t}$, respectively.

We use $*$ as the special Boolean query that checks a first-order theory for unsatisfiability. Thus, $\mathsf{cert}(*, \mathcal{T}, \mathcal{A}) = \mathsf{t}$ if and only if $\mathcal{T} \cup \mathcal{A}$ is unsatisfiable.

A query $\mathcal{Q}$ with a query predicate $Q$ is a *union of conjunctive queries* (UCQ) if it is a datalog program in which each rule contains $Q$ in the head but not in the body. A UCQ $\mathcal{Q}$ is a *conjunctive query* (CQ) if it contains exactly one rule.

A union of conjunctive queries $\mathcal{Q}$ is *ground* if, for each rule $r \in \mathcal{Q}$, each variable occurring in the body of $r$ also occurs in the head of $r$. Roughly speaking, when computing $\mathsf{cert}(\mathcal{Q}, \mathcal{T}, \mathcal{A})$ for a ground $\mathcal{Q}$, all variables in $\mathcal{Q}$ can be matched only to the individuals in $\mathcal{T}$ and $\mathcal{A}$, but not to unnamed objects whose existence is guaranteed by existential quantifiers. Many state of the art reasoners used in practice support only ground UCQs. Note that $\mathcal{Q} = \{A(x) \rightarrow Q(x), R(x, y) \rightarrow Q(x, y)\}$ is not a ground UCQ; in fact, $\mathcal{Q}$ is not even a valid first-order theory since predicate $Q$ does not have unique arity. To obtain a UCQ, one can 'pad' the head of the first rule—that is, one can introduce a special fresh individual *null* and rewrite the rules as $\mathcal{Q}' = \{A(x) \rightarrow Q(x, null), R(x, y) \rightarrow Q(x, y)\}$.

By the properties of first-order logic entailment, $\mathsf{cert}$ satisfies the following properties for each query $\mathcal{Q}$, all TBoxes $\mathcal{T}$ and $\mathcal{T}'$, and all ABoxes $\mathcal{A}$ and $\mathcal{A}'$.

1. *Monotonicity*: $\mathcal{T} \subseteq \mathcal{T}'$ and $\mathcal{A} \subseteq \mathcal{A}'$ imply

   - $\mathsf{cert}(*, \mathcal{T}, \mathcal{A}) = \mathsf{t}$ implies $\mathsf{cert}(*, \mathcal{T}', \mathcal{A}') = \mathsf{t}$, and
   - $\mathsf{cert}(\mathcal{Q}, \mathcal{T}, \mathcal{A}) \subseteq \mathsf{cert}(\mathcal{Q}, \mathcal{T}', \mathcal{A}')$.

2. *Invariance under renamings*: For each renaming $\mu$ and each tuple of individuals $\vec{a}$,

   - $\mathsf{cert}(*, \mathcal{T}, \mathcal{A}) = \mathsf{t}$ implies $\mathsf{cert}(*, \mu(\mathcal{T}), \mu(\mathcal{A})) = \mathsf{t}$, and
   - $\vec{a} \in \mathsf{cert}(\mathcal{Q}, \mathcal{T}, \mathcal{A})$ implies $\mu(\vec{a}) \in \mathsf{cert}(\mu(\mathcal{Q}), \mu(\mathcal{T}), \mu(\mathcal{A}))$.

## 2.4 Rewritings

Intuitively, a *rewriting* of a query $\mathcal{Q}$ w.r.t. a TBox $\mathcal{T}$ is another query that captures all the information from $\mathcal{T}$ that is relevant for answering $\mathcal{Q}$ over an arbitrary ABox $\mathcal{A}$ (Calvanese et al., 2007; Artale, Calvanese, Kontchakov, & Zakharyaschev, 2009; Pérez-Urbina et al., 2010). In practice, UCQs (Calvanese et al., 2007) and datalog (Pérez-Urbina et al., 2010) are the most widely used target languages for query rewriting. For the sake of generality, however, in this paper we use a notion of a datalog$^{\pm,\vee}$ rewriting.

**Definition 2.2.** *Let $\mathcal{Q}$ be a query and let $\mathcal{T}$ be a TBox. A datalog$^{\pm,\vee}$ rewriting (or simply a rewriting) of $\mathcal{Q}$ w.r.t. $\mathcal{T}$ is a triple $\mathcal{R} = \langle \mathcal{R}_D, \mathcal{R}_\perp, \mathcal{R}_Q \rangle$ where*

- *$\mathcal{R}_D$ is a datalog$^{\pm,\vee}$ program not containing $\perp$ or $Q$ such that $\mathcal{T} \models \mathcal{R}_D$,*

- *$\mathcal{R}_\perp$ is a datalog program such that $\mathsf{head}(r) = \perp$ for each $r \in \mathcal{R}_\perp$, and*

- *$\mathcal{R}_Q$ is a UCQ whose query predicate is $Q$,*

*such that the following properties hold for each ABox $\mathcal{A}$:*

- *$\mathsf{cert}(*, \mathcal{T}, \mathcal{A}) = \mathsf{cert}(*, \mathcal{R}_D \cup \mathcal{R}_\perp, \mathcal{A})$, and*

- *if $\mathsf{cert}(*, \mathcal{T}, \mathcal{A}) = \mathsf{f}$, then $\mathsf{cert}(\mathcal{Q}, \mathcal{T}, \mathcal{A}) = \mathsf{cert}(\mathcal{R}_Q, \mathcal{R}_D \cup \mathcal{R}_\perp, \mathcal{A})$.*

*Rewriting $\mathcal{R}$ is a* datalog rewriting *if $\mathcal{R}_D$ is a datalog program. Furthermore, rewriting $\mathcal{R}$ is a* UCQ rewriting *if $\mathcal{R}_D = \emptyset$; such an $\mathcal{R}$ is usually written as just $\mathcal{R} = \langle \mathcal{R}_\perp, \mathcal{R}_Q \rangle$.*

Note that Definition 2.2 requires $\mathcal{T} \models \mathcal{R}_D$ to hold, which precludes rewritings consisting of axioms that are unsound w.r.t. $\mathcal{T}$. For example, let $\mathcal{Q} = \{A(x) \rightarrow Q(x)\}$ and $\mathcal{T} = \emptyset$; then, $\mathcal{R}_D = \{B(x) \rightarrow A(x)\}$ does not satisfy the definition of a rewriting since formula $B(x) \rightarrow A(x)$ is not a logical consequence of $\mathcal{T}$.

For a wide range of $\mathcal{T}$ and $\mathcal{Q}$, a datalog$^{\pm,\vee}$ rewriting of $\mathcal{Q}$ w.r.t. $\mathcal{T}$ can be computed using straightforward equivalence-preserving transformations of $\mathcal{T}$; this can be further optimised by eliminating axioms from $\mathcal{T}$ that are irrelevant to answering $\mathcal{Q}$. Furthermore, several algorithms for computing UCQ and datalog rewritings have been proposed in the literature. For example, Calvanese et al. (2007) showed how to compute a UCQ rewriting in cases when $\mathcal{T}$ is expressed in a logic from the DL-Lite family, and this approach can be extended to the OWL 2 QL profile of OWL 2 (Motik et al., 2009a). Similarly, Pérez-Urbina et al. (2010) proposed an algorithm for computing the simplest possible datalog rewriting when $\mathcal{T}$ is expressed in the description logic $\mathcal{ELHIO}$.

Rewritings produced by known algorithms often contain predicates that do not occur in $\mathcal{T}$ and $\mathcal{Q}$; such predicates are sometimes called *fresh*. For example, many rewriting algorithms normalise TBoxes by replacing complex concepts with fresh atomic concepts. A rewriting $\mathcal{R} = \langle \mathcal{R}_D, \mathcal{R}_\perp, \mathcal{R}_Q \rangle$ obtained in such a way is unlikely to satisfy the requirement that $\mathcal{T} \models \mathcal{R}_D$. However, predicates occurring in $\mathcal{R}$ but not in $\mathcal{T}$ can often be eliminated via *unfolding*. For example, let $\mathcal{Q} = \{A(x) \rightarrow Q(x)\}$ and $\mathcal{T} = \{\exists R.\exists S.B \sqsubseteq A\}$, and assume that a rewriting algorithm produces

$$\mathcal{R}_D = \{S(x,y) \wedge B(x) \rightarrow C(x),\ R(x,y) \wedge C(y) \rightarrow A(x)\}.$$

To satisfy Definition 2.2, predicate $C$ can be unfolded and $\mathcal{R}_D$ replaced with

$$\mathcal{R}'_D = \{R(x,y) \wedge S(y,z) \wedge B(z) \rightarrow A(x)\},$$

for which $\mathcal{T} \models \mathcal{R}'_D$ holds. Unfolding, however, may not always be possible (e.g., this might be the case when fresh predicates occur in recursive axioms), which may limit the applicability of some of the results presented in this paper.

## 3. Completeness Guarantees for Incomplete Reasoners

In this section, we introduce the formal framework that will allow us to establish completeness guarantees for incomplete reasoners. Our results are not restricted to any particular description logic, but are applicable to all TBoxes that satisfy the following criterion.

**Definition 3.1.** *A TBox $\mathcal{T}$ is* admissible *if a description logic $\mathcal{DL}$ exists such that $\mathcal{T}$ is a $\mathcal{DL}$-TBox, and both checking TBox satisfiability and answering Boolean UCQs w.r.t. an arbitrary ABox are decidable for $\mathcal{DL}$.*

### 3.1 Concrete and Abstract Reasoners

Concrete reasoners are complex software systems that differ greatly in the functionality and the supported interfaces, and that use a range of different implementation techniques. To make our results general and independent from specific implementation techniques, we introduce the notion of an *abstract reasoner*. An abstract reasoner can be thought of as an idealised reasoner that captures the intended behaviour and salient features of a class of concrete reasoners. A concrete reasoner belonging to this class may use arbitrary algorithms, as long as their observable behaviour mirrors that of the abstract reasoner.

**Definition 3.2.** *An* abstract reasoner ans *for a description logic $\mathcal{DL}$ is a computable function that takes as input an arbitrary $\mathcal{DL}$-TBox $\mathcal{T}$, an arbitrary ABox $\mathcal{A}$, and either the special* unsatisfiability *query $*$ or an arbitrary UCQ $\mathcal{Q}$. The return value of* ans *is defined as follows:*

- *ans$(*, \mathcal{T}, \mathcal{A})$ is either* t *or* f;

- *if* ans$(*, \mathcal{T}, \mathcal{A}) = $ t, *then* ans$(\mathcal{Q}, \mathcal{T}, \mathcal{A})$ *is of no interest and can be arbitrary; and*

- *if* ans$(*, \mathcal{T}, \mathcal{A}) = $ f, *then* ans$(\mathcal{Q}, \mathcal{T}, \mathcal{A})$ *is a finite set of tuples of individuals, where the arity of each tuple is equal to the arity the query predicate of $\mathcal{Q}$.*

*An abstract reasoner* ans *for $\mathcal{DL}$ is said to be* applicable to *a TBox $\mathcal{T}$ if $\mathcal{T}$ is a $\mathcal{DL}$-TBox.*

Intuitively, ans$(*, \mathcal{T}, \mathcal{A})$ asks the abstract reasoner to check whether $\mathcal{T} \cup \mathcal{A}$ is unsatisfiable, and ans$(\mathcal{Q}, \mathcal{T}, \mathcal{A})$ asks the abstract reasoner to evaluate $\mathcal{Q}$ w.r.t. $\mathcal{T} \cup \mathcal{A}$. If $\mathcal{T} \cup \mathcal{A}$ is unsatisfiable, then each tuple of constants of the same arity as the query predicate $Q$ is an answer to $\mathcal{Q}$ on $\mathcal{T} \cup \mathcal{A}$; therefore, the result of ans$(\mathcal{Q}, \mathcal{T}, \mathcal{A})$ is of interest only if ans$(*, \mathcal{T}, \mathcal{A}) = $ f—that is, if ans identifies $\mathcal{T} \cup \mathcal{A}$ as satisfiable.

**Example 3.3.** Consider the abstract reasoners rdf, rdfs, rl, and classify which, given as input a UCQ $\mathcal{Q}$, a TBox $\mathcal{T}$, and an ABox $\mathcal{A}$, compute the answer to $\mathcal{Q}$ w.r.t. $\mathcal{T}$ and $\mathcal{A}$ as described next.

Abstract reasoner rdf ignores $\mathcal{T}$ and evaluates $\mathcal{Q}$ w.r.t. $\mathcal{A}$; more precisely, rdf$(*, \mathcal{T}, \mathcal{A}) = $ f and rdf$(\mathcal{Q}, \mathcal{T}, \mathcal{A}) = $ cert$(\mathcal{Q}, \emptyset, \mathcal{A})$. Thus, rdf captures the behaviour of RDF reasoners.

Abstract reasoner rdfs evaluates $\mathcal{Q}$ w.r.t. $\mathcal{A}$ and a datalog program $\mathcal{P}_{\mathsf{rdfs}}$ that is constructed by translating each RDFS axiom $\alpha$ in $\mathcal{T}$ into an equivalent datalog rule; more precisely, rdfs$(*, \mathcal{T}, \mathcal{A}) = $ f and rdfs$(\mathcal{Q}, \mathcal{T}, \mathcal{A}) = $ cert$(\mathcal{Q}, \mathcal{P}_{\mathsf{rdfs}}, \mathcal{A})$. Thus, rdfs captures the behaviour of RDFS reasoners such as Sesame.

Abstract reasoner rl evaluates $\mathcal{Q}$ w.r.t. $\mathcal{A}$ and a datalog program $\mathcal{P}_{\mathsf{rl}}$ that is constructed by translating each OWL 2 RL axiom $\alpha$ in $\mathcal{T}$ into an equivalent datalog rule; more precisely, rl$(*, \mathcal{T}, \mathcal{A}) = $ cert$(*, \mathcal{P}_{\mathsf{rl}}, \mathcal{A})$ and rl$(\mathcal{Q}, \mathcal{T}, \mathcal{A}) = $ cert$(\mathcal{Q}, \mathcal{P}_{\mathsf{rl}}, \mathcal{A})$. Thus, rl captures the behaviour of OWL 2 RL reasoners such as Jena and Oracle's Semantic Data Store.

Abstract reasoner classify first classifies $\mathcal{T}$ using a complete OWL 2 DL reasoner; that is, it computes a TBox $\mathcal{T}'$ containing each subclass axiom $A \sqsubseteq B$ such that $\mathcal{T} \models A \sqsubseteq B$, and $A$ and $B$ are atomic concepts occurring in $\mathcal{T}$. The abstract reasoner then proceeds as rl, but considers $\mathcal{T} \cup \mathcal{T}'$ instead of $\mathcal{T}$; more precisely, classify$(*, \mathcal{T}, \mathcal{A}) = $ rl$(*, \mathcal{T} \cup \mathcal{T}', \mathcal{A}_{in})$ and classify$(\mathcal{Q}, \mathcal{T}, \mathcal{A}) = $ rl$(\mathcal{Q}, \mathcal{T} \cup \mathcal{T}', \mathcal{A})$. In this way, classify captures the behaviour of OWL 2 RL reasoners such as Minerva and DLE-Jena that try to be 'more complete' by materialising certain consequences of $\mathcal{T}$. ◇

An 'ideal' abstract reasoner is one such that, for an arbitrary UCQ $\mathcal{Q}$, TBox $\mathcal{T}$, and ABox $\mathcal{A}$, we have ans$(*, \mathcal{T}, \mathcal{A}) = $ cert$(*, \mathcal{T}, A)$, and ans$(\mathcal{Q}, \mathcal{T}, \mathcal{A}) = $ cert$(\mathcal{Q}, \mathcal{T}, A)$ whenever ans$(*, \mathcal{T}, \mathcal{A}) = $ f. We next introduce and discuss several properties of abstract reasoners that are likely to affect how close they come to this ideal and that may also be relevant to the applicability of our results.

The following notion of *soundness* describes abstract reasoners that return only answers that logically follow from $\mathcal{Q}$, $\mathcal{T}$, and $\mathcal{A}$.

**Definition 3.4.** *An abstract reasoner* ans *for* $\mathcal{DL}$ *is* sound *if the following conditions hold for each UCQ* $\mathcal{Q}$*,* $\mathcal{DL}$*-TBox* $\mathcal{T}$*, and ABox* $\mathcal{A}$*:*

- ans$(*, \mathcal{T}, \mathcal{A}) = $ t *implies* cert$(*, \mathcal{T}, \mathcal{A}) = $ t*; and*

- ans$(*, \mathcal{T}, \mathcal{A}) = $ f *implies* ans$(\mathcal{Q}, \mathcal{T}, \mathcal{A}) \subseteq $ cert$(\mathcal{Q}, \mathcal{T}, \mathcal{A})$.

The following notion of *monotonicity* describes abstract reasoners for which extending the input TBox and ABox never leads to dropping answers. We also consider a weaker notion of $(\mathcal{Q}, \mathcal{T})$-*monotonicity*, in which the input query $\mathcal{Q}$ and TBox $\mathcal{T}$ are fixed.

**Definition 3.5.** *An abstract reasoner* ans *for* $\mathcal{DL}$ *is* monotonic *if the following conditions hold for each UCQ* $\mathcal{Q}$*, all* $\mathcal{DL}$*-TBoxes* $\mathcal{T}$ *and* $\mathcal{T}'$*, and all ABoxes* $\mathcal{A}$ *and* $\mathcal{A}'$ *such that* $\mathcal{T} \subseteq \mathcal{T}'$ *and* $\mathcal{A} \subseteq \mathcal{A}'$*:*

- ans$(*, \mathcal{T}, \mathcal{A}) = $ t *implies* ans$(*, \mathcal{T}', \mathcal{A}') = $ t*; and*

- ans$(*, \mathcal{T}, \mathcal{A}) = $ f *and* ans$(*, \mathcal{T}', \mathcal{A}') = $ f *imply* ans$(\mathcal{Q}, \mathcal{T}, \mathcal{A}) \subseteq $ ans$(\mathcal{Q}, \mathcal{T}', \mathcal{A}')$.

*Given a UCQ* $\mathcal{Q}$ *and a* $\mathcal{DL}$*-TBox* $\mathcal{T}$*,* ans *is* $(\mathcal{Q}, \mathcal{T})$-monotonic *if the following conditions hold for all ABoxes* $\mathcal{A}$ *and* $\mathcal{A}'$ *such that* $\mathcal{A} \subseteq \mathcal{A}'$*:*

- ans$(*, \mathcal{T}, \mathcal{A}) = $ t *implies* ans$(*, \mathcal{T}, \mathcal{A}') = $ t*; and*

- ans$(*, \mathcal{T}, \mathcal{A}) = $ f *and* ans$(*, \mathcal{T}, \mathcal{A}') = $ f *imply* ans$(\mathcal{Q}, \mathcal{T}, \mathcal{A}) \subseteq $ ans$(\mathcal{Q}, \mathcal{T}, \mathcal{A}')$.

As discussed in Section 2.3, the logical consequences of a first-order theory are invariant under renaming and merging of individuals. To define analogous properties for abstract reasoners, we first introduce the notions of $\mathcal{T}$-stable and $(\mathcal{Q}, \mathcal{T})$-stable renamings—that is, renamings that leave all individuals occurring in $\mathcal{T}$ (respectively, in $\mathcal{Q}$ and $\mathcal{T}$) unchanged.

**Definition 3.6.** *Let* $\mathcal{Q}$ *be a query, let* $\mathcal{T}$ *be a TBox, and let* $\mu$ *be a renaming. Then,* $\mu$ *is* $\mathcal{T}$-stable *if* $\mu(a) = a$ *for each individual* $a \in $ dom$(\mu) \cap $ ind$(\mathcal{T})$*; furthermore,* $\mu$ *is* $(\mathcal{Q}, \mathcal{T})$-stable *if* $\mu(a) = a$ *for each individual* $a \in $ dom$(\mu) \cap $ ind$(\mathcal{Q} \cup \mathcal{T})$.

The following notion of *weak faithfulness* describes abstract reasoners whose answers are invariant under replacement of individuals with fresh individuals. Furthermore, *weak* $(\mathcal{Q}, \mathcal{T})$-*faithfulness* relaxes this property to the case when $\mathcal{Q}$ and $\mathcal{T}$ are fixed.

**Definition 3.7.** *An abstract reasoner* ans *for* $\mathcal{DL}$ *is* weakly faithful *if the following conditions hold for each UCQ* $\mathcal{Q}$*,* $\mathcal{DL}$*-TBox* $\mathcal{T}$*, ABox* $\mathcal{A}$*, injective renaming* $\mu$*, and tuple* $\vec{a}$*:*

- ans$(*, \mathcal{T}, \mathcal{A}) = $ t *and* ind$(\mathcal{T} \cup \mathcal{A}) \subseteq $ dom$(\mu)$ *imply* ans$(*, \mu(\mathcal{T}), \mu(\mathcal{A})) = $ t*; and*

- ans$(*, \mathcal{T}, \mathcal{A}) = $ f, ind$(\mathcal{Q} \cup \mathcal{T} \cup \mathcal{A}) \subseteq $ dom$(\mu)$*, and* $\vec{a} \in $ ans$(\mathcal{Q}, \mathcal{T}, \mathcal{A})$ *imply* ans$(*, \mu(\mathcal{T}), \mu(\mathcal{A})) = $ f *and* $\mu(\vec{a}) \in $ ans$(\mu(\mathcal{Q}), \mu(\mathcal{T}), \mu(\mathcal{A}))$.

*Given a UCQ $\mathcal{Q}$ and a $\mathcal{DL}$-TBox $\mathcal{T}$, $\mathsf{ans}$ is weakly $(\mathcal{Q}, \mathcal{T})$-faithful if the following conditions hold for each ABox $\mathcal{A}$, injective renaming $\mu$, and tuple $\vec{a}$:*

- $\mathsf{ans}(*, \mathcal{T}, \mathcal{A}) = \mathsf{t}$, $\mathsf{ind}(\mathcal{T} \cup \mathcal{A}) \subseteq \mathsf{dom}(\mu)$, *and* $\mu$ *is* $\mathcal{T}$-stable imply $\mathsf{ans}(*, \mathcal{T}, \mu(\mathcal{A})) = \mathsf{t}$; *and*

- $\mathsf{ans}(*, \mathcal{T}, \mathcal{A}) = \mathsf{f}$, $\mathsf{ind}(\mathcal{Q} \cup \mathcal{T} \cup \mathcal{A}) \subseteq \mathsf{dom}(\mu)$, $\mu$ *is* $(\mathcal{Q}, \mathcal{T})$-stable, *and* $\vec{a} \in \mathsf{ans}(\mathcal{Q}, \mathcal{T}, \mathcal{A})$ *imply* $\mathsf{ans}(*, \mathcal{T}, \mu(\mathcal{A})) = \mathsf{f}$ *and* $\mu(\vec{a}) \in \mathsf{ans}(\mathcal{Q}, \mathcal{T}, \mu(\mathcal{A}))$.

The following notion of *strong faithfulness* describes abstract reasoners whose answers are invariant under merging of individuals. Furthermore, *strong $(\mathcal{Q}, \mathcal{T})$-faithfulness* relaxes this property to the case when $\mathcal{Q}$ and $\mathcal{T}$ are fixed.

**Definition 3.8.** *An abstract reasoner $\mathsf{ans}$ for $\mathcal{DL}$ is strongly faithful if the following conditions hold for each UCQ $\mathcal{Q}$, $\mathcal{DL}$-TBox $\mathcal{T}$, ABox $\mathcal{A}$, renaming $\mu$, and tuple $\vec{a}$:*

- $\mathsf{ans}(*, \mathcal{T}, \mathcal{A}) = \mathsf{t}$ *implies* $\mathsf{ans}(*, \mu(\mathcal{T}), \mu(\mathcal{A})) = \mathsf{t}$; *and*

- $\mathsf{ans}(*, \mathcal{T}, \mathcal{A}) = \mathsf{f}$, $\vec{a} \in \mathsf{ans}(\mathcal{Q}, \mathcal{T}, \mathcal{A})$, *and* $\mathsf{ans}(*, \mu(\mathcal{T}), \mu(\mathcal{A})) = \mathsf{f}$ *imply* $\mu(\vec{a}) \in \mathsf{ans}(\mu(\mathcal{Q}), \mu(\mathcal{T}), \mu(\mathcal{A}))$.

*Given a UCQ $\mathcal{Q}$ and a $\mathcal{DL}$-TBox $\mathcal{T}$, $\mathsf{ans}$ is strongly $(\mathcal{Q}, \mathcal{T})$-faithful if the following conditions hold for each ABox $\mathcal{A}$, renaming $\mu$, and tuple $\vec{a}$:*

- $\mathsf{ans}(*, \mathcal{T}, \mathcal{A}) = \mathsf{t}$ *and* $\mu$ *is* $\mathcal{T}$-stable imply $\mathsf{ans}(*, \mathcal{T}, \mu(\mathcal{A})) = \mathsf{t}$; *and*

- $\mathsf{ans}(*, \mathcal{T}, \mathcal{A}) = \mathsf{f}$, $\mu$ *is* $(\mathcal{Q}, \mathcal{T})$-stable, $\vec{a} \in \mathsf{ans}(\mathcal{Q}, \mathcal{T}, \mathcal{A})$, *and* $\mathsf{ans}(*, \mathcal{T}, \mu(\mathcal{A})) = \mathsf{f}$ *imply* $\mu(\vec{a}) \in \mathsf{ans}(\mathcal{Q}, \mathcal{T}, \mu(\mathcal{A}))$.

The results that we present in the rest of this paper are applicable only to abstract reasoners that satisfy various combinations of these properties; as a minimum, we require $(\mathcal{Q}, \mathcal{T})$-monotonicity and weak $(\mathcal{Q}, \mathcal{T})$-faithfulness. The abstract reasoners described in Example 3.3 all satisfy these properties. Testing if this is the case for concrete reasoners may, however, be infeasible in practice; indeed, we are not aware of a technique that would allow one to check whether a concrete reasoner satisfies the required properties. We believe, however, that all concrete reasoners commonly used in practice are intended to be sound, monotonic, and at least weakly faithful, and that strong faithfulness is a reasonable assumption in most cases. If a concrete reasoner fails to satisfy some of these properties on certain inputs, this is likely to be due to implementation bugs; thus, any consequent failure of completeness can be seen as a bug, and detecting such situations should be viewed as a part of a more general problem of testing software systems.

We next present several examples of abstract reasoners that do not satisfy some of the mentioned properties.

**Example 3.9.** Consider an abstract reasoner that behaves as $\mathsf{rdf}$ whenever the number of assertions in the input ABox is smaller than a certain threshold, and that returns the empty set of answers for larger ABoxes. Intuitively, such an abstract reasoner characterises a concrete RDF reasoner that processes inputs only up to a certain size. Such a reasoner is not $(\mathcal{Q}, \mathcal{T})$-monotonic for an arbitrary $\mathcal{Q}$ and $\mathcal{T}$.                    ◇

**Example 3.10.** Consider an abstract reasoner that behaves like rdf, but that, for trust reasons, removes from each input ABox all assertions whose individuals are 'blacklisted' (e.g., they come from an untrusted source). Such an abstract reasoner is not weakly $(\mathcal{Q}, \mathcal{T})$-faithful for an arbitrary $\mathcal{Q}$ and $\mathcal{T}$. ◇

Example 3.10 suggests that, for an abstract reasoner to be weakly faithful, it should not make decisions that depend on specific names of individuals.

**Example 3.11.** Consider an abstract reasoner rl$^{\neq}$ that, given as input a UCQ $\mathcal{Q}$, a TBox $\mathcal{T}$, and an ABox $\mathcal{A}$, proceeds as follows. First, rl$^{\neq}$ computes the ABox $\mathcal{A}'$ obtained by evaluating the datalog program $\mathcal{P}_{rl}$ from Example 3.3 over $\mathcal{A}$. Second, rl$^{\neq}$ computes the query $\mathcal{Q}^{\neq}$ obtained from $\mathcal{Q}$ by adding to the body of each rule $r \in \mathcal{Q}$ an inequality $x \not\approx y$ for all pairs of distinct variables $x$ and $y$ occurring in $r$. Third, rl$^{\neq}$ evaluates $\mathcal{Q}^{\neq}$ over $\mathcal{A}'$ by considering $\mathcal{A}'$ as a *database*—that is, as a finite first-order interpretation in which each individual is mapped to itself (and thus different individuals are distinct). Thus, rl$^{\neq}$ characterises concrete reasoners that evaluate queries by matching different variables to different individuals. Abstract reasoner rl$^{\neq}$ is sound, monotonic, and weakly faithful, but it is not strongly faithful. For example, given query $\mathcal{Q} = \{R(x, y) \rightarrow Q(x)\}$, ABox $\mathcal{A} = \{R(a, b)\}$, and renaming $\mu = \{a \mapsto c, b \mapsto c\}$, we have rl$^{\neq}(\mathcal{Q}, \emptyset, \mathcal{A}) = \{a\}$, but rl$^{\neq}(\mathcal{Q}, \emptyset, \mu(\mathcal{A})) = \emptyset$. ◇

Example 3.11 suggests that, for an abstract reasoner to be strongly faithful, it should allow distinct variables in queries and axioms to be mapped to the same individuals.

We next identify classes of abstract reasoners that we use throughout this paper. Note that soundness is not required, which contributes to the generality of our results.

**Definition 3.12.** *Given a UCQ $\mathcal{Q}$ and a TBox $\mathcal{T}$, $\mathcal{C}_w^{\mathcal{Q}, \mathcal{T}}$ ($\mathcal{C}_s^{\mathcal{Q}, \mathcal{T}}$) is the class of all $(\mathcal{Q}, \mathcal{T})$-monotonic and weakly (strongly) $(\mathcal{Q}, \mathcal{T})$-faithful abstract reasoners applicable to $\mathcal{T}$.*

Finally, note that all the abstract reasoners introduced in Example 3.3 are sound, monotonic, and strongly (and therefore also weakly) faithful. Consequently, all concrete reasoners based on reasoning techniques outlined in Example 3.3 can be considered sound, monotonic, and strongly faithful, modulo implementation bugs.

## 3.2 Completeness of Abstract Reasoners

We next define the central notion of abstract reasoner completeness for a given query $\mathcal{Q}$ and TBox $\mathcal{T}$. Intuitively, a $(\mathcal{Q}, \mathcal{T})$-complete abstract reasoner is indistinguishable from a complete abstract reasoner when applied to $\mathcal{Q}$, $\mathcal{T}$, and an arbitrary ABox $\mathcal{A}$.

**Definition 3.13.** *Let $\mathcal{DL}$ be a description logic, and let ans be an abstract reasoner for $\mathcal{DL}$. Then, ans is $(\mathcal{Q}, \mathcal{T})$-complete for a UCQ $\mathcal{Q}$ and a $\mathcal{DL}$-TBox $\mathcal{T}$ if the following conditions hold for each ABox $\mathcal{A}$:*

- *if $\mathsf{cert}(*, \mathcal{T}, \mathcal{A}) = \mathsf{t}$, then $\mathsf{ans}(*, \mathcal{T}, \mathcal{A}) = \mathsf{t}$;*

- *if $\mathsf{cert}(*, \mathcal{T}, \mathcal{A}) = \mathsf{f}$ and $\mathsf{ans}(*, \mathcal{T}, \mathcal{A}) = \mathsf{f}$, then $\mathsf{cert}(\mathcal{Q}, \mathcal{T}, \mathcal{A}) \subseteq \mathsf{ans}(\mathcal{Q}, \mathcal{T}, \mathcal{A})$.*

*Finally, ans is complete if it is $(\mathcal{Q}, \mathcal{T})$-complete for each UCQ $\mathcal{Q}$ and each $\mathcal{DL}$-TBox $\mathcal{T}$.*

**Example 3.14.** Consider the $\mathcal{EL}$-TBox $\mathcal{T}$ consisting of the following axioms; the translation of the axioms into first-order logic is shown after the $\rightsquigarrow$ symbol.

$$\exists\mathsf{takesCo}.\mathsf{MathCo} \sqsubseteq \mathsf{St} \quad \rightsquigarrow \quad \forall x,y.[\mathsf{takesCo}(x,y) \wedge \mathsf{MathCo}(y) \rightarrow \mathsf{St}(x)] \tag{7}$$

$$\mathsf{CalcCo} \sqsubseteq \mathsf{MathCo} \quad \rightsquigarrow \quad \forall x.[\mathsf{CalcCo}(x) \rightarrow \mathsf{MathCo}(x)] \tag{8}$$

$$\mathsf{MathSt} \sqsubseteq \exists\mathsf{takesCo}.\mathsf{MathCo} \quad \rightsquigarrow \quad \forall x.[\mathsf{MathSt}(x) \rightarrow \exists y.[\mathsf{takesCo}(x,y) \wedge \mathsf{MathCo}(y)]] \tag{9}$$

$$\mathsf{St} \sqcap \mathsf{Prof} \sqsubseteq \bot \quad \rightsquigarrow \quad \forall x.[\mathsf{St}(x) \wedge \mathsf{Prof}(x) \rightarrow \bot] \tag{10}$$

Axiom (7) states that everyone taking a maths course is a student; axiom (8) states that each calculus course is also a maths course; axiom (9) states that each maths student takes some maths course; and axiom (10) states that no person can be both a student and a professor. Axiom (8) is an RDFS axiom, and all other axioms in $\mathcal{T}$ apart from (9) are OWL 2 RL axioms. Consider also query (11) that retrieves students taking a maths course.

$$\mathcal{Q} = \{\mathsf{St}(x) \wedge \mathsf{takesCo}(x,y) \wedge \mathsf{MathCo}(y) \rightarrow Q(x)\} \tag{11}$$

None of the abstract reasoners rdf, rdfs, rl, and classify from Example 3.3 are complete in general for answering UCQs over $\mathcal{EL}$-TBoxes. Furthermore, for $\mathcal{Q}$ and $\mathcal{T}$ from the previous paragraph, abstract reasoners rdf, rdfs, and rl are not $(\mathcal{Q}, \mathcal{T})$-complete, as all of them return the empty set of answers for ABox $\mathcal{A} = \{\mathsf{MathSt}(c)\}$. In contrast, in the following sections we will show that abstract reasoner classify is $(\mathcal{Q}, \mathcal{T})$-complete—that is, that it returns all certain answers to $\mathcal{Q}$, $\mathcal{T}$, and an arbitrary ABox $\mathcal{A}$. $\diamond$

### 3.3 Test Suites

Checking $(\mathcal{Q}, \mathcal{T})$-completeness of a concrete reasoner by applying the reasoner to all possible ABoxes and comparing the reasoner's answers with that of a complete reasoner is clearly infeasible in practice since there are infinitely many candidate input ABoxes. To obtain a practical approach, we need a *finite* number of tests. We formalise this idea using the following definition.

**Definition 3.15.** *Let $\mathcal{T}$ be a TBox. A $\mathcal{T}$-test suite is a pair $\mathbf{S} = \langle \mathbf{S}_\bot, \mathbf{S}_\mathcal{Q} \rangle$ where*

- $\mathbf{S}_\bot$ *is a finite set of ABoxes such that $\mathsf{cert}(*, \mathcal{T}, \mathcal{A}) = \mathsf{t}$ for each $\mathcal{A} \in \mathbf{S}_\bot$, and*

- $\mathbf{S}_\mathcal{Q}$ *is a finite set of pairs $\langle \mathcal{A}, \mathcal{Y} \rangle$ where $\mathcal{A}$ is an ABox such that $\mathsf{cert}(*, \mathcal{T}, \mathcal{A}) = \mathsf{f}$ and $\mathcal{Y}$ is a UCQ.*

*An abstract reasoner ans applicable to $\mathcal{T}$ passes a $\mathcal{T}$-test suite $\mathbf{S}$ if ans satisfies the following two conditions:*

- *for each $\mathcal{A} \in \mathbf{S}_\bot$, we have $\mathsf{ans}(*, \mathcal{T}, \mathcal{A}) = \mathsf{t}$, and*

- *for each $\langle \mathcal{A}, \mathcal{Y} \rangle \in \mathbf{S}_\mathcal{Q}$, if $\mathsf{ans}(*, \mathcal{T}, \mathcal{A}) = \mathsf{f}$, then $\mathsf{cert}(\mathcal{Y}, \mathcal{T}, \mathcal{A}) \subseteq \mathsf{ans}(\mathcal{Y}, \mathcal{T}, \mathcal{A})$.*

*Let $\mathcal{Q}$ be a UCQ, and let $\mathcal{C}$ be a class of abstract reasoners applicable to $\mathcal{T}$. Then, $\mathbf{S}$ is* exhaustive *for $\mathcal{C}$ and $\mathcal{Q}$ if each ans $\in \mathcal{C}$ that passes $\mathbf{S}$ is $(\mathcal{Q}, \mathcal{T})$-complete.*

*A $\mathcal{T}$-test suite $\mathbf{S}$ is $\mathcal{Q}$-simple if $\mathcal{Q}$ is the only query occurring in $\mathbf{S}_\mathcal{Q}$; then, $\mathbf{S}_\mathcal{Q}$ is commonly written as just a set of ABoxes, and $\langle \mathcal{A}, \mathcal{Q} \rangle \in \mathbf{S}_\mathcal{Q}$ is commonly abbreviated as $\mathcal{A} \in \mathbf{S}_\mathcal{Q}$.*

Intuitively, a $\mathcal{T}$-test suite $\mathbf{S} = \langle \mathbf{S}_\perp, \mathbf{S}_Q \rangle$ determines the tests that an abstract reasoner should be subjected to. For a reasoner to pass $\mathbf{S}$, it must correctly identify each ABox $\mathcal{A} \in \mathbf{S}_\perp$ as unsatisfiable, and for each ABox–query pair $\langle \mathcal{A}, \mathcal{Y} \rangle \in \mathbf{S}_Q$ the reasoner must correctly answer $\mathcal{Y}$ w.r.t. $\mathcal{T}$ and $\mathcal{A}$.

Given $\mathcal{Q}$ and $\mathcal{T}$, our goal is to identify a $\mathcal{T}$-test suite $\mathbf{S}$ that is exhaustive for $\mathcal{Q}$—that is, a test suite such that each abstract reasoner that passes $\mathbf{S}$ is guaranteed to be $(\mathcal{Q}, \mathcal{T})$-complete. Depending on the properties of abstract reasoners, however, different test suites may or may not achieve this goal. Therefore, the notion of exhaustiveness is relative to a class of abstract reasoners $\mathcal{C}$: if $\mathbf{S}$ is exhaustive for some class of abstract reasoners $\mathcal{C}$, then $\mathbf{S}$ can be used to test an arbitrary abstract reasoner in $\mathcal{C}$. Note that $\mathbf{S}$ depends on the target class of abstract reasoners, but not on the actual abstract reasoner being tested; in order words, the construction of $\mathbf{S}$ depends on the properties that one can assume to hold for the target abstract reasoner. Furthermore, if an abstract reasoner not contained in $\mathcal{C}$ passes $\mathbf{S}$, this will in general not imply a $(\mathcal{Q}, \mathcal{T})$-completeness guarantee.

**Example 3.16.** Let $\mathcal{Q}$ and $\mathcal{T}$ be as specified in Example 3.14, and let $\mathcal{A}_1$–$\mathcal{A}_6$ be the following ABoxes.

$$\begin{aligned}
&\mathcal{A}_1 = \{\mathsf{takesCo}(c, d), \mathsf{MathCo}(d)\} &\quad &\mathcal{A}_2 = \{\mathsf{takesCo}(c, c), \mathsf{MathCo}(c)\} \\
&\mathcal{A}_3 = \{\mathsf{takesCo}(c, d), \mathsf{CalcCo}(d)\} &\quad &\mathcal{A}_4 = \{\mathsf{takesCo}(c, c), \mathsf{CalcCo}(c)\} \\
&\mathcal{A}_5 = \{\mathsf{MathSt}(c)\} &\quad &\mathcal{A}_6 = \{\mathsf{St}(c), \mathsf{Prof}(c)\}
\end{aligned}$$

In the following sections, we will show that the $\mathcal{Q}$-simple $\mathcal{T}$-test suite $\mathbf{S} = \langle \mathbf{S}_\perp, \mathbf{S}_Q \rangle$ defined by $\mathbf{S}_\perp = \{\mathcal{A}_6\}$ and $\mathbf{S}_Q = \{\mathcal{A}_1, \ldots, \mathcal{A}_5\}$ is exhaustive for the class $\mathcal{C}_w^{\mathcal{Q}, \mathcal{T}}$ and $\mathcal{Q}$; consequently, $\mathbf{S}$ can be used to test all abstract reasoners from Example 3.3.

In particular, note that abstract reasoners rdf and rdfs fail all tests in $\mathbf{S}_\perp$ and $\mathbf{S}_Q$, and that abstract reasoner rl fails the test $\mathcal{A}_5 \in \mathbf{S}_Q$; furthermore, all failed tests provide a counterexample of $(\mathcal{Q}, \mathcal{T})$-completeness. In contrast, abstract reasoner classify from Example 3.14 passes the tests in $\mathbf{S}$, which implies that the abstract reasoner is indeed $(\mathcal{Q}, \mathcal{T})$-complete.

Finally, consider a variant of the abstract reasoner classify that, similarly to the abstract reasoner described in Example 3.9, returns the empty set of answers if the input ABox contains more than, say, ten assertions. Such an abstract reasoner is not $(\mathcal{Q}, \mathcal{T})$-monotonic and hence does not belong to $\mathcal{C}_w^{\mathcal{Q}, \mathcal{T}}$. This abstract reasoner clearly passes $\mathbf{S}$; however, since it does not belong to $\mathcal{C}_w^{\mathcal{Q}, \mathcal{T}}$, passing $\mathbf{S}$ (correctly) does not imply that the abstract reasoner is $(\mathcal{Q}, \mathcal{T})$-complete. $\diamondsuit$

We next state the following property, the proof of which is trivial.

**Proposition 3.17.** *Let $\mathcal{Q}$ be a UCQ, let $\mathcal{T}$ be a TBox, and let $\mathcal{C}_1$ and $\mathcal{C}_2$ be classes of abstract reasoners applicable to $\mathcal{T}$ such that $\mathcal{C}_1 \subseteq \mathcal{C}_2$.*

1. *If a $\mathcal{T}$-test suite $\mathbf{S}$ is exhaustive for $\mathcal{C}_2$ and $\mathcal{Q}$, then $\mathbf{S}$ is also exhaustive for $\mathcal{C}_1$ and $\mathcal{Q}$.*

2. *If no $\mathcal{T}$-test suite exists that is exhaustive for $\mathcal{C}_1$ and $\mathcal{Q}$, then no $\mathcal{T}$-test suite exists that is exhaustive for $\mathcal{C}_2$ and $\mathcal{Q}$.*

Therefore, when proving existence of a $\mathcal{T}$-test suite exhaustive for $\mathcal{Q}$, the most general result is the one that applies to the largest possible class of abstract reasoners. Furthermore, in the following section we will identify cases for which no $\mathcal{T}$-test suite exhaustive for $\mathcal{Q}$ can be found; by Proposition 3.17 it suffices to provide nonexistence results for the smallest possible class of abstract reasoners.

We finish this section by pointing out an important practically relevant property of $\mathcal{Q}$-simple $\mathcal{T}$-test suites, which has been illustrated in Example 3.16.

**Proposition 3.18.** *Let* $\mathbf{S} = \langle \mathbf{S}_\perp, \mathbf{S}_Q \rangle$ *be a* $\mathcal{Q}$-simple $\mathcal{T}$-test suite and let ans *be an abstract reasoner applicable to* $\mathcal{T}$. *If* ans *does not pass* $\mathbf{S}$, *then* ans *is not* $(\mathcal{Q}, \mathcal{T})$-complete.

*Proof.* The ABox in $\mathbf{S}_\perp$ or $\mathbf{S}_Q$ for which ans does not satisfy the conditions from Definition 3.15 is a counterexample for the $(\mathcal{Q}, \mathcal{T})$-completeness of ans. $\qquad\square$

Thus, a $\mathcal{Q}$-simple $\mathcal{T}$-test suite $\mathbf{S}$ exhaustive for $\mathcal{C}$ and $\mathcal{Q}$ provides a *sufficient and necessary test* for $(\mathcal{Q}, \mathcal{T})$-completeness of the abstract reasoners in $\mathcal{C}$. In contrast, if $\mathbf{S}$ is not $\mathcal{Q}$-simple, we show in Section 3.7 that then $\mathbf{S}$ provides only a sufficient, but not also a necessary test for $(\mathcal{Q}, \mathcal{T})$-completeness of the abstract reasoners in $\mathcal{C}$.

### 3.4 Negative Results

In Sections 3.5 (resp. Section 3.6) we identify restrictions on a UCQ $\mathcal{Q}$ and a TBox $\mathcal{T}$ that guarantee existence of $\mathcal{T}$-test suites exhaustive for $\mathcal{C}_w^{\mathcal{Q},\mathcal{T}}$ (resp. $\mathcal{C}_s^{\mathcal{Q},\mathcal{T}}$) and $\mathcal{Q}$. Before presenting these positive results, we first outline the limits of $(\mathcal{Q}, \mathcal{T})$-completeness testing and thus justify the restrictions we use in the following sections.

#### 3.4.1 MONOTONICITY AND WEAK FAITHFULNESS

Our approaches for testing $(\mathcal{Q}, \mathcal{T})$-completeness of abstract reasoners are applicable only to reasoners that are $(\mathcal{Q}, \mathcal{T})$-monotonic and weakly $(\mathcal{Q}, \mathcal{T})$-faithful. In this section, we provide a formal justification for these requirements in the form of the following two theorems.

- Theorem 3.19 shows that exhaustive test suites do not exist if we consider the class of abstract reasoners satisfying all properties from Section 3.1 *apart from* $(\mathcal{Q}, \mathcal{T})$-monotonicity; this includes soundness, strong faithfulness (which implies weak faithfulness), and monotonicity w.r.t. the *TBox only*.

- Theorem 3.20 shows that exhaustive test suites do not exist if we consider the class of abstract reasoners satisfying all properties defined in Section 3.1 *with the exception of* $(\mathcal{Q}, \mathcal{T})$-weak faithfulness; these properties include soundness and monotonicity.

The negative results of Theorems 3.19 and 3.20 are very strong: they hold for smallest classes of abstract reasoners we can define based on the notions introduced in Section 3.1 (by Proposition 3.17, the smaller the class of abstract reasoners, the more general the negative result); and they hold regardless of the $\mathcal{Q}$ and $\mathcal{T}$ considered (modulo a minor technicality: unlike Theorem 3.19, Theorem 3.20 requires $\mathcal{T}$ to be satisfiable).

The proof of Theorem 3.19 can intuitively be understood as follows. We first assume that $\mathbf{S}$ is a $\mathcal{T}$-test suite exhaustive for $\mathcal{Q}$ and the class of abstract reasoners to which the theorem applies. Then, we specify an abstract reasoner ans that 'does the right thing' (i.e., it returns the correct answer) when it is given as input the query $\mathcal{Q}$, the TBox $\mathcal{T}$, and an arbitrary ABox containing at most as many assertions as the largest test ABox in **S**; otherwise, ans returns a sound, but an incomplete answer. We finally show the following three properties of ans.

- Abstract reasoner ans belongs to the relevant class of abstract reasoners.

- Abstract reasoner ans passes **S**.

- Abstract reasoner ans is incomplete for at least one input ABox.

These three properties then show that **S** is not exhaustive for $\mathcal{Q}$ and the relevant class of abstract reasoners. Intuitively, this means that the class of abstract reasoners is 'too large', allowing abstract reasoners to treat their input in an erratic way.

**Theorem 3.19.** *Let $\mathcal{Q}$ be an arbitrary UCQ, and let $\mathcal{T}$ be an arbitrary admissible TBox. Then, no $\mathcal{T}$-test suite exists that is exhaustive for $\mathcal{Q}$ and the class of all sound and strongly faithful abstract reasoners applicable to $\mathcal{T}$ satisfying the following conditions for each TBox $\mathcal{T}'$ with $\mathcal{T} \subseteq \mathcal{T}'$ and each ABox $\mathcal{A}$:*

- $\mathsf{ans}(*, \mathcal{T}, \mathcal{A}) = \mathsf{t}$ *implies* $\mathsf{ans}(*, \mathcal{T}', \mathcal{A}) = \mathsf{t}$;

- $\mathsf{ans}(*, \mathcal{T}, \mathcal{A}) = \mathsf{f}$ *and* $\mathsf{ans}(*, \mathcal{T}', \mathcal{A}) = \mathsf{f}$ *imply* $\mathsf{ans}(\mathcal{Q}, \mathcal{T}, \mathcal{A}) \subseteq \mathsf{ans}(\mathcal{Q}, \mathcal{T}', \mathcal{A})$.

*Proof.* Consider an arbitrary $\mathcal{T}$-test suite $\mathbf{S} = \langle \mathbf{S}_\perp, \mathbf{S}_\mathcal{Q} \rangle$. Let $n$ be the maximum number of assertions in an ABox from **S**. Furthermore, let ans be the abstract reasoner that takes as input a UCQ $\mathcal{Q}_{in}$, an $\mathcal{FOL}$-TBox $\mathcal{T}_{in}$, and an ABox $\mathcal{A}_{in}$. The result of $\mathsf{ans}(*, \mathcal{T}_{in}, \mathcal{A}_{in})$ is determined as follows.

1. Try to find a renaming $\xi$ such that $\mathsf{dom}(\xi) = \mathsf{ind}(\mathcal{T})$ and $\xi(\mathcal{T}) \subseteq \mathcal{T}_{in}$; if no such $\xi$ exists, then return $\mathsf{f}$.

2. If $\mathcal{A}_{in}$ contains at most $n$ assertions, then check the satisfiability of $\xi(\mathcal{T}) \cup \mathcal{A}_{in}$ using a sound and complete reasoner; return $\mathsf{t}$ if $\xi(\mathcal{T}) \cup \mathcal{A}_{in}$ is unsatisfiable.

3. Return $\mathsf{f}$.

Furthermore, the result of $\mathsf{ans}(\mathcal{Q}_{in}, \mathcal{T}_{in}, \mathcal{A}_{in})$ is determined as follows.

4. Try to find a renaming $\xi$ such that $\mathsf{dom}(\xi) = \mathsf{ind}(\mathcal{Q} \cup \mathcal{T})$, $\xi(\mathcal{T}) \subseteq \mathcal{T}_{in}$, and $\xi(\mathcal{Q}) = \mathcal{Q}_{in}$; if no such $\xi$ exists, then return $\emptyset$.

5. If $\mathcal{A}_{in}$ contains at most $n$ assertions, then compute $\mathsf{cert}(\xi(\mathcal{Q}), \xi(\mathcal{T}), \mathcal{A}_{in})$ using a sound and complete reasoner and return the result.

6. Return $\emptyset$.

Since $\mathcal{T}$ is admissible, checks in steps 2 and 5 can be performed in finite time; furthermore, step 1 can be realised by enumerating all mappings from $\mathsf{ind}(\mathcal{T})$ to $\mathsf{ind}(\mathcal{T}_{in})$, and step 4 can be realised analogously; consequently, ans can be implemented such that it terminates on all inputs. To see that ans is sound and monotonic w.r.t. the TBox, consider arbitrary input $\mathcal{Q}_{in}$, $\mathcal{T}_{in}$, $\mathcal{T}'_{in}$, and $\mathcal{A}_{in}$ such that $\mathcal{T}_{in} \subseteq \mathcal{T}'_{in}$.

- Assume that $\mathsf{ans}(*, \mathcal{T}_{in}, \mathcal{A}_{in}) = \mathsf{t}$. Then, on $\mathcal{Q}_{in}$, $\mathcal{T}_{in}$, and $\mathcal{A}_{in}$ the abstract reasoner returns in step 2 because $\xi(\mathcal{T}) \cup \mathcal{A}_{in}$ is unsatisfiable; but then, since $\xi(\mathcal{T}) \subseteq \mathcal{T}_{in}$, we have that $\mathcal{T}_{in} \cup \mathcal{A}_{in}$ is unsatisfiable as well, as required for soundness. Furthermore, since $\xi(\mathcal{T}) \subseteq \mathcal{T}_{in} \subseteq \mathcal{T}'_{in}$, on $\mathcal{Q}_{in}$, $\mathcal{T}'_{in}$, and $\mathcal{A}_{in}$ the abstract reasoner returns in step 2 as well, so $\mathsf{ans}(*, \mathcal{T}'_{in}, \mathcal{A}_{in}) = \mathsf{t}$, as required for monotonicity w.r.t. the TBox.

- Assume that $\vec{a} \in \mathsf{ans}(\mathcal{Q}_{in}, \mathcal{T}_{in}, \mathcal{A}_{in})$. Then, on $\mathcal{Q}_{in}$, $\mathcal{T}_{in}$, and $\mathcal{A}_{in}$ the abstract reasoner returns in step 5, and therefore we have $\vec{a} \in \mathsf{cert}(\xi(\mathcal{Q}), \xi(\mathcal{T}), \mathcal{A}_{in})$; but then, since $\xi(\mathcal{Q}) = \mathcal{Q}_{in}$ and $\xi(\mathcal{T}) \subseteq \mathcal{T}_{in}$, we have $\vec{a} \in \mathsf{cert}(\mathcal{Q}_{in}, \mathcal{T}_{in}, \mathcal{A}_{in})$, as required for soundness. Furthermore, since $\xi(\mathcal{T}) \subseteq \mathcal{T}_{in} \subseteq \mathcal{T}'_{in}$, on $\mathcal{Q}_{in}$, $\mathcal{T}'_{in}$, and $\mathcal{A}_{in}$ the abstract reasoner returns in step 5 as well, so $\vec{a} \in \mathsf{ans}(\mathcal{Q}_{in}, \mathcal{T}'_{in}, \mathcal{A}_{in})$, as required for monotonicity w.r.t. the TBox.

To see that $\mathsf{ans}$ is strongly faithful, consider an arbitrary renaming $\lambda$. If renaming $\xi$ exists such that $\xi(Q) = \mathcal{Q}_{in}$ and $\xi(\mathcal{T}) \subseteq \mathcal{T}_{in}$, then clearly renaming $\xi'$ exists such that $\xi'(Q) = \lambda(\mathcal{Q}_{in})$ and $\xi'(\mathcal{T}) \subseteq \lambda(\mathcal{T}_{in})$. Consequently, if $\mathsf{ans}(*, \mathcal{T}_{in}, \mathcal{A}_{in})$ returns in step 2, then $\mathsf{ans}(*, \lambda(\mathcal{T}_{in}), \lambda(\mathcal{A}_{in}))$ returns in step 2 as well; similarly, if $\mathsf{ans}(\mathcal{Q}_{in}, \mathcal{T}_{in}, \mathcal{A}_{in})$ returns in step 5, then $\mathsf{ans}(\lambda(\mathcal{Q}_{in}), \lambda(\mathcal{T}_{in}), \lambda(\mathcal{A}_{in}))$ returns in step 5 as well; clearly, $\mathsf{ans}$ is strongly faithful. Finally, it is straightforward to see that $\mathsf{ans}$ passes $\mathbf{S}$.

Now let $\mathcal{A}$ be an ABox containing at least $n + 1$ assertions such that $\mathsf{cert}(\mathcal{Q}, \emptyset, \mathcal{A}) \neq \emptyset$; such $\mathcal{A}$ clearly exists. If $\mathcal{T} \cup \mathcal{A}$ is unsatisfiable, then $\mathsf{ans}(*, \mathcal{T}, \mathcal{A}) = \mathsf{f}$; furthermore, if $\mathcal{T} \cup \mathcal{A}$ is satisfiable, then $\mathsf{ans}(\mathcal{Q}, \mathcal{T}, \mathcal{A}) = \emptyset$; consequently, $\mathsf{ans}$ is not $(\mathcal{Q}, \mathcal{T})$-complete. Thus, $\mathbf{S}$ is not exhaustive for $\mathcal{Q}$ and the class of abstract reasoners considered in this theorem. $\quad\square$

We next prove Theorem 3.20. The proof is similar to the proof of Theorem 3.19, and the main difference is in the abstract reasoner $\mathsf{ans}$ we construct. In particular, given a test suite $\mathbf{S}$, we take $\mathsf{ans}$ to return the correct answer on the query $\mathcal{Q}$, the TBox $\mathcal{T}$, and each ABox that contains only the individuals occurring in $\mathbf{S}$; otherwise, the abstract reasoner returns a sound, but an incomplete answer. Again, the class of the abstract reasoners is 'too large', allowing $\mathsf{ans}$ to treat inputs in an erratic way.

Unlike Theorem 3.19, the following theorem requires $\mathcal{T}$ to be satisfiable; to understand why, consider an arbitrary unsatisfiable TBox $\mathcal{T}$ and UCQ $\mathcal{Q}$. Let $\mathbf{S} = \langle \mathbf{S}_\perp, \mathbf{S}_Q \rangle$ be the $\mathcal{T}$-test suite defined by $\mathbf{S}_\perp = \{\emptyset\}$ (i.e., $\mathbf{S}_\perp$ contains a single empty ABox) and $\mathbf{S}_Q = \emptyset$ (i.e., $\mathbf{S}_Q$ contains no ABoxes), and consider an arbitrary monotonic abstract reasoner $\mathsf{ans}$ that passes $\mathbf{S}_\perp$. Since $\mathsf{ans}$ passes $\mathbf{S}$, we have $\mathsf{ans}(*, \mathcal{T}, \emptyset) = \mathsf{t}$; but then, since $\mathsf{ans}$ is monotonic, for an arbitrary ABox $\mathcal{A}$ we have $\mathsf{ans}(*, \mathcal{T}, \mathcal{A}) = \mathsf{t}$ as well, which in turn implies that $\mathsf{ans}$ is $(\mathcal{Q}, \mathcal{T})$-complete. Failure to satisfy weak faithfulness is thus irrelevant if $\mathcal{T}$ is unsatisfiable.

**Theorem 3.20.** *Let $\mathcal{T}$ be an arbitrary admissible and satisfiable TBox and let $\mathcal{Q}$ be an arbitrary UCQ. Then, no $\mathcal{T}$-test suite exists that is exhaustive for $\mathcal{Q}$ and the class of all sound and monotonic abstract reasoners applicable to $\mathcal{T}$.*

*Proof.* Consider an arbitrary $\mathcal{T}$-test suite $\mathbf{S} = \langle \mathbf{S}_\perp, \mathbf{S}_Q \rangle$. Let $I$ be the set of all individuals occurring in $\mathbf{S}$, $\mathcal{Q}$, and $\mathcal{T}$. Furthermore, let $\mathsf{ans}$ be the abstract reasoner that takes as input a UCQ $\mathcal{Q}_{in}$, an $\mathcal{FOL}$-TBox $\mathcal{T}_{in}$, and an ABox $\mathcal{A}_{in}$. The result of $\mathsf{ans}(*, \mathcal{T}_{in}, \mathcal{A}_{in})$ is determined as follows.

1. If $\mathcal{T} \not\subseteq \mathcal{T}_{in}$, then return $\mathsf{f}$.

2. Let $\mathcal{A}_{in,I}$ be the set of assertions in $\mathcal{A}_{in}$ that mention only the individuals in $I$.

3. Check the satisfiability of $\mathcal{T} \cup \mathcal{A}_{in,I}$ using a sound and complete reasoner; return $\mathsf{t}$ if $\mathcal{T} \cup \mathcal{A}_{in,I}$ is unsatisfiable, and return $\mathsf{f}$ otherwise.

Furthermore, given a UCQ $\mathcal{Q}_{in}$, the result of $\mathsf{ans}(\mathcal{Q}_{in}, \mathcal{T}_{in}, \mathcal{A}_{in})$ is determined as follows:

4. If $\mathcal{T} \not\subseteq \mathcal{T}_{in}$ or $\mathcal{Q} \neq \mathcal{Q}_{in}$, then return $\emptyset$.

5. Let $\mathcal{A}_{in,I}$ be the set of assertions in $\mathcal{A}_{in}$ that mention only the individuals in $I$.

6. Compute $\mathsf{cert}(\mathcal{Q}, \mathcal{T}, \mathcal{A}_{in,I})$ using a sound and complete reasoner and return the result.

That $\mathsf{ans}$ can be implemented such that it terminates on all inputs can be shown as in the proof of Theorem 3.19. Furthermore, the soundness of $\mathsf{ans}$ follows from the following two observations.

- Assume that $\mathsf{ans}(*, \mathcal{T}_{in}, \mathcal{A}_{in}) = \mathsf{t}$. Then, the abstract reasoner returns in step 3 since $\mathcal{T} \cup \mathcal{A}_{in,I}$ is unsatisfiable; but then, since $\mathcal{T} \subseteq \mathcal{T}_{in}$ and $\mathcal{A}_{in,I} \subseteq \mathcal{A}_{in}$, we have that $\mathcal{T}_{in} \cup \mathcal{A}_{in}$ is unsatisfiable as well, as required.

- Assume that $\vec{a} \in \mathsf{ans}(\mathcal{Q}_{in}, \mathcal{T}_{in}, \mathcal{A}_{in})$. Then, the abstract reasoner returns in step 6, and therefore we have $\vec{a} \in \mathsf{cert}(\mathcal{Q}, \mathcal{T}, \mathcal{A}_{in,I})$; but then, since $\mathcal{Q} = \mathcal{Q}_{in}$, $\mathcal{T} \subseteq \mathcal{T}_{in}$, and $\mathcal{A}_{in,I} \subseteq \mathcal{A}_{in}$, we have $\vec{a} \in \mathsf{cert}(\mathcal{Q}_{in}, \mathcal{T}_{in}, \mathcal{A}_{in})$, as required.

For monotonicity, consider an arbitrary $\mathcal{T}'_{in}$ and $\mathcal{A}'_{in}$ such that $\mathcal{T}_{in} \subseteq \mathcal{T}'_{in}$ and $\mathcal{A}_{in} \subseteq \mathcal{A}'_{in}$; clearly, we have $\mathcal{T} \subseteq \mathcal{T}'_{in}$ and $\mathcal{A}_{in,I} \subseteq \mathcal{A}'_{in,I}$; but then, by monotonicity of first-order logic, $\mathsf{ans}(*, \mathcal{T}_{in}, \mathcal{A}_{in}) = \mathsf{t}$ implies $\mathsf{ans}(*, \mathcal{T}'_{in}, \mathcal{A}'_{in}) = \mathsf{t}$, and $\mathsf{ans}(\mathcal{Q}, \mathcal{T}_{in}, \mathcal{A}_{in}) \subseteq \mathsf{ans}(\mathcal{Q}, \mathcal{T}'_{in}, \mathcal{A}'_{in})$. Finally, it is straightforward to see that $\mathsf{ans}$ passes $\mathbf{S}$.

Now consider an arbitrary ABox $\mathcal{A}$ such that $\mathsf{ind}(\mathcal{A}) \cap I = \emptyset$ and $\mathsf{cert}(\mathcal{Q}, \emptyset, \mathcal{A}) \neq \emptyset$; such $\mathcal{A}$ clearly exists. If $\mathcal{T} \cup \mathcal{A}$ is unsatisfiable, since the ABox constructed in step 2 is empty and $\mathcal{T}$ is satisfiable, we have $\mathsf{ans}(*, \mathcal{T}, \mathcal{A}) = \mathsf{f}$; furthermore, if $\mathcal{T} \cup \mathcal{A}$ is satisfiable, since the ABox constructed in step 5 is empty, $\mathsf{ans}(\mathcal{Q}, \mathcal{T}, \mathcal{A})$ cannot contain individuals not occurring in $I$; consequently, $\mathsf{ans}$ is not $(\mathcal{Q}, \mathcal{T})$-complete. Thus, $\mathbf{S}$ is not exhaustive for $\mathcal{Q}$ and the class of abstract reasoners considered in this theorem. $\square$

### 3.4.2 Monotonicity and Weak Faithfulness do not Suffice

Next, we show that $(\mathcal{Q}, \mathcal{T})$-monotonicity and $(\mathcal{Q}, \mathcal{T})$-faithfulness in general do not guarantee existence of a $\mathcal{T}$-test suite exhaustive for $\mathcal{Q}$. In particular, Theorem 3.21 shows that, if $\mathcal{T}$ contains a single recursive axiom, no test suite exists that is exhaustive for the class of all sound, monotonic, and strongly faithful abstract reasoners (and by Proposition 3.17 for $\mathcal{C}_s^{\mathcal{Q},\mathcal{T}}$ and $\mathcal{C}_w^{\mathcal{Q},\mathcal{T}}$ as well, for each UCQ $\mathcal{Q}$). Although our result is applicable only to a particular $\mathcal{Q}$ and $\mathcal{T}$, it is straightforward to adapt the proof to any TBox with a recursive axiom that is 'relevant' to the given query. Example 3.22, however, shows that the concept of 'relevance' is rather difficult to formalise: even if $\mathcal{T}$ entails a recursive axiom, this axiom is not necessarily 'relevant' to answering the query. In order not to complicate matters any further, we state the following result for fixed $\mathcal{Q}$ and $\mathcal{T}$, and we hope that our proof clearly illustrates the limitations incurred by recursive axioms.

**Theorem 3.21.** *For $\mathcal{Q} = \{A(x) \rightarrow Q(x)\}$ and $\mathcal{T} = \{\exists R.A \sqsubseteq A\}$, no $\mathcal{T}$-test suite exists that is exhaustive for $\mathcal{Q}$ and the class of all sound, monotonic, and strongly faithful abstract reasoners applicable to $\mathcal{T}$.*

*Proof.* Consider an arbitrary $\mathcal{T}$-test suite $\mathbf{S} = \langle \mathbf{S}_\perp, \mathbf{S}_Q \rangle$. Since $\mathbf{S}$ is a $\mathcal{T}$-test suite, $\mathbf{S}_\perp$ contains only ABoxes $\mathcal{A}$ such that $\mathcal{T} \cup \mathcal{A}$ is unsatisfiable; clearly, no such ABox exists for $\mathcal{T}$ as stated in the theorem, so $\mathbf{S}_\perp = \emptyset$. Let $\mathbf{S}_Q$ be an arbitrary, but finite, set of pairs $\langle \mathcal{A}, \mathcal{Y} \rangle$ with $\mathcal{A}$ an ABox and $\mathcal{Y}$ a UCQ, and let $n$ be the maximum number of assertions in an ABox in $\mathbf{S}_Q$. Furthermore, consider the following ABox, where $a_i \neq a_j$ for all $1 \leq i < j \leq n+1$:

$$\mathcal{A}_{n+1} = \{R(a_0, a_1), \ldots, R(a_n, a_{n+1}), A(a_{n+1})\}$$

We next construct an abstract reasoner $\mathsf{pEval}_n$ with the following properties:

(P1) for each $\langle \mathcal{A}, \mathcal{Y} \rangle \in \mathbf{S}_Q$, we have $\mathsf{cert}(\mathcal{Y}, \mathcal{T}, \mathcal{A}) \subseteq \mathsf{pEval}_n(\mathcal{Y}, \mathcal{T}, \mathcal{A})$;

(P2) $a_0 \notin \mathsf{pEval}_n(\mathcal{Q}, \mathcal{T}, \mathcal{A}_{n+1})$; and

(P3) $\mathsf{pEval}_n$ is sound, monotonic, and strongly faithful.

Note that $a_0 \in \mathsf{cert}(\mathcal{Q}, \mathcal{T}, \mathcal{A}_{n+1})$, so the above three properties imply that $\mathbf{S}$ is not exhaustive for $\mathcal{Q}$ and the class of abstract reasoners considered in this theorem.

Abstract reasoner $\mathsf{pEval}_n$ accepts as input an $\mathcal{FOL}$-TBox $\mathcal{T}_{in}$ and an ABox $\mathcal{A}_{in}$. The result of $\mathsf{pEval}_n(*, \mathcal{T}_{in}, \mathcal{A}_{in})$ is determined as follows.

1. Return $\mathsf{f}$.

Furthermore, given a UCQ $\mathcal{Q}_{in}$, the result of $\mathsf{pEval}_n(\mathcal{Q}_{in}, \mathcal{T}_{in}, \mathcal{A}_{in})$ is determined as follows.

2. If $\mathcal{T} \not\subseteq \mathcal{T}_{in}$ or $\mathcal{Q} \neq \mathcal{Q}_{in}$, return $\emptyset$.

3. $\mathcal{A}_{sat} := \mathcal{A}_{in}$

4. Repeat the following computation $n$ times:

   - $\mathcal{A}_{sat} := \mathcal{A}_{sat} \cup \{\sigma(A(x)) \mid \sigma$ is a substitution s.t. $\{\sigma(R(x,y)), \sigma(A(y))\} \subseteq \mathcal{A}_{sat}\}$

5. Return $\mathsf{cert}(\mathcal{Q}, \emptyset, \mathcal{A}_{sat})$.

Abstract reasoner $\mathsf{pEval}_n$ clearly satisfies Property (P2) because deriving the assertion $A(a_0)$ requires $n+1$ iterations of the loop in step 4. Furthermore, $\mathsf{pEval}_n$ also satisfies (P1) because every ABox $\mathcal{A}$ occurring in $\mathbf{S}_Q$ contains at most $n$ individuals and $\mathcal{T}$ can be seen as the rule $R(x,y) \wedge A(y) \rightarrow A(x)$, which $\mathsf{pEval}_n$ applies $n$ times to the input ABox $\mathcal{A}_{in}$.

We finally show (P3). Abstract reasoner $\mathsf{pEval}_n$ is clearly sound. Furthermore, for each renaming $\mu$ we have $\mu(\mathcal{T}) = \mathcal{T}$ and $\mu(\mathcal{Q}) = \mathcal{Q}$, so $\mathsf{pEval}_n$ is clearly strongly faithful.

To show that $\mathsf{pEval}_n$ is monotonic, consider arbitrary $\mathcal{T}_{in}$, $\mathcal{T}'_{in}$, $\mathcal{A}_{in}$, and $\mathcal{A}'_{in}$ such that $\mathcal{T}_{in} \subseteq \mathcal{T}'_{in}$ and $\mathcal{A}_{in} \subseteq \mathcal{A}'_{in}$; since $\mathsf{pEval}_n(*, \mathcal{T}_{in}, \mathcal{A}_{in}) = \mathsf{f}$ for each input, the following are the only relevant cases.

- $\mathsf{pEval}_n$ returns $\emptyset$ in step 2 on input $\mathcal{Q}_{in}$, $\mathcal{T}'_{in}$, and $\mathcal{A}'_{in}$, in which case either $\mathcal{T} \not\subseteq \mathcal{T}'_{in}$ or $\mathcal{Q} \neq \mathcal{Q}_{in}$. Since $\mathcal{T}_{in} \subseteq \mathcal{T}'_{in}$, clearly $\mathsf{pEval}_n$ also returns $\emptyset$ in step 2 on input $\mathcal{Q}_{in}$, $\mathcal{T}_{in}$, and $\mathcal{A}_{in}$, and monotonicity holds.

- $\mathsf{pEval}_n$ returns in step 5 on input $\mathcal{Q}_{in}$, $\mathcal{T}'_{in}$, and $\mathcal{A}'_{in}$. Then, $\mathsf{pEval}_n$ can return in either step 2 or step 5 on input $\mathcal{Q}_{in}$, $\mathcal{T}_{in}$ and $\mathcal{A}_{in}$; in the former case, monotonicity holds trivially, and in the latter case, $\mathsf{pEval}_n(\mathcal{Q}_{in}, \mathcal{T}_{in}, \mathcal{A}_{in}) \subseteq \mathsf{pEval}_n(\mathcal{Q}_{in}, \mathcal{T}'_{in}, \mathcal{A}'_{in})$ follows directly from the fact that $\mathcal{A}_{in} \subseteq \mathcal{A}'_{in}$. $\qquad\square$

The following example shows that the presence of recursive axioms in $\mathcal{T}$ does not preclude the existence of a $\mathcal{T}$-test suite exhaustive for $\mathcal{Q}$.

**Example 3.22.** Consider $\mathcal{Q}$ and $\mathcal{T}$ defined as follows:

$$\mathcal{Q} = \{A(x) \wedge B(x) \rightarrow Q(x)\}$$
$$\mathcal{T} = \{\exists R.A \sqsubseteq A,\ B \sqsubseteq \exists R.A\}$$

Note that $\mathcal{T}$ contains the axiom mentioned in Theorem 3.21; however, note also that $\mathcal{T} \models B \sqsubseteq A$, and so

$$\mathcal{R} = \langle \emptyset, \{B(x) \rightarrow Q(x)\} \rangle$$

is a UCQ rewriting of $\mathcal{Q}$ w.r.t. $\mathcal{T}$. In Section 3.5 we show that the existence of a UCQ rewriting of $\mathcal{Q}$ w.r.t. $\mathcal{T}$ guarantees existence of a $\mathcal{Q}$-simple $\mathcal{T}$-test suite that is exhaustive for $\mathcal{C}_w^{\mathcal{Q},\mathcal{T}}$ (and hence also for $\mathcal{C}_s^{\mathcal{Q},\mathcal{T}}$) and $\mathcal{Q}$; for example, $\mathbf{S} = \langle \emptyset, \{\ \{B(a)\}\ \} \rangle$ is one such $\mathcal{T}$-test suite. Intuitively, $\mathcal{T} \models B \sqsubseteq A$ is the only consequence of $\mathcal{T}$ that is relevant for answering $\mathcal{Q}$; hence, for $\mathcal{T}' = \{B \sqsubseteq A\}$ and $\mathcal{Q}' = \{A(x) \rightarrow Q(x)\}$, we have that $\mathsf{cert}(\mathcal{Q}, \mathcal{T}, \mathcal{A}) = \mathsf{cert}(\mathcal{Q}', \mathcal{T}', \mathcal{A})$ for an arbitrary ABox $\mathcal{A}$. Hence, the recursive axiom in $\mathcal{T}$ is 'irrelevant' for answering $\mathcal{Q}$, and therefore its presence in $\mathcal{T}$ does not preclude the existence of a $\mathcal{T}$-test suite that is exhaustive for $\mathcal{C}_w^{\mathcal{Q},\mathcal{T}}$ and $\mathcal{Q}$. $\qquad\diamond$

## 3.5 Testing $(\mathcal{Q}, \mathcal{T})$-Monotonic and Weakly $(\mathcal{Q}, \mathcal{T})$-Faithful Abstract Reasoners

In this section, we identify a sufficient condition that guarantees existence of a $\mathcal{Q}$-simple $\mathcal{T}$-test suite $\mathbf{S}$ exhaustive for $\mathcal{C}_w^{\mathcal{Q},\mathcal{T}}$ and $\mathcal{Q}$; by Proposition 3.17, this result applies to $\mathcal{C}_s^{\mathcal{Q},\mathcal{T}}$ as well. Roughly speaking, such $\mathbf{S}$ can always be computed by 'instantiating' the rules in a UCQ rewriting of $\mathcal{Q}$ w.r.t. $\mathcal{T}$ in a suitable way. The requirement that $\mathcal{Q}$ should be UCQ-rewritable w.r.t. $\mathcal{T}$ invalidates the negative result of Theorem 3.21 since no UCQ rewriting of $\mathcal{Q} = \{A(x) \rightarrow Q(x)\}$ w.r.t. $\mathcal{T} = \{\exists R.A \sqsubseteq A\}$ exists.

This result allows one to compute $\mathcal{Q}$-simple $\mathcal{T}$-test suites exhaustive for $\mathcal{Q}$ in numerous practically relevant cases. In particular, a UCQ rewriting is guaranteed to exist if $\mathcal{T}$ is expressed in the DLs underpinning the QL profile of OWL 2 (Motik et al., 2009a; Calvanese et al., 2007); furthermore, as illustrated in Example 3.22, a UCQ rewriting may exist even if $\mathcal{T}$ is expressed in other fragments of OWL 2 such as the OWL 2 EL (Motik et al., 2009a; Baader, Brandt, & Lutz, 2005). In practice, such rewritings can be computed using systems such as QuOnto (Acciarri et al., 2005) and REQUIEM (Pérez-Urbina et al., 2010).

We establish the desired result in two steps. First, in Section 3.5.1 we present a general characterisation of $\mathcal{Q}$-simple $\mathcal{T}$-test suites exhaustive for $\mathcal{C}_w^{\mathcal{Q},\mathcal{T}}$ and $\mathcal{Q}$. Then, in Section 3.5.2 we use this characterisation to establish the desired connection between rewritings and $\mathcal{Q}$-simple $\mathcal{T}$-test suites exhaustive for $\mathcal{Q}$.

### 3.5.1 Characterisation of Simple and Exhaustive Test Suites

We next prove that a $\mathcal{Q}$-simple $\mathcal{T}$-test suite $\mathbf{S} = \langle \mathbf{S}_\perp, \mathbf{S}_Q \rangle$ is exhaustive for $\mathcal{C}_w^{\mathcal{Q},\mathcal{T}}$ and $\mathcal{Q}$ if and only if $\mathbf{S}$ contains an isomorphic copy of each 'data pattern' (i.e., a subset of an ABox) that can produce a certain answer to $\mathcal{Q}$ and $*$ w.r.t. $\mathcal{T}$, but that 'preserves' the identity of the individuals occurring in $\mathcal{T}$ and $\mathcal{Q}$. To show that this is not just a sufficient, but also a necessary condition for the existence of an exhaustive $\mathcal{T}$-test suite, we observe that, if $\mathbf{S}$ does not contain one such copy of a 'data pattern', we can always find an abstract reasoner in $\mathcal{C}_w^{\mathcal{Q},\mathcal{T}}$ that passes $\mathbf{S}$ but that misses certain answers obtained via the missing data pattern and that is therefore not $(\mathcal{Q}, \mathcal{T})$-complete.

**Theorem 3.23.** *Let $\mathcal{Q}$ be a UCQ, let $\mathcal{T}$ be an admissible TBox, and let $\mathbf{S} = \langle \mathbf{S}_\perp, \mathbf{S}_Q \rangle$ be a $\mathcal{Q}$-simple $\mathcal{T}$-test suite. Then, $\mathbf{S}$ is exhaustive for $\mathcal{C}_w^{\mathcal{Q},\mathcal{T}}$ and $\mathcal{Q}$ if and only if the following properties are satisfied for each ABox $\mathcal{A}$.*

1. *If $\mathcal{T} \cup \mathcal{A}$ is unsatisfiable, then there exist an ABox $\mathcal{A}' \in \mathbf{S}_\perp$ and an injective $\mathcal{T}$-stable renaming $\mu$ such that $\mathsf{dom}(\mu) = \mathsf{ind}(\mathcal{T} \cup \mathcal{A}')$ and $\mu(\mathcal{A}') \subseteq \mathcal{A}$.*

2. *If $\mathcal{T} \cup \mathcal{A}$ is satisfiable, then for each tuple $\vec{a} \in \mathsf{cert}(\mathcal{Q}, \mathcal{T}, \mathcal{A})$ there exist an ABox $\mathcal{A}' \in \mathbf{S}_Q$, a tuple $\vec{b} \in \mathsf{cert}(\mathcal{Q}, \mathcal{T}, \mathcal{A}')$, and an injective $(\mathcal{Q}, \mathcal{T})$-stable renaming $\mu$ such that $\mu(\vec{b}) = \vec{a}$, $\mathsf{dom}(\mu) = \mathsf{ind}(\mathcal{Q} \cup \mathcal{T} \cup \mathcal{A}')$, and $\mu(\mathcal{A}') \subseteq \mathcal{A}$.*

*Proof.* ($\Leftarrow$) Let $\mathbf{S}$ be an arbitrary $\mathcal{Q}$-simple $\mathcal{T}$-test suite that satisfies Properties 1 and 2; we next show that $\mathbf{S}$ is exhaustive for $\mathcal{C}_w^{\mathcal{Q},\mathcal{T}}$ and $\mathcal{Q}$. Consider an arbitrary abstract reasoner $\mathsf{ans} \in \mathcal{C}_w^{\mathcal{Q},\mathcal{T}}$ that passes $\mathbf{S}$—that is, $\mathsf{ans}$ satisfies the following two properties:

(a) $\mathsf{ans}(*, \mathcal{T}, \mathcal{A}') = \mathsf{t}$ for each $\mathcal{A}' \in \mathbf{S}_\perp$, and

(b) $\mathsf{ans}(*, \mathcal{T}, \mathcal{A}') = \mathsf{f}$ implies $\mathsf{cert}(\mathcal{Q}, \mathcal{T}, \mathcal{A}') \subseteq \mathsf{ans}(\mathcal{Q}, \mathcal{T}, \mathcal{A}')$ for each $\mathcal{A}' \in \mathbf{S}_Q$.

We next show that $\mathsf{ans}$ is $(\mathcal{Q}, \mathcal{T})$-complete—that is, that $\mathsf{ans}$ satisfies the two conditions in Definition 3.13 for an arbitrary ABox $\mathcal{A}$. For an arbitrary such $\mathcal{A}$, we have the following two possibilities, depending on the satisfiability of $\mathcal{T} \cup \mathcal{A}$.

Assume that $\mathcal{T} \cup \mathcal{A}$ is unsatisfiable. Since $\mathbf{S}$ satisfies Property 1, there exist an ABox $\mathcal{A}' \in \mathbf{S}_\perp$ and an injective $\mathcal{T}$-stable renaming $\mu$ s.t. $\mathsf{dom}(\mu) = \mathsf{ind}(\mathcal{T} \cup \mathcal{A}')$ and $\mu(\mathcal{A}') \subseteq \mathcal{A}$. By Condition (a) we have $\mathsf{ans}(*, \mathcal{T}, \mathcal{A}') = \mathsf{t}$. Since $\mathsf{ans}$ is weakly $(\mathcal{Q}, \mathcal{T})$-faithful, $\mu$ is injective and $\mathcal{T}$-stable, and $\mathsf{dom}(\mu) = \mathsf{ind}(\mathcal{T} \cup \mathcal{A}')$, we have $\mathsf{ans}(*, \mathcal{T}, \mu(\mathcal{A}')) = \mathsf{t}$; finally, since $\mathsf{ans}$ is $(\mathcal{Q}, \mathcal{T})$-monotonic and $\mu(\mathcal{A}') \subseteq \mathcal{A}$, we have $\mathsf{ans}(*, \mathcal{T}, \mathcal{A}) = \mathsf{t}$, as required by Definition 3.13.

Assume that $\mathcal{T} \cup \mathcal{A}$ is satisfiable and $\mathsf{ans}(*, \mathcal{T}, \mathcal{A}) = \mathsf{f}$. Furthermore, consider an arbitrary tuple $\vec{a} \in \mathsf{cert}(\mathcal{Q}, \mathcal{T}, \mathcal{A})$. Since $\mathbf{S}$ satisfies Property 2, there exist an ABox $\mathcal{A}' \in \mathbf{S}_Q$, a tuple $\vec{b} \in \mathsf{cert}(\mathcal{Q}, \mathcal{T}, \mathcal{A}')$, and an injective $(\mathcal{Q}, \mathcal{T})$-stable renaming $\mu$ such that $\mu(\vec{b}) = \vec{a}$, $\mathsf{dom}(\mu) = \mathsf{ind}(\mathcal{Q} \cup \mathcal{T} \cup \mathcal{A}')$, and $\mu(\mathcal{A}') \subseteq \mathcal{A}$. Since $\mu(\mathcal{A}') \subseteq \mathcal{A}$, $\mathsf{ans}(*, \mathcal{T}, \mathcal{A}) = \mathsf{f}$, and $\mathsf{ans}$ is $(\mathcal{Q}, \mathcal{T})$-monotonic, we have $\mathsf{ans}(*, \mathcal{T}, \mu(\mathcal{A}')) = \mathsf{f}$; furthermore, $\mathsf{ind}(\mathcal{T} \cup \mathcal{A}') \subseteq \mathsf{dom}(\mu)$, $\mu$ is injective and $(\mathcal{Q}, \mathcal{T})$-stable, and $\mathsf{ans}$ is weakly $(\mathcal{Q}, \mathcal{T})$-faithful, so $\mathsf{ans}(*, \mathcal{T}, \mu(\mathcal{A}')) = \mathsf{f}$ implies $\mathsf{ans}(*, \mathcal{T}, \mathcal{A}') = \mathsf{f}$. But then, by Condition (b) we have $\mathsf{cert}(\mathcal{Q}, \mathcal{T}, \mathcal{A}') \subseteq \mathsf{ans}(\mathcal{Q}, \mathcal{T}, \mathcal{A}')$, so $\vec{b} \in \mathsf{ans}(\mathcal{Q}, \mathcal{T}, \mathcal{A}')$. Since $\mathsf{ans}$ is weakly $(\mathcal{Q}, \mathcal{T})$-faithful, $\mu$ is injective and $(\mathcal{Q}, \mathcal{T})$-stable, and $\mathsf{dom}(\mu) = \mathsf{ind}(\mathcal{Q} \cup \mathcal{T} \cup \mathcal{A}')$, we have $\mu(\vec{b}) \in \mathsf{ans}(\mathcal{Q}, \mathcal{T}, \mu(\mathcal{A}'))$; since $\mu(\vec{b}) = \vec{a}$, we have $\vec{a} \in \mathsf{ans}(\mathcal{Q}, \mathcal{T}, \mu(\mathcal{A}'))$; finally, since $\mathsf{ans}$ is $(\mathcal{Q}, \mathcal{T})$-monotonic and $\mu(\mathcal{A}') \subseteq \mathcal{A}$, we then have $\vec{a} \in \mathsf{ans}(\mathcal{Q}, \mathcal{T}, \mathcal{A})$, as required by Definition 3.13.

($\Rightarrow$) Assume that $\mathbf{S}$ is exhaustive for $\mathcal{C}_w^{\mathcal{Q}, \mathcal{T}}$ and $\mathcal{Q}$; we next show that Properties 1 and 2 are satisfied for an arbitrary ABox $\mathcal{A}$. To this end, we consider a particular abstract reasoner $\mathsf{ans}$ for which we prove that $\mathsf{ans} \in \mathcal{C}_w^{\mathcal{Q}, \mathcal{T}}$ and that $\mathsf{ans}$ passes $\mathbf{S}$; this abstract reasoner will help us identify the ABox, the tuple, and the renaming required to prove Properties 1 and 2.

Let $\mathsf{ans}$ be the abstract reasoner that takes as input a UCQ $\mathcal{Q}_{in}$, an $\mathcal{FOL}$-TBox $\mathcal{T}_{in}$, and an ABox $\mathcal{A}_{in}$. The result of $\mathsf{ans}(*, \mathcal{T}_{in}, \mathcal{A}_{in})$ is determined as follows.

1. If $\mathcal{T} \neq \mathcal{T}_{in}$, then return $\mathsf{f}$.

2. For each ABox $\mathcal{A}' \in \mathbf{S}_\perp$, do the following.

    (a) Check the satisfiability of $\mathcal{T} \cup \mathcal{A}'$ using a sound, complete, and terminating reasoner.

    (b) If $\mathcal{T} \cup \mathcal{A}'$ is unsatisfiable, and if an injective $\mathcal{T}$-stable renaming $\mu$ exists such that $\mathsf{dom}(\mu) = \mathsf{ind}(\mathcal{T} \cup \mathcal{A}')$ and $\mu(\mathcal{A}') \subseteq \mathcal{A}_{in}$, then return $\mathsf{t}$.

3. Return $\mathsf{f}$.

Furthermore, the result of $\mathsf{ans}(\mathcal{Q}_{in}, \mathcal{T}_{in}, \mathcal{A}_{in})$ is determined as follows.

4. If $\mathcal{T} \neq \mathcal{T}_{in}$ or $\mathcal{Q} \neq \mathcal{Q}_{in}$, then return $\emptyset$.

5. $\mathsf{Out} := \emptyset$.

6. For each tuple $\vec{a}$ of constants occurring in $\mathcal{A}_{in}$ of arity equal to the arity of the query predicate of $\mathcal{Q}$, and for each $\mathcal{A}' \in \mathbf{S}_Q$ do the following.

    (a) Compute $\mathsf{C} := \mathsf{cert}(\mathcal{Q}, \mathcal{T}, \mathcal{A}')$ using a sound, complete and terminating reasoner.

    (b) If a tuple $\vec{b} \in \mathsf{C}$ and an injective $(\mathcal{Q}, \mathcal{T})$-stable renaming $\mu$ exist such that $\mu(\vec{b}) = \vec{a}$, $\mathsf{dom}(\mu) = \mathsf{ind}(\mathcal{Q} \cup \mathcal{T} \cup \mathcal{A}')$, and $\mu(\mathcal{A}') \subseteq \mathcal{A}_{in}$, then add $\vec{a}$ to $\mathsf{Out}$.

7. Return $\mathsf{Out}$.

We next show that $\mathsf{ans}$ belongs to $\mathcal{C}_w^{\mathcal{Q}, \mathcal{T}}$; to this end, we prove that $\mathsf{ans}$ terminates on all inputs, and that it is $(\mathcal{Q}, \mathcal{T})$-monotonic and weakly $(\mathcal{Q}, \mathcal{T})$-faithful.

*Termination.* Since $\mathcal{T}$ is admissible, checking satisfiability of $\mathcal{T} \cup \mathcal{A}'$ and the computation of $\mathsf{cert}(\mathcal{Q}, \mathcal{T}, \mathcal{A}')$ are decidable, so the relevant sound, complete and terminating reasoners exist. Furthermore, checking whether a $\mathcal{T}$-stable (resp. $(\mathcal{Q}, \mathcal{T})$-stable) injective renaming $\mu$ exists can be done by enumerating all renamings from $\mathsf{ind}(\mathcal{T} \cup \mathcal{A}')$ (resp. $\mathsf{ind}(\mathcal{Q} \cup \mathcal{T} \cup \mathcal{A}')$) to $\mathsf{ind}(\mathcal{T} \cup \mathcal{A}_{in})$ (resp. $\mathsf{ind}(\mathcal{Q} \cup \mathcal{T} \cup \mathcal{A}_{in})$). Therefore, $\mathsf{ans}$ can be implemented such that it terminates on each input.

$(\mathcal{Q}, \mathcal{T})$-*Monotonicity.* Consider arbitrary input $\mathcal{Q}_{in}$, $\mathcal{T}_{in}$, $\mathcal{A}_{in}$, and $\mathcal{A}'_{in}$ such that $\mathcal{A}_{in} \subseteq \mathcal{A}'_{in}$.

- Assume that $\mathsf{ans}(*, \mathcal{T}_{in}, \mathcal{A}_{in}) = \mathsf{t}$, so on $\mathcal{T}_{in}$ and $\mathcal{A}_{in}$ the abstract reasoner terminates in step 2(b) for some $\mathcal{A}' \in \mathbf{S}_\perp$ and $\mu$. But then, since $\mu(\mathcal{A}') \subseteq \mathcal{A}_{in} \subseteq \mathcal{A}'_{in}$, on $\mathcal{T}_{in}$ and $\mathcal{A}'_{in}$ the abstract reasoner also terminates in step 2(b), so $\mathsf{ans}(*, \mathcal{T}_{in}, \mathcal{A}'_{in}) = \mathsf{t}$, as required.

- Assume that $\mathsf{ans}(*, \mathcal{T}_{in}, \mathcal{A}_{in}) = \mathsf{f}$ and $\mathsf{ans}(*, \mathcal{T}_{in}, \mathcal{A}'_{in}) = \mathsf{f}$, and consider an arbitrary tuple $\vec{a} \in \mathsf{ans}(\mathcal{Q}_{in}, \mathcal{T}_{in}, \mathcal{A}_{in})$. Then $\vec{a}$ is added to $\mathsf{Out}$ in step 7(b) for some $\mathcal{A}' \in \mathbf{S}_Q$ and $\mu$. But then, since $\mu(\mathcal{A}') \subseteq \mathcal{A}_{in} \subseteq \mathcal{A}'_{in}$, on $\mathcal{Q}_{in}$, $\mathcal{T}_{in}$, and $\mathcal{A}'_{in}$ the abstract reasoner also adds $\vec{a}$ to $\mathsf{Out}$ in step 7(b), so $\vec{a} \in \mathsf{ans}(\mathcal{Q}_{in}, \mathcal{T}_{in}, \mathcal{A}'_{in})$, as required.

*Weak $(\mathcal{Q}, \mathcal{T})$-Faithfulness.* Consider an arbitrary input $\mathcal{Q}_{in}$, $\mathcal{T}_{in}$, and $\mathcal{A}_{in}$, and an arbitrary injective renaming $\lambda$.

- Assume that $\mathsf{ans}(*, \mathcal{T}_{in}, \mathcal{A}_{in}) = \mathsf{t}$, $\mathsf{dom}(\lambda) \subseteq \mathsf{ind}(\mathcal{T} \cup \mathcal{A})$, and $\lambda$ is $\mathcal{T}$-stable. Thus, on $\mathcal{T}_{in}$ and $\mathcal{A}_{in}$ the abstract reasoner terminates in step 2(b) for some $\mathcal{A}' \in \mathbf{S}_\perp$ and $\mu$. Let $\mu'$ be the renaming such that $\mu'(c) = \lambda(\mu(c))$ for each $c \in \mathsf{ind}(\mathcal{T} \cup \mathcal{A}')$. Clearly, we have $\mathsf{dom}(\mu') = \mathsf{ind}(\mathcal{T} \cup \mathcal{A}')$, renaming $\mu'$ is $\mathcal{T}$-stable and injective, and $\mu'(\mathcal{A}') \subseteq \lambda(\mathcal{A}_{in})$. Thus, on $\mathcal{T}_{in}$ and $\lambda(\mathcal{A}_{in})$ the abstract reasoner terminates in step 2(b), so we therefore have $\mathsf{ans}(*, \mathcal{T}_{in}, \lambda(\mathcal{A}_{in})) = \mathsf{t}$, as required.

- Assume that $\mathsf{ans}(*, \mathcal{T}_{in}, \mathcal{A}_{in}) = \mathsf{f}$, $\mathsf{dom}(\lambda) \subseteq \mathsf{ind}(\mathcal{Q} \cup \mathcal{T} \cup \mathcal{A})$, and $\lambda$ is $(\mathcal{Q}, \mathcal{T})$-stable, and consider an arbitrary truple $\vec{a} \in \mathsf{ans}(\mathcal{Q}_{in}, \mathcal{T}_{in}, \mathcal{A}_{in})$. Then $\vec{a}$ is added to $\mathsf{Out}$ in step 7(b) for some $\mathcal{A}' \in \mathbf{S}_Q$, $\mu$, and $\vec{b}$. Let $\mu'$ be the renaming defined s.t. $\mu'(c) = \lambda(\mu(c))$ for each individual $c \in \mathsf{ind}(\mathcal{Q} \cup \mathcal{T} \cup \mathcal{A}')$. Clearly, we have $\mathsf{dom}(\mu') = \mathsf{ind}(\mathcal{Q} \cup \mathcal{T} \cup \mathcal{A}')$, renaming $\mu'$ is $(\mathcal{Q}, \mathcal{T})$-stable and injective, $\mu'(\mathcal{A}') \subseteq \lambda(\mathcal{A}_{in})$, and $\mu'(\vec{b}) = \lambda(\vec{a})$. Thus, on $\mathcal{Q}_{in}$, $\mathcal{T}_{in}$, and $\lambda(\mathcal{A}_{in})$ the abstract reasoner terminates in step 7(b) and we clearly have $\lambda(\vec{a}) \in \mathsf{ans}(\mathcal{Q}_{in}, \mathcal{T}_{in}, \lambda(\mathcal{A}_{in}))$, as required.

This concludes the proof that $\mathsf{ans} \in \mathcal{C}_w^{\mathcal{Q}, \mathcal{T}}$. Furthermore, $\mathsf{ans}$ clearly passes $\mathbf{S}$; but then, since $\mathbf{S}$ is exhaustive for $\mathcal{C}_w^{\mathcal{Q}, \mathcal{T}}$ and $\mathcal{Q}$, abstract reasoner $\mathsf{ans}$ is $(\mathcal{Q}, \mathcal{T})$-complete. We next prove the main claim of this theorem. To this end, consider an arbitrary ABox $\mathcal{A}$; we have the following possibilities, depending on the satisfiability of $\mathcal{T} \cup \mathcal{A}$.

- Assume that $\mathcal{T} \cup \mathcal{A}$ is unsatisfiable. Then $\mathsf{ans}(*, \mathcal{T}, \mathcal{A}) = \mathsf{t}$, so the abstract reasoner returns in step 2(b) for some ABox $\mathcal{A}' \in \mathbf{S}_\perp$ and some $\mathcal{T}$-stable renaming $\mu$ such that $\mu(\mathcal{A}') \subseteq \mathcal{A}$ and $\mathsf{dom}(\mu) = \mathsf{ind}(\mathcal{T} \cup \mathcal{A}')$. Thus, Property 1 holds as required.

- Assume that $\mathcal{T} \cup \mathcal{A}$ is satisfiable, and consider an arbitrary tuple $\vec{a} \in \mathsf{cert}(\mathcal{Q}, \mathcal{T}, \mathcal{A})$. Then $\mathsf{ans}(*, \mathcal{T}, \mathcal{A}) = \mathsf{f}$ and $\vec{a} \in \mathsf{ans}(\mathcal{Q}, \mathcal{T}, \mathcal{A})$, so $\vec{a}$ is added to $\mathsf{Out}$ in step 7(b) for some ABox $\mathcal{A}' \in \mathbf{S}_Q$, tuple $\vec{b} \in \mathsf{cert}(\mathcal{Q}, \mathcal{T}, \mathcal{A}')$, and an injective $(\mathcal{Q}, \mathcal{T})$-stable renaming $\mu$ such that $\mu(\vec{b}) = \vec{a}$, $\mathsf{dom}(\mu) = \mathsf{ind}(\mathcal{Q} \cup \mathcal{T} \cup \mathcal{A}')$, and $\mu(\mathcal{A}') \subseteq \mathcal{A}$. Thus, Property 2 holds as required. □

The following example illustrates Theorem 3.23.

**Example 3.24.** Let $\mathcal{Q}$ and $\mathcal{T}$ be as specified in Example 3.14, and let $\mathbf{S} = \langle \mathbf{S}_\perp, \mathbf{S}_Q \rangle$ be specified in Example 3.16. As we show in Section 3.5.2, $\mathbf{S}$ is exhaustive for $\mathcal{C}_w^{\mathcal{Q}, \mathcal{T}}$ and $\mathcal{Q}$.

Consider now an ABox $\mathcal{A} = \{\mathsf{St}(a), \mathsf{MathSt}(b), \mathsf{takesCo}(a, b_1)\}$. Clearly, $\mathcal{T} \cup \mathcal{A}$ is satisfiable and $\mathsf{cert}(\mathcal{Q}, \mathcal{T}, \mathcal{A}) = \{b\}$. By Theorem 3.23, this certain answer can be obtained by evaluating $\mathcal{Q}$ w.r.t. $\mathcal{T}$ and an ABox in $\mathbf{S}_Q$. Indeed, note that ABox $\mathcal{A}_5 \in \mathbf{S}_Q$ is isomorphic to the subset $\mathcal{A}' = \{\mathsf{MathSt}(b)\}$ of $\mathcal{A}$ via renaming $\mu = \{b \mapsto c\}$, and that applying $\mathcal{Q}$ to $\mathcal{T}$ and $\mathcal{A}_5$ produces $c$, which is isomorphic to $b$ via $\mu$.

Note also that, if we remove $\mathcal{A}_5$ from **S**, we no longer have a $\mathcal{T}$-test suite that is exhaustive for $\mathcal{Q}$. For example, abstract reasoner rl from Example 3.16 would pass such a test suite, but it would not return the required certain answers when applied to $\mathcal{A}_5$ (and, consequently, when applied to $\mathcal{A}$ either). ◇

### 3.5.2 Computing Test Suites Exhaustive for $\mathcal{C}_w^{\mathcal{Q},\mathcal{T}}$

Based on Theorem 3.23, in this section we show that a $\mathcal{T}$-test suite exhaustive for $\mathcal{C}_w^{\mathcal{Q},\mathcal{T}}$ and $\mathcal{Q}$ can be obtained by instantiating a UCQ rewriting $\mathcal{R}$ of $\mathcal{Q}$ w.r.t. $\mathcal{T}$—that is, by replacing all variables in $\mathcal{R}$ with individuals in all possible ways. Please note that such an instantiation must be *full*, in the sense that all possible replacements must be considered. This is because the class $\mathcal{C}_w^{\mathcal{Q},\mathcal{T}}$ can contain abstract reasoners such as rl$^{\neq}$ from Example 3.11 that are not strongly faithful and that may incorrectly handle the case when distinct variables are bound to the same individuals.

**Definition 3.25.** *Let $I$ be a set of individuals, let $r$ be a datalog rule, and let $\sigma$ be a substitution. Then, $\sigma$ is an* instantiation substitution *for $r$ w.r.t. $I$ if $\sigma(x) \in I$ for each variable $x$ occurring in $r$. If the latter holds, then the* instantiation *of $r$ w.r.t. $\sigma$ is the ABox*

$$\mathcal{A}_\sigma^r := \{\sigma(B) \mid B \in \mathsf{body}(r)\}.$$

*Let $\mathcal{Q}$ be a UCQ, let $\mathcal{T}$ be a TBox, let $\mathcal{R} = \langle \mathcal{R}_\perp, \mathcal{R}_\mathcal{Q} \rangle$ be a UCQ rewriting of $\mathcal{Q}$ w.r.t. $\mathcal{T}$, let $m$ be the maximum number of distinct variables occurring in a rule in $\mathcal{R}$, and let $I$ be a set containing all individuals occurring in $\mathcal{R}$, $\mathcal{Q}$, and $\mathcal{T}$, as well as $m$ fresh individuals. The* full instantiation *of $\mathcal{R}$ w.r.t. $I$ is the pair $\mathbf{E}^{\mathcal{R},I} = \langle \mathbf{E}_\perp^{\mathcal{R},I}, \mathbf{E}_\mathcal{Q}^{\mathcal{R},I} \rangle$ where $\mathbf{E}_\perp^{\mathcal{R},I}$ and $\mathbf{E}_\mathcal{Q}^{\mathcal{R},I}$ are the smallest sets of ABoxes such that*

- *$\mathcal{A}_\sigma^r \in \mathbf{E}_\perp^{\mathcal{R},I}$ for each $r \in \mathcal{R}_\perp$ and each instantiation substitution $\sigma$ for $r$ w.r.t. $I$, and*

- *$\mathcal{A}_\sigma^r \in \mathbf{E}_\mathcal{Q}^{\mathcal{R},I}$ for each $r \in \mathcal{R}_\mathcal{Q}$ and each instantiation substitution $\sigma$ for $r$ w.r.t. $I$ such that $\mathsf{cert}(*, \mathcal{R}_\perp, \mathcal{A}_\sigma^r) = \mathsf{f}$.*

*$\mathbf{E}^{\mathcal{R},I}$ is clearly unique up to the renaming of the fresh individuals in $I$, so $I$ is typically left implicit, and one talks of the full instantiation $\mathbf{E}^\mathcal{R} = \langle \mathbf{E}_\perp^\mathcal{R}, \mathbf{E}_\mathcal{Q}^\mathcal{R} \rangle$ of $\mathcal{R}$.*

**Example 3.26.** Let $\mathcal{Q}$ and $\mathcal{T}$ be as specified in Example 3.14, and let $\mathcal{R} = \langle \mathcal{R}_\perp, \mathcal{R}_\mathcal{Q} \rangle$ be such that $\mathcal{R}_\perp = \{\mathsf{St}(x) \wedge \mathsf{Prof}(x) \to \perp\}$ and $\mathcal{R}_\mathcal{Q}$ consists of the following datalog rules:

$$\mathsf{takesCo}(x, y) \wedge \mathsf{MathCo}(y) \to Q(x)$$
$$\mathsf{takesCo}(x, y) \wedge \mathsf{CalcCo}(y) \to Q(x)$$
$$\mathsf{MathSt}(x) \to Q(x)$$

Then, $\mathcal{R}$ is a UCQ rewriting of $\mathcal{Q}$ w.r.t. $\mathcal{T}$, and one can see that the $\mathcal{Q}$-simple $\mathcal{T}$-test suite $\mathbf{S} = \langle \mathbf{S}_\perp, \mathbf{S}_\mathcal{Q} \rangle$ from Example 3.16 is the full instantiation of $\mathcal{R}$. ◇

The following theorem shows that the full instantiation of a UCQ rewriting of $\mathcal{Q}$ w.r.t. $\mathcal{T}$ is a $\mathcal{Q}$-simple $\mathcal{T}$-test suite exhaustive for $\mathcal{C}_w^{\mathcal{Q},\mathcal{T}}$ and $\mathcal{Q}$. According to this theorem, the $\mathcal{T}$-test suite **S** in Example 3.26 is exhaustive for $\mathcal{C}_w^{\mathcal{Q},\mathcal{T}}$ and $\mathcal{Q}$.

**Theorem 3.27.** *Let $\mathcal{Q}$ be a UCQ, let $\mathcal{T}$ be an admissible TBox, let $\mathcal{R} = \langle \mathcal{R}_\perp, \mathcal{R}_Q \rangle$ be a UCQ rewriting of $\mathcal{Q}$ w.r.t. $\mathcal{T}$, and let $\mathbf{E}^\mathcal{R} = \langle \mathbf{E}_\perp^\mathcal{R}, \mathbf{E}_Q^\mathcal{R} \rangle$ be the full instantiation of $\mathcal{R}$. Then, $\mathbf{E}^\mathcal{R}$ is a $\mathcal{Q}$-simple $\mathcal{T}$-test suite that is exhaustive for $\mathcal{C}_w^{\mathcal{Q},\mathcal{T}}$ and $\mathcal{Q}$.*

*Proof.* Let $I$ be the set of individuals that $\mathbf{E}^\mathcal{R}$ is obtained from. We first show that $\mathbf{E}^\mathcal{R}$ is a $\mathcal{Q}$-simple $\mathcal{T}$-test suite—that is, that it satisfies the two properties in Definition 3.15.

- Consider an arbitrary ABox $\mathcal{A} \in \mathbf{E}_\perp^\mathcal{R}$. Then, a rule $r \in \mathcal{R}_\perp$ and an instantiation substitution $\sigma$ for $r$ exist such that $\mathcal{A} = \mathcal{A}_\sigma^r$; clearly $\mathsf{cert}(*, \{r\}, \mathcal{A}) = \mathsf{t}$; since $\mathcal{R}$ is a UCQ rewriting, $\mathcal{T} \cup \mathcal{A}$ is unsatisfiable, as required.

- Consider an arbitrary ABox $\mathcal{A} \in \mathbf{E}_Q^\mathcal{R}$. Then, $\mathsf{cert}(*, \mathcal{R}_\perp, \mathcal{A}) = \mathsf{f}$ by Definition 3.25; since $\mathcal{R}$ is a UCQ rewriting, $\mathcal{T} \cup \mathcal{A}$ is satisfiable, as required.

We next show that $\mathbf{E}^\mathcal{R}$ satisfies Properties 1 and 2 of Theorem 3.23 for an arbitrary ABox $\mathcal{A}$.

(Property 1) Assume that $\mathcal{T} \cup \mathcal{A}$ is unsatisfiable. Since $\mathcal{R}$ is a UCQ rewriting, by Definition 2.2 we have $\mathsf{cert}(*, \mathcal{R}_\perp, \mathcal{A}) = \mathsf{t}$; but then, a rule $r \in \mathcal{R}_\perp$ and a substitution $\rho$ exist such that $\mathcal{A}_\rho^r \subseteq \mathcal{A}$ and $\mathsf{cert}(*, \{r\}, \mathcal{A}_\rho^r) = \mathsf{t}$. Let $\nu$ be an injective renaming such that for each individual $c$ occurring in $\mathcal{R}$ or $\mathcal{T}$ we have $\nu(c) = c$, and for each individual $d$ occurring in $\mathcal{A}_\rho^r$ but not in $\mathcal{R}$ and $\mathcal{T}$ we have that $\nu(d)$ is a fresh individual in $I$; such $\nu$ exists since the number of variables in $r$ is smaller or equal to the number of fresh individuals in $I$. Let $\sigma$ be an instantiation substitution for $r$ such that $\sigma(x) = \nu(\rho(x))$ for each variable $x$ occurring in $r$; then $\mathcal{A}_\sigma^r \in \mathbf{E}_\perp^\mathcal{R}$ holds since $\mathbf{E}^\mathcal{R}$ is the full instantiation of $\mathcal{R}$ w.r.t. $I$. Let $\mu$ be any injective renaming that coincides with the inverse of $\nu$ on each individual occurring in $\mathcal{A}_\sigma^r$, $\mathcal{R}$, or $\mathcal{T}$; such $\mu$ exists since $\nu$ is injective and the range of $\nu$ contains each individual occurring in $\mathcal{A}_\sigma^r$, $\mathcal{R}$, and $\mathcal{T}$. Clearly $\mu(\mathcal{A}_\sigma^r) = \mathcal{A}_\rho^r$ holds, so $\mu(\mathcal{A}_\sigma^r) \subseteq \mathcal{A}$. Furthermore, $\mu$ is clearly $\mathcal{T}$-stable. Thus, Property (1) is satisfied for $\mathcal{A}_\sigma^r \in \mathbf{E}_\perp^\mathcal{R}$ and $\mu$.

(Property 2) Assume that $\mathcal{T} \cup \mathcal{A}$ is satisfiable, and consider an arbitrarily chosen tuple $\vec{a} \in \mathsf{cert}(\mathcal{Q}, \mathcal{T}, \mathcal{A})$. Since $\mathcal{R}$ is a UCQ rewriting, by Definition 2.2 we have $\mathsf{cert}(*, \mathcal{R}_\perp, \mathcal{A}) = \mathsf{f}$ and $\vec{a} \in \mathsf{cert}(\mathcal{R}_Q, \mathcal{R}_\perp, \mathcal{A})$; but then, clearly $\vec{a} \in \mathsf{cert}(\mathcal{R}_Q, \emptyset, \mathcal{A})$ as well. Then, a rule $r \in \mathcal{R}_Q$ and a substitution $\rho$ exist such that $\mathcal{A}_\rho^r \subseteq \mathcal{A}$ and $\vec{a} \in \mathsf{cert}(\{r\}, \emptyset, \mathcal{A}_\rho^r)$. Let $\nu$ be an injective renaming such that for each individual $c$ occurring in $\mathcal{R}$, $\mathcal{Q}$, or $\mathcal{T}$ we have $\nu(c) = c$, and for each individual $d$ occurring in $\mathcal{A}_\rho^r$ but not in $\mathcal{R}$, $\mathcal{Q}$, and $\mathcal{T}$ we have that $\nu(d)$ is a fresh individual in $I$; such $\nu$ clearly exists since the number of variables in $r$ is smaller or equal to the number of fresh individuals in $I$. Let $\sigma$ be an instantiation substitution for $r$ such that $\sigma(x) = \nu(\rho(x))$ for each variable $x$ occurring in $r$; then $\mathcal{A}_\sigma^r \in \mathbf{E}_Q^\mathcal{R}$ holds since $\mathbf{E}^\mathcal{R}$ is the full instantiation of $\mathcal{R}$ w.r.t. $I$. Let $\mu$ be any injective renaming that coincides with the inverse of $\nu$ on each individual occurring in $\mathcal{A}_\sigma^r$, $\mathcal{R}$, $\mathcal{Q}$, or $\mathcal{T}$; such $\mu$ exists since $\nu$ is injective and the range of $\nu$ contains each individual occurring in $\mathcal{A}_\sigma^r$, $\mathcal{R}$, $\mathcal{Q}$, and $\mathcal{T}$. Furthermore, clearly a tuple $\vec{b} \in \mathsf{cert}(\{r\}, \emptyset, \mathcal{A}_\sigma^r)$ exists such that $\sigma(\mathsf{head}(r)) = Q(\vec{b})$; since $\mathcal{R}$ is a UCQ rewriting and $\mathcal{T} \cup \mathcal{A}_\sigma^r$ is satisfiable, we have $\vec{b} \in \mathsf{cert}(\mathcal{Q}, \mathcal{T}, \mathcal{A}_\sigma^r)$; furthermore, since $\mu$ is injective, $\mu(\vec{b}) = \vec{a}$ clearly holds. But then, Property (2) is satisfied for $\mathcal{A}_\sigma^r \in \mathbf{E}_Q^\mathcal{R}$, $\mu$, and $\vec{b}$. $\qquad\square$

### 3.5.3 Minimising Exhaustive Test Suites

In practice, it is clearly beneficial to compute test suites that are as 'small' as possible. This goal can be achieved by applying known techniques for minimising UCQ rewritings (Calvanese et al., 2007; Pérez-Urbina, Horrocks, & Motik, 2009). By Theorem 3.27, the smallest such rewriting can be instantiated to obtain an exhaustive test suite.

State of the art query rewriting systems employ *subsumption* and *condensation* techniques in order to reduce the size of a rewriting. A datalog rule $r$ subsumes a datalog rule $r'$ if a substitution $\sigma$ exists such that $\sigma(r) \subseteq r'$; intuitively, $r$ is then 'more general' than $r'$. If a rewriting contains such rules $r$ and $r'$, then $r'$ can be safely removed from the rewriting. Furthermore, if a rule $r$ contains distinct unifiable body atoms $B_i$ and $B_j$, a condensation of $r$ is the rule $\sigma(r)$ where $\sigma$ is the most general unifier of $B_i$ and $B_j$. If a rewriting contains such rule $r$ and $\sigma(r)$ subsumes $r$, the rule can safely be replaced with $\sigma(r)$. The following example illustrates how these techniques can be used to obtain small test suites.

**Example 3.28.** Let $\mathcal{Q}$ and $\mathcal{T}$ be as specified in Example 3.14, and let $\mathcal{R}$ be the rewriting of $\mathcal{Q}$ w.r.t. $\mathcal{T}$ from Example 3.26. Then $\mathcal{R}' = \langle \mathcal{R}_\perp, \mathcal{R}'_Q \rangle$ where $\mathcal{R}'_Q$ consists of the following rules is also a UCQ rewriting of $\mathcal{Q}$ w.r.t. $\mathcal{T}$.

$$\mathsf{takesCo}(x,y) \wedge \mathsf{takesCo}(x,z) \wedge \mathsf{MathCo}(y) \to Q(x) \tag{12}$$

$$\mathsf{takesCo}(x,x) \wedge \mathsf{CalcCo}(x) \wedge \mathsf{MathCo}(x) \to Q(x) \tag{13}$$

$$\mathsf{takesCo}(x,y) \wedge \mathsf{CalcCo}(y) \to Q(x) \tag{14}$$

$$\mathsf{St}(x) \wedge \mathsf{MathSt}(x) \to Q(x) \tag{15}$$

$$\mathsf{MathSt}(x) \to Q(x) \tag{16}$$

By Theorem 3.27, the full instantiation of $\mathcal{R}'$ is also a $\mathcal{T}$-test suite exhaustive for $\mathcal{C}_w^{\mathcal{Q},\mathcal{T}}$ and $\mathcal{Q}$. The rewriting $\mathcal{R}'$, however, contains redundancy and hence the resulting test suite is unnecessarily large. In particular, by applying condensation to query (12), subsumption to queries (13) and (14), and subsumption again to queries (15) and (16), we can obtain the simpler rewriting $\mathcal{R}$. ◇

Finally, note that the test suites obtained via full instantiation can contain isomorphic ABoxes. Clearly, all isomorphic copies of an ABox can safely be eliminated from a test suite without losing exhaustiveness for $\mathcal{C}_w^{\mathcal{Q},\mathcal{T}}$ and $\mathcal{Q}$.

## 3.6 Testing $(\mathcal{Q}, \mathcal{T})$-Monotonic and Strongly $(\mathcal{Q}, \mathcal{T})$-Faithful Abstract Reasoners

Due to full instantiation, test suites obtained by Definition 3.25 can be exponentially larger than the rewriting they are generated from. As a result, even rewritings of moderate size can yield test suites containing thousands of ABoxes. Intuitively, full instantiation is required to obtain a test suite exhaustive for the class $\mathcal{C}_w^{\mathcal{Q},\mathcal{T}}$ because this class contains abstract reasoners such as $\mathsf{rl}^{\neq}$ from Example 3.11, which do not correctly handle the case when distinct variables in a query are matched to the same individual.

In this section, we show that test suites exhaustive for the class $\mathcal{C}_s^{\mathcal{Q},\mathcal{T}}$ can be obtained by an *injective* instantiation of a rewriting—that is, by replacing each variable with a distinct fresh individual. Test suites obtained in such a way are linear in the size of the rewriting, and are thus substantially smaller than test suites obtained by full instantiation.

**Example 3.29.** Let $\mathcal{Q}$ and $\mathcal{T}$ be as specified in Example 3.14, and let $\mathbf{S} = \langle \mathbf{S}_\perp, \mathbf{S}_Q \rangle$ be the $\mathcal{Q}$-simple $\mathcal{T}$-test suite from Example 3.16. Furthermore, consider the abstract reasoner $\mathsf{rl}^{\neq}$ from Example 3.11 that is weakly, but not strongly $(\mathcal{Q}, \mathcal{T})$-faithful. It is easy to check that $\mathsf{rl}^{\neq}$ returns complete answers on $\mathcal{A}_1$ and $\mathcal{A}_3$, but not on $\mathcal{A}_2$ and $\mathcal{A}_4$. Therefore, by Theorem 3.27, for $\mathbf{S}$ to be exhaustive for $\mathcal{C}_w^{\mathcal{Q}, \mathcal{T}}$ and $\mathcal{Q}$, we must include in $\mathbf{S}_Q$ ABoxes $\mathcal{A}_2$ and $\mathcal{A}_4$, which are respectively obtained from ABoxes $\mathcal{A}_1$ and $\mathcal{A}_3$ by merging individual $d$ into $c$.

Strongly $(\mathcal{Q}, \mathcal{T})$-faithful abstract reasoners, however, correctly handle inputs obtained by merging individuals. Based on this observation, in this section we show that the $\mathcal{Q}$-simple $\mathcal{T}$-test suite $\mathbf{S}' = \langle \mathbf{S}_\perp, \mathbf{S}'_Q \rangle$ where $\mathbf{S}'_Q = \{\mathcal{A}_1, \mathcal{A}_3, \mathcal{A}_5\}$, obtained by injectively instantiating the rewriting $\mathcal{R}$ from Example 3.26, is exhaustive for $\mathcal{C}_s^{\mathcal{Q}, \mathcal{T}}$ and $\mathcal{Q}$. $\diamond$

As in Section 3.5, we first develop a characterisation of $\mathcal{Q}$-simple $\mathcal{T}$-test suites that are exhaustive for $\mathcal{C}_s^{\mathcal{Q}, \mathcal{T}}$ and $\mathcal{Q}$; this result is analogous to Theorem 3.23.

**Theorem 3.30.** *Let $\mathcal{Q}$ be a UCQ, let $\mathcal{T}$ be an admissible TBox, and let $\mathbf{S} = \langle \mathbf{S}_\perp, \mathbf{S}_Q \rangle$ be a $\mathcal{Q}$-simple $\mathcal{T}$-test suite. Then, $\mathbf{S}$ is exhaustive for $\mathcal{C}_s^{\mathcal{Q}, \mathcal{T}}$ and $\mathcal{Q}$ if and only if the following properties are satisfied for each ABox $\mathcal{A}$.*

1. *If $\mathcal{T} \cup \mathcal{A}$ is unsatisfiable, then there exist an ABox $\mathcal{A}' \in \mathbf{S}_\perp$ and a $\mathcal{T}$-stable renaming $\mu$ such that $\mathsf{dom}(\mu) = \mathsf{ind}(\mathcal{T} \cup \mathcal{A}')$ and $\mu(\mathcal{A}') \subseteq \mathcal{A}$.*

2. *If $\mathcal{T} \cup \mathcal{A}$ is satisfiable, then for each tuple $\vec{a} \in \mathsf{cert}(\mathcal{Q}, \mathcal{T}, \mathcal{A})$ there exist an ABox $\mathcal{A}' \in \mathbf{S}_Q$, a tuple $\vec{b} \in \mathsf{cert}(\mathcal{Q}, \mathcal{T}, \mathcal{A}')$, and a $(\mathcal{Q}, \mathcal{T})$-stable renaming $\mu$ such that $\mu(\vec{b}) = \vec{a}$, $\mathsf{dom}(\mu) = \mathsf{ind}(\mathcal{Q} \cup \mathcal{T} \cup \mathcal{A}')$, and $\mu(\mathcal{A}') \subseteq \mathcal{A}$.*

*Proof.* ($\Leftarrow$) Let $\mathbf{S}$ be an arbitrary $\mathcal{Q}$-simple $\mathcal{T}$-test suite that satisfies Properties 1 and 2; we next show that $\mathbf{S}$ is exhaustive for $\mathcal{C}_s^{\mathcal{Q}, \mathcal{T}}$ and $\mathcal{Q}$. Consider an arbitrary abstract reasoner $\mathsf{ans} \in \mathcal{C}_s^{\mathcal{Q}, \mathcal{T}}$ that passes $\mathbf{S}$—that is, $\mathsf{ans}$ satisfies the following two properties:

(a) $\mathsf{ans}(*, \mathcal{T}, \mathcal{A}') = \mathsf{t}$ for each $\mathcal{A}' \in \mathbf{S}_\perp$, and

(b) $\mathsf{ans}(*, \mathcal{T}, \mathcal{A}') = \mathsf{f}$ implies $\mathsf{cert}(\mathcal{Q}, \mathcal{T}, \mathcal{A}') \subseteq \mathsf{ans}(\mathcal{Q}, \mathcal{T}, \mathcal{A}')$ for each $\mathcal{A}' \in \mathbf{S}_Q$.

We next show that $\mathsf{ans}$ is $(\mathcal{Q}, \mathcal{T})$-complete—that is, that $\mathsf{ans}$ satisfies the two conditions in Definition 3.13 for an arbitrary ABox $\mathcal{A}$. For an arbitrary such $\mathcal{A}$, we have the following two possibilities, depending on the satisfiability of $\mathcal{T} \cup \mathcal{A}$.

Assume that $\mathcal{T} \cup \mathcal{A}$ is unsatisfiable. Since $\mathbf{S}$ satisfies Property 1, there exist an ABox $\mathcal{A}' \in \mathbf{S}_\perp$ and a $\mathcal{T}$-stable renaming $\mu$ such that $\mathsf{dom}(\mu) = \mathsf{ind}(\mathcal{T} \cup \mathcal{A}')$ and $\mu(\mathcal{A}') \subseteq \mathcal{A}$. By Condition (a) we have $\mathsf{ans}(*, \mathcal{T}, \mathcal{A}') = \mathsf{t}$. Since $\mathsf{ans}$ is strongly $(\mathcal{Q}, \mathcal{T})$-faithful and $\mu$ is $\mathcal{T}$-stable, we have $\mathsf{ans}(*, \mathcal{T}, \mu(\mathcal{A}')) = \mathsf{t}$; finally, since $\mathsf{ans}$ is $(\mathcal{Q}, \mathcal{T})$-monotonic and $\mu(\mathcal{A}') \subseteq \mathcal{A}$, we have $\mathsf{ans}(*, \mathcal{T}, \mathcal{A}) = \mathsf{t}$, as required by Definition 3.13.

Assume that $\mathcal{T} \cup \mathcal{A}$ is satisfiable and $\mathsf{ans}(*, \mathcal{T}, \mathcal{A}) = \mathsf{f}$. Furthermore, consider an arbitrary tuple $\vec{a} \in \mathsf{cert}(\mathcal{Q}, \mathcal{T}, \mathcal{A})$. Since $\mathbf{S}$ satisfies Property 2, there exist an ABox $\mathcal{A}' \in \mathbf{S}_Q$, a tuple $\vec{b} \in \mathsf{cert}(\mathcal{Q}, \mathcal{T}, \mathcal{A}')$, and a $(\mathcal{Q}, \mathcal{T})$-stable renaming $\mu$ such that $\mu(\vec{b}) = \vec{a}$, $\mu(\mathcal{A}') \subseteq \mathcal{A}$, and $\mathsf{dom}(\mu) = \mathsf{ind}(\mathcal{Q} \cup \mathcal{T} \cup \mathcal{A}')$. Since $\mu(\mathcal{A}') \subseteq \mathcal{A}$, $\mathsf{ans}(*, \mathcal{T}, \mathcal{A}) = \mathsf{f}$, and $\mathsf{ans}$ is $(\mathcal{Q}, \mathcal{T})$-monotonic, we have $\mathsf{ans}(*, \mathcal{T}, \mu(\mathcal{A}')) = \mathsf{f}$; furthermore, $\mu$ is $(\mathcal{Q}, \mathcal{T})$-stable and $\mathsf{ans}$ is strongly faithful, so $\mathsf{ans}(*, \mathcal{T}, \mu(\mathcal{A}')) = \mathsf{f}$ implies $\mathsf{ans}(*, \mathcal{T}, \mathcal{A}') = \mathsf{f}$. But then, by Condition (b) we have $\mathsf{cert}(\mathcal{Q}, \mathcal{T}, \mathcal{A}') \subseteq \mathsf{ans}(\mathcal{Q}, \mathcal{T}, \mathcal{A}')$, so $\vec{b} \in \mathsf{ans}(\mathcal{Q}, \mathcal{T}, \mathcal{A}')$. Now $\mathsf{ans}$ is strongly $(\mathcal{Q}, \mathcal{T})$-faithful and $\mu$ is $(\mathcal{Q}, \mathcal{T})$-stable, so $\mu(\vec{b}) \in \mathsf{ans}(\mathcal{Q}, \mathcal{T}, \mu(\mathcal{A}'))$; since $\mu(\vec{b}) = \vec{a}$, we have $\vec{a} \in \mathsf{ans}(\mathcal{Q}, \mathcal{T}, \mu(\mathcal{A}'))$; finally, since $\mathsf{ans}$ is $(\mathcal{Q}, \mathcal{T})$-monotonic and $\mu(\mathcal{A}') \subseteq \mathcal{A}$, we have $\vec{a} \in \mathsf{ans}(\mathcal{Q}, \mathcal{T}, \mathcal{A})$, as required by Definition 3.13.

($\Rightarrow$) Assume that $\mathbf{S}$ is exhaustive for $\mathcal{C}_s^{\mathcal{Q},\mathcal{T}}$ and $\mathcal{Q}$; we next show that Properties 1 and 2 are satisfied for an arbitrary ABox $\mathcal{A}$. To this end, we consider a particular abstract reasoner $\mathsf{ans}$ for which we prove that $\mathsf{ans} \in \mathcal{C}_s^{\mathcal{Q},\mathcal{T}}$ and that $\mathsf{ans}$ passes $\mathbf{S}$; this abstract reasoner will help us identify the ABox, the tuple, and the renaming required to prove Properties 1 and 2.

Let $\mathsf{ans}$ be the abstract reasoner that takes as input a UCQ $\mathcal{Q}_{in}$, an $\mathcal{FOL}$-TBox $\mathcal{T}_{in}$, and an ABox $\mathcal{A}_{in}$. The result of $\mathsf{ans}(*, \mathcal{T}_{in}, \mathcal{A}_{in})$ is determined as follows.

1. If $\mathcal{T} \neq \mathcal{T}_{in}$, then return $\mathsf{f}$.

2. For each ABox $\mathcal{A}' \in \mathbf{S}_\perp$, do the following.

    (a) Check the satisfiability of $\mathcal{T} \cup \mathcal{A}'$ using a sound, complete, and terminating reasoner.

    (b) If $\mathcal{T} \cup \mathcal{A}'$ is unsatisfiable, and if a $\mathcal{T}$-stable renaming $\mu$ exists such that $\mathsf{dom}(\mu) = \mathsf{ind}(\mathcal{T} \cup \mathcal{A}')$ and $\mu(\mathcal{A}') \subseteq \mathcal{A}_{in}$, then return $\mathsf{t}$.

3. Return $\mathsf{f}$.

Furthermore, the result of $\mathsf{ans}(\mathcal{Q}_{in}, \mathcal{T}_{in}, \mathcal{A}_{in})$ is determined as follows.

4. If $\mathcal{T} \neq \mathcal{T}_{in}$ or $\mathcal{Q} \neq \mathcal{Q}_{in}$, then return $\emptyset$.

5. $\mathsf{Out} := \emptyset$.

6. For each tuple $\vec{a}$ of constants occurring in $\mathcal{A}_{in}$ of arity equal to the arity of the query predicate of $\mathcal{Q}$, and for each $\mathcal{A}' \in \mathbf{S}_Q$ do the following.

    (a) Compute $\mathsf{C} := \mathsf{cert}(\mathcal{Q}, \mathcal{T}, \mathcal{A}')$ using a sound, complete and terminating reasoner.

    (b) If a tuple $\vec{b} \in \mathsf{C}$ and a $(\mathcal{Q}, \mathcal{T})$-stable renaming $\mu$ exist such that $\mu(\vec{b}) = \vec{a}$, $\mathsf{dom}(\mu) = \mathsf{ind}(\mathcal{Q} \cup \mathcal{T} \cup \mathcal{A}')$, and $\mu(\mathcal{A}') \subseteq \mathcal{A}_{in}$, then add $\vec{a}$ to $\mathsf{Out}$.

7. Return $\mathsf{Out}$.

We next show that $\mathsf{ans}$ belongs to $\mathcal{C}_s^{\mathcal{Q},\mathcal{T}}$. The proofs that $\mathsf{ans}$ terminates and that it is $(\mathcal{Q}, \mathcal{T})$-monotonic are analogous to the proofs in Theorem 3.23. To show strong $(\mathcal{Q}, \mathcal{T})$-faithfulness, consider an arbitrary $\mathcal{Q}_{in}$, $\mathcal{T}_{in}$, and $\mathcal{A}_{in}$, and an arbitrary renaming $\lambda$.

- Assume that $\mathsf{ans}(*, \mathcal{T}_{in}, \mathcal{A}_{in}) = \mathsf{t}$ and $\lambda$ is $\mathcal{T}$-stable. Thus, on $\mathcal{T}_{in}$ and $\mathcal{A}_{in}$ the abstract reasoner terminates in step 2(b) for some $\mathcal{A}' \in \mathbf{S}_\perp$ and $\mu$. Let $\mu'$ be the renaming such that $\mu'(c) = \lambda(\mu(c))$ for each $c \in \mathsf{ind}(\mathcal{T} \cup \mathcal{A}')$. Clearly, we have $\mathsf{dom}(\mu') = \mathsf{ind}(\mathcal{T} \cup \mathcal{A}')$, renaming $\mu'$ is $\mathcal{T}$-stable, and $\mu'(\mathcal{A}') \subseteq \lambda(\mathcal{A}_{in})$. Thus, on $\mathcal{T}_{in}$ and $\lambda(\mathcal{A}_{in})$ the abstract reasoner terminates in step 2(b), so we have $\mathsf{ans}(*, \mathcal{T}_{in}, \lambda(\mathcal{A}_{in})) = \mathsf{t}$, as required.

- Assume that $\mathsf{ans}(*, \mathcal{T}_{in}, \mathcal{A}_{in}) = \mathsf{f}$ and $\lambda$ is $(\mathcal{Q}, \mathcal{T})$-stable, and consider an arbitrary tuple $\vec{a} \in \mathsf{ans}(\mathcal{Q}_{in}, \mathcal{T}_{in}, \mathcal{A}_{in})$. Then $\vec{a}$ is added to $\mathsf{Out}$ in step 7(b) for some $\mathcal{A}' \in \mathbf{S}_Q$, $\mu$, and $\vec{b}$. Let $\mu'$ be the renaming defined such that $\mu'(c) = \lambda(\mu(c))$ for each individual $c \in \mathsf{ind}(\mathcal{Q} \cup \mathcal{T} \cup \mathcal{A}')$. Clearly, we have $\mathsf{dom}(\mu') = \mathsf{ind}(\mathcal{Q} \cup \mathcal{T} \cup \mathcal{A}')$, mapping $\mu'$ is $(\mathcal{Q}, \mathcal{T})$-stable, $\mu'(\mathcal{A}') \subseteq \lambda(\mathcal{A}_{in})$, and $\mu'(\vec{b}) = \lambda(\vec{a})$. Thus, on $\mathcal{Q}_{in}$, $\mathcal{T}_{in}$, and $\lambda(\mathcal{A}_{in})$ the abstract reasoner terminates in step 7(b), so $\lambda(\vec{a}) \in \mathsf{ans}(\mathcal{Q}_{in}, \mathcal{T}_{in}, \lambda(\mathcal{A}_{in}))$, as required.

This concludes the proof that $\mathsf{ans} \in \mathcal{C}_s^{\mathcal{Q},\mathcal{T}}$. Furthermore, $\mathsf{ans}$ clearly passes $\mathbf{S}$; but then, since $\mathbf{S}$ is exhaustive for $\mathcal{C}_s^{\mathcal{Q},\mathcal{T}}$ and $\mathcal{Q}$, abstract reasoner $\mathsf{ans}$ is $(\mathcal{Q}, \mathcal{T})$-complete. The main claim of this theorem can now be shown as in Theorem 3.23. □

We next use Theorem 3.30 to show that a $\mathcal{Q}$-simple $\mathcal{T}$-test suite that is exhaustive for $\mathcal{C}_s^{\mathcal{Q},\mathcal{T}}$ and $\mathcal{Q}$ can be obtained as an injective instantiation of a UCQ rewriting of $\mathcal{Q}$ w.r.t. $\mathcal{T}$.

**Definition 3.31.** *Let $\mathcal{Q}$ be a UCQ, let $\mathcal{T}$ be a TBox, let $\mathcal{R} = \langle \mathcal{R}_\perp, \mathcal{R}_\mathcal{Q} \rangle$ be a UCQ rewriting of $\mathcal{Q}$ w.r.t. $\mathcal{T}$, and let $\lambda$ be a substitution mapping each variable occurring in $\mathcal{R}$ into a distinct fresh individual. The* injective instantiation *of $\mathcal{R}$ w.r.t. $\lambda$ is the pair $\mathbf{I}^{\mathcal{R},\lambda} = \langle \mathbf{I}_\perp^{\mathcal{R},\lambda}, \mathbf{I}_Q^{\mathcal{R},\lambda} \rangle$ where $\mathbf{I}_\perp^{\mathcal{R},\lambda}$ and $\mathbf{I}_Q^{\mathcal{R},\lambda}$ are the smallest sets of ABoxes such that*

- $\mathcal{A}_\lambda^r \in \mathbf{I}_\perp^{\mathcal{R},\lambda}$ *for each $r \in \mathcal{R}_\perp$, and*

- $\mathcal{A}_\lambda^r \in \mathbf{I}_Q^{\mathcal{R},\lambda}$ *for each $r \in \mathcal{R}_Q$ such that $\mathsf{cert}(*, \mathcal{R}_\perp, \mathcal{A}_\lambda^r) = \mathsf{f}$.*

$\mathbf{I}^{\mathcal{R},\lambda}$ *is clearly unique up to the renaming of the fresh individuals in $\lambda$, so $\lambda$ is typically left implicit, and one talks of the* injective instantiation *$\mathbf{I}^\mathcal{R} = \langle \mathbf{I}_\perp^\mathcal{R}, \mathbf{I}_Q^\mathcal{R} \rangle$ of $\mathcal{R}$.*

**Theorem 3.32.** *Let $\mathcal{Q}$ be a UCQ, let $\mathcal{T}$ be an admissible TBox, let $\mathcal{R} = \langle \mathcal{R}_\perp, \mathcal{R}_Q \rangle$ be a UCQ rewriting of $\mathcal{Q}$ w.r.t. $\mathcal{T}$, and let $\mathbf{I}^\mathcal{R} = \langle \mathbf{I}_\perp^\mathcal{R}, \mathbf{I}_Q^\mathcal{R} \rangle$ be the injective instantiation of $\mathcal{R}$. Then, $\mathbf{I}^\mathcal{R}$ is a $\mathcal{Q}$-simple $\mathcal{T}$-test suite that is exhaustive for $\mathcal{C}_s^{\mathcal{Q},\mathcal{T}}$ and $\mathcal{Q}$.*

*Proof.* Let $\lambda$ be the substitution that $\mathbf{I}^\mathcal{R}$ is obtained from. We first show that $\mathbf{I}^\mathcal{R}$ is a $\mathcal{Q}$-simple $\mathcal{T}$-test suite—that is, that it satisfies the two properties in Definition 3.15.

- Consider an arbitrary $\mathcal{A} \in \mathbf{I}_\perp^\mathcal{R}$. Then, a rule $r \in \mathcal{R}_\perp$ exist such that $\mathcal{A} = \mathcal{A}_\lambda^r$; clearly $\mathsf{cert}(*, \{r\}, \mathcal{A}) = \mathsf{t}$; since $\mathcal{R}$ is a UCQ rewriting, $\mathcal{T} \cup \mathcal{A}$ is unsatisfiable, as required.

- Consider an arbitrary $\mathcal{A} \in \mathbf{I}_Q^\mathcal{R}$. Then, $\mathsf{cert}(*, \mathcal{R}_\perp, \mathcal{A}) = \mathsf{f}$ by Definition 3.31; since $\mathcal{R}$ is a UCQ rewriting, $\mathcal{T} \cup \mathcal{A}$ is satisfiable, as required.

We next show that $\mathbf{I}^\mathcal{R}$ satisfies Properties 1 and 2 of Theorem 3.30 for an arbitrary ABox $\mathcal{A}$.

(Property 1) Assume that $\mathcal{T} \cup \mathcal{A}$ is unsatisfiable. Since $\mathcal{R}$ is a UCQ rewriting, by Definition 2.2 we have $\mathsf{cert}(*, \mathcal{R}_\perp, \mathcal{A}) = \mathsf{t}$; but then, a rule $r \in \mathcal{R}_\perp$ and a substitution $\rho$ exist such that $\mathcal{A}_\rho^r \subseteq \mathcal{A}$ and $\mathsf{cert}(*, \{r\}, \mathcal{A}_\rho^r) = \mathsf{t}$. Let $\mu$ be a renaming such that for each individual $c$ occurring in $\mathcal{R}$ or $\mathcal{T}$ we have $\mu(c) = c$, and for each variable $x$ in $r$ we have $\mu(\lambda(x)) = \rho(x)$. Clearly, $\mu(\mathcal{A}_\lambda^r) = \mathcal{A}_\rho^r$, so $\mu(\mathcal{A}_\lambda^r) \subseteq \mathcal{A}$. Furthermore, it is clear that $\mu$ is $\mathcal{T}$-stable. Thus, Property (1) holds for $\mathcal{A}_\lambda^r \in \mathbf{I}_\perp^\mathcal{R}$ and $\mu$.

(Property 2) Assume that $\mathcal{T} \cup \mathcal{A}$ is satisfiable, and consider an arbitrarily chosen tuple $\vec{a} \in \mathsf{cert}(\mathcal{Q}, \mathcal{T}, \mathcal{A})$. Since $\mathcal{R}$ is a UCQ rewriting, by Definition 2.2 we have $\mathsf{cert}(*, \mathcal{R}_\perp, \mathcal{A}) = \mathsf{f}$ and $\vec{a} \in \mathsf{cert}(\mathcal{R}_Q, \mathcal{R}_\perp, \mathcal{A})$; but then, clearly $\vec{a} \in \mathsf{cert}(\mathcal{R}_Q, \emptyset, \mathcal{A})$ as well. Then, a rule $r \in \mathcal{R}_Q$ and a substitution $\rho$ exist such that $\mathcal{A}_\rho^r \subseteq \mathcal{A}$ and $\vec{a} \in \mathsf{cert}(\{r\}, \emptyset, \mathcal{A}_\rho^r)$. Let $\mu$ be the renaming such that for each individual $c$ occurring in $\mathcal{R}$, $\mathcal{Q}$, or $\mathcal{T}$ we have $\mu(c) = c$, and for each variable $x$ in $r$ we have $\mu(\lambda(x)) = \rho(x)$. Clearly, $\mu(\mathcal{A}_\lambda^r) = \mathcal{A}_\rho^r$, so $\mu(\mathcal{A}_\lambda^r) \subseteq \mathcal{A}$. Furthermore, it is clear that $\mu$ is $(\mathcal{Q}, \mathcal{T})$-stable. Finally, clearly a tuple $\vec{b} \in \mathsf{cert}(\{r\}, \emptyset, \mathcal{A}_\lambda^r)$ exists such that $\lambda(\mathsf{head}(r)) = Q(\vec{b})$; since $\mathcal{R}$ is a UCQ rewriting and $\mathcal{T} \cup \mathcal{A}_\lambda^r$ is satisfiable, we have $\vec{b} \in \mathsf{cert}(\mathcal{Q}, \mathcal{T}, \mathcal{A}_\lambda^r)$; furthermore, $\mu(\vec{b}) = \vec{a}$ clearly holds. But then, Property (2) is satisfied for $\mathcal{A}_\lambda^r \in \mathbf{I}_Q^{\mathcal{R}}$, $\mu$, and $\vec{b}$. $\qquad\square$

### 3.7 Dealing with Recursive Axioms

The negative result in Theorem 3.21 (which applies to both $\mathcal{C}_w^{\mathcal{Q},\mathcal{T}}$ and $\mathcal{C}_s^{\mathcal{Q},\mathcal{T}}$) depends on the presence of a recursive axiom in the TBox; thus, the positive results in Sections 3.5 and 3.6 require the input UCQ to be rewritable w.r.t. the input TBox, which effectively prohibits recursion in TBox axioms. Instead of disallowing recursive axioms, in this section we overcome the limitation of Theorem 3.21 by placing additional requirements on the abstract reasoners by requiring them to be *first-order reproducible*. Intuitively, the latter means that the reasoner's behaviour can be seen as complete reasoning in some unknown first-order theory. Such abstract reasoners are not allowed to partially evaluate recursive axioms, which invalidates the approach used to prove Theorem 3.21.

We show that a $\mathcal{T}$-test suite exhaustive for $\mathcal{Q}$ and the class of first-order reproducible abstract reasoners can be obtained by instantiating a datalog$^{\pm,\vee}$ rewriting of $\mathcal{Q}$ w.r.t. $\mathcal{T}$. Such rewritings exist for a wide range of TBoxes and queries, which in turn allows our results to be applicable to a range of practically interesting cases. In contrast to test suites computed from a UCQ rewriting, however, the test suites obtained from a datalog$^{\pm,\vee}$ rewriting may not be $\mathcal{Q}$-simple. In fact, we show in Section 3.7.2 that, for certain $\mathcal{Q}$ and $\mathcal{T}$, a $\mathcal{T}$-test suite exhaustive for $\mathcal{Q}$ and the class of first-order reproducible abstract reasoners exists, but no such test suite is $\mathcal{Q}$-simple. This has an important practically-relevant consequence: if a $\mathcal{T}$-test suite $\mathbf{S}$ is not $\mathcal{Q}$-simple, a first-order reproducible abstract reasoner that passes $\mathbf{S}$ is guaranteed to be $(\mathcal{Q}, \mathcal{T})$-complete; however, if an abstract reasoner does not pass $\mathbf{S}$, in general we cannot conclude that the reasoner is not $(\mathcal{Q}, \mathcal{T})$-complete.

#### 3.7.1 FIRST-ORDER REPRODUCIBLE ABSTRACT REASONERS

State of the art concrete reasoners such as Oracle's reasoner, Jena, OWLim, Minerva, Virtuoso, and DLE-Jena are all implemented as RDF triple stores extended with deductive database features. Given $\mathcal{T}$ and $\mathcal{A}$ as input, these reasoners first precompute all assertions that follow from $\mathcal{T} \cup \mathcal{A}$ in a preprocessing step. In practice, this step is commonly implemented by (a technique that can be seen as) evaluating a datalog program over $\mathcal{A}$. After preprocessing, these reasoners can then answer an arbitrary UCQ $\mathcal{Q}$ by simply evaluating $\mathcal{Q}$ in the precomputed set of assertions.

Motivated by this observation, we next introduce the new class of *first-order reproducible* abstract reasoners—that is, abstract reasoners whose behaviour can be conceived as complete reasoning in some unknown first-order theory. Note that this theory is not required to be a datalog program; for example, it can contain existential quantifiers, which can be used to capture the behaviour of concrete reasoners such as Jena and OWLim (Bishop, Kiryakov,

Ognyanoff, Peikov, Tashev, & Velkov, 2011) that handle existential quantifiers in the input by introducing fresh individuals.

**Definition 3.33.** *An abstract reasoner* ans *for a description logic* $\mathcal{DL}$ *is* first-order reproducible *if, for each* $\mathcal{DL}$-TBox $\mathcal{T}$, *a set of first-order sentences* $\mathcal{F}_{\mathcal{T}}$ *exists such that, for each ABox* $\mathcal{A}$,

- ans$(*, \mathcal{T}, \mathcal{A}) = $ cert$(*, \mathcal{F}_{\mathcal{T}}, \mathcal{A})$, *and*

- *if* ans$(*, \mathcal{T}, \mathcal{A}) = $ f, *then for each UCQ* $\mathcal{Q}$, *we have* ans$(\mathcal{Q}, \mathcal{T}, \mathcal{A}) = $ cert$(\mathcal{Q}, \mathcal{F}_{\mathcal{T}}, \mathcal{A})$.

*If* $\mathcal{F}_{\mathcal{T}}$ *contains predicates and/or individuals not occurring in* $\mathcal{T}$, *these are assumed to be 'internal' to* ans *and not accessible in queries, TBoxes, ABoxes, test suites, and so on. Given a TBox* $\mathcal{T}$, $\mathcal{C}_f^{\mathcal{T}}$ *is the class of all first-order reproducible abstract reasoners applicable to* $\mathcal{T}$.

**Example 3.34.** Abstract reasoners rdf, rdfs, rl and classify from Example 3.3 are all first-order reproducible. Indeed, theory $\mathcal{F}_{\mathcal{T}}$ is empty in the case of rdf, and it is precisely $\mathcal{P}_{\mathsf{rdfs}}$ and $\mathcal{P}_{\mathsf{rl}}$ in the cases of rdfs and rl, respectively. Finally, for abstract reasoner classify, theory $\mathcal{F}_{\mathcal{T}}$ is the union of $\mathcal{P}_{\mathsf{rl}}$ and the program containing the axiom $\forall x.[A(x) \to B(x)]$ for each atomic subsumption $A \sqsubseteq B$ entailed by the input TBox. $\diamond$

Please note that a first-order reproducible abstract reasoner ans does not need to actually construct $\mathcal{F}_{\mathcal{T}}$: it only matters that *some* (possibly unknown) theory $\mathcal{F}_{\mathcal{T}}$ exists that characterises the reasoner's behaviour as specified in Definition 3.33.

Since $\mathcal{Q} \cup \mathcal{F}_{\mathcal{T}} \cup \mathcal{A}' \models \mathcal{Q} \cup \mathcal{F}_{\mathcal{T}} \cup \mathcal{A}$ whenever $\mathcal{A} \subseteq \mathcal{A}'$, each first-order reproducible abstract reasoner is $(\mathcal{Q}, \mathcal{T})$-monotonic for arbitrary $\mathcal{Q}$ and $\mathcal{T}$. Furthermore, it is straightforward to see that each first-order reproducible abstract reasoner is also strongly $(\mathcal{Q}, \mathcal{T})$-faithful. Consequently, we have $\mathcal{C}_f^{\mathcal{T}} \subseteq \mathcal{C}_s^{\mathcal{Q}, \mathcal{T}}$ for each UCQ $\mathcal{Q}$ and each TBox $\mathcal{T}$.

We next show that the negative result in Theorem 3.21 does not directly apply to the class $\mathcal{C}_f^{\mathcal{T}}$. In particular, we show that the abstract reasoner pEval$_n$ used to prove Theorem 3.21 is not first-order reproducible. Intuitively, pEval$_n$ can be understood as *partial* evaluation of a datalog program—that is, the rules in the program are applied to the facts only a fixed number of times rather then until a fixpoint is reached.

**Proposition 3.35.** *For each positive integer* $n$, *the abstract reasoner* pEval$_n$ *defined in the proof of Theorem 3.21 is not first-order reproducible.*

*Proof.* Let $\mathcal{T} = \{\exists R.A \sqsubseteq A\}$, let $\mathcal{Q} = \{A(x) \to Q(x)\}$, and consider an arbitrary nonnegative integer $n$. Furthermore, assume that pEval$_n \in \mathcal{C}_f^{\mathcal{T}}$; then, a finite set of first-order sentences $\mathcal{F}_{\mathcal{T}}$ exists such that pEval$_n(\mathcal{Q}, \mathcal{T}, \mathcal{A}) = $ cert$(\mathcal{Q}, \mathcal{F}_{\mathcal{T}}, \mathcal{A})$ for each ABox $\mathcal{A}$.

Let $k$ be a positive integer; furthermore, let $r_k$ be the datalog rule and let $\mathcal{A}_k$ be the ABox defined as follows, for $a_0, \ldots, a_k$ arbitrary distinct but fixed individuals not occurring in $\mathcal{Q} \cup \mathcal{F}_{\mathcal{T}}$:

$$r_k = R(x_0, x_1) \wedge \ldots \wedge R(x_{k-1}, x_k) \wedge A(x_k) \to A(x_0)$$
$$\mathcal{A}_k = \{R(a_0, a_1), \ldots, R(a_{k-1}, a_k), A(a_k)\}$$

The following condition holds by Proposition 2.1:

$$\mathcal{F}_{\mathcal{T}} \models r_k \quad \text{if and only if} \quad \mathcal{F}_{\mathcal{T}} \cup \mathcal{A}_k \models A(a_0) \tag{17}$$

By the definition of $\mathsf{pEval}_n$, we have

$$a_0 \in \mathsf{pEval}_n(\mathcal{Q}, \mathcal{T}, \mathcal{A}_k) \quad \text{for each } 1 \leq k \leq n, \text{ and}$$
$$a_0 \notin \mathsf{pEval}_n(\mathcal{Q}, \mathcal{T}, \mathcal{A}_k) \quad \text{for each } k > n.$$

Since $\mathsf{pEval}_n(\mathcal{Q}, \mathcal{T}, \mathcal{A}) = \mathsf{cert}(\mathcal{Q}, \mathcal{F}_{\mathcal{T}}, \mathcal{A})$, we have

$$a_0 \in \mathsf{cert}(\mathcal{Q}, \mathcal{F}_{\mathcal{T}}, \mathcal{A}_k) \quad \text{for each } 1 \leq k \leq n, \text{ and}$$
$$a_0 \notin \mathsf{cert}(\mathcal{Q}, \mathcal{F}_{\mathcal{T}}, \mathcal{A}_k) \quad \text{for each } k > n.$$

Since $\mathcal{Q}$ contains only the atom $A(x)$ in the body, we have

$$\mathcal{F}_{\mathcal{T}} \cup \mathcal{A}_k \models A(a_0) \quad \text{for each } 1 \leq k \leq n, \text{ and}$$
$$\mathcal{F}_{\mathcal{T}} \cup \mathcal{A}_k \not\models A(a_0) \quad \text{for each } k > n.$$

By condition (17), we then have

$$\mathcal{F}_{\mathcal{T}} \models r_k \quad \text{for each } 1 \leq k \leq n$$
$$\mathcal{F}_{\mathcal{T}} \not\models r_k \quad \text{for each } k > n.$$

This, however, contradicts the obvious observation that $r_1 \models r_k$ for each $k \geq 1$. □

Note that the proof of Proposition 3.35 relies on the fact that the theory $\mathcal{F}_{\mathcal{T}}$ only depends on the input TBox, and not on the input query. As shown next, had we defined first-order reproducible abstract reasoners by allowing $\mathcal{F}_{\mathcal{T}}$ to depend also on the input query, then the negative result from Theorem 3.21 would have applied.

**Definition 3.36.** *An abstract reasoner* $\mathsf{ans}$ *for* $\mathcal{DL}$ *first-order q-reproducible if, for each UCQ* $\mathcal{Q}$ *and each* $\mathcal{DL}$-*TBox* $\mathcal{T}$, *a finite set of first-order sentences* $\mathcal{F}_{\mathcal{Q},\mathcal{T}}$ *exists such that, for each ABox* $\mathcal{A}$,

- $\mathsf{ans}(*, \mathcal{T}, \mathcal{A}) = \mathsf{cert}(*, \mathcal{F}_{\mathcal{Q},\mathcal{T}}, \mathcal{A})$, *and*

- *if* $\mathsf{ans}(*, \mathcal{T}, \mathcal{A}) = \mathsf{f}$, *then* $\mathsf{ans}(\mathcal{Q}, \mathcal{T}, \mathcal{A}) = \mathsf{cert}(\mathcal{Q}, \mathcal{F}_{\mathcal{Q},\mathcal{T}}, \mathcal{A})$.

**Theorem 3.37.** *For* $\mathcal{Q} = \{A(x) \rightarrow Q(x)\}$ *and* $\mathcal{T} = \{\exists R.A \sqsubseteq A\}$, *no* $\mathcal{T}$-*test suite exists that is exhaustive for* $\mathcal{Q}$ *and the class of all sound, monotonic, strongly faithful, and q-reproducible abstract reasoners applicable to* $\mathcal{T}$.

*Proof.* To prove this claim, it suffices to show that, for each nonnegative integer $n$, the abstract reasoner $\mathsf{pEval}_n$ defined in the proof of Theorem 3.21 is first-order q-reproducible. Consider an arbitrary nonnegative integer $n$, an arbitrary $\mathcal{DL}$-TBox $\mathcal{T}'$, and an arbitrary UCQ $\mathcal{Q}'$. We define $\mathcal{F}_{\mathcal{Q}',\mathcal{T}'}$ such that, if $\mathcal{T} \not\subseteq \mathcal{T}'$ or $\mathcal{Q}' \neq \mathcal{Q}$, then $\mathcal{F}_{\mathcal{Q}',\mathcal{T}'} = \emptyset$; otherwise, $\mathcal{F}_{\mathcal{Q}',\mathcal{T}'}$ consists of the following $n$ rules:

$$A(x_0) \rightarrow Q(x_0)$$
$$R(x_0, x_1) \wedge A(x_1) \rightarrow Q(x_0)$$
$$\cdots$$
$$R(x_0, x_1) \wedge R(x_1, x_2) \wedge \ldots \wedge R(x_{n-1}, x_n) \wedge A(x_n) \rightarrow Q(x_0)$$

Clearly, $\mathsf{pEval}_n(*, \mathcal{T}', \mathcal{A}') = \mathsf{cert}(*, \mathcal{F}_{\mathcal{Q}', \mathcal{T}'}, \mathcal{A}') = \mathsf{f}$ for each UCQ $\mathcal{Q}'$, $\mathcal{DL}$-TBox $\mathcal{T}'$ and ABox $\mathcal{A}'$, as required. Furthermore, for each $\mathcal{Q}'$ and $\mathcal{T}'$ such that either $\mathcal{T} \not\subseteq \mathcal{T}'$ or $\mathcal{Q}' \neq \mathcal{Q}$ and each ABox $\mathcal{A}'$, we have $\mathsf{pEval}_n(\mathcal{Q}', \mathcal{T}', \mathcal{A}') = \mathsf{cert}(\mathcal{Q}', \mathcal{F}_{\mathcal{Q}', \mathcal{T}'}, \mathcal{A}') = \emptyset$. Finally, for $\mathcal{Q}' = \mathcal{Q}$, each $\mathcal{T}'$ such that $\mathcal{T} \subseteq \mathcal{T}'$, and each ABox $\mathcal{A}'$, we clearly have $\mathsf{pEval}_n(\mathcal{Q}', \mathcal{T}', \mathcal{A}') = \mathsf{cert}(\mathcal{Q}', \mathcal{F}_{\mathcal{Q}', \mathcal{T}'}, \mathcal{A}')$, as required. □

### 3.7.2 Simple vs. Non-Simple Test Suites

Proposition 3.18 from Section 3.3 shows that each $\mathcal{Q}$-simple $\mathcal{T}$-test suite that is exhaustive for $\mathcal{Q}$ and a class of abstract reasoners provides a sufficient and necessary test for $(\mathcal{Q}, \mathcal{T})$-completeness. We next show that an analogous result does not hold if $\mathcal{T}$ contains recursive axioms, even if we consider only first-order reproducible abstract reasoners. As in Theorem 3.21, we prove the claim for a fixed $\mathcal{Q}$ and $\mathcal{T}$ since the concept of 'relevant recursive axioms' might be difficult to formalise; however, our proof can easily be adapted to other UCQs and TBoxes. Our result essentially states that no $\mathcal{T}$-test suite exists that provides a necessary and sufficient condition for $(\mathcal{Q}, \mathcal{T})$-completeness of each abstract reasoner in $\mathcal{C}_f^{\mathcal{T}}$; consequently, by Proposition 3.18 each $\mathcal{T}$-test suite exhaustive for $\mathcal{C}_f^{\mathcal{T}}$ and $\mathcal{Q}$ is not $\mathcal{Q}$-simple. Furthermore, in Section 3.7.3 we show how to compute a $\mathcal{T}$-test suite exhaustive for $\mathcal{C}_f^{\mathcal{T}}$ and $\mathcal{Q}$, so the following claim does not hold vacuously.

**Theorem 3.38.** *Let $\mathcal{Q} = \{A(x) \wedge B(x) \rightarrow Q(x)\}$, let $\mathcal{T} = \{\exists R.A \sqsubseteq A\}$, and let $\mathcal{C}$ be the class of all sound, monotonic, strongly faithful, and first-order reproducible abstract reasoners applicable to $\mathcal{T}$. Then, no $\mathcal{T}$-test suite $\mathbf{S}$ exists that satisfies the following two properties:*

  *1. $\mathbf{S}$ is exhaustive for $\mathcal{C}$ and $\mathcal{Q}$; and*

  *2. for each abstract reasoner $\mathsf{ans} \in \mathcal{C}$, if $\mathsf{ans}$ is $(\mathcal{Q}, \mathcal{T})$-complete then $\mathsf{ans}$ passes $\mathbf{S}$.*

*Proof.* Assume that a $\mathcal{T}$-test suite $\mathbf{S} = \langle \mathbf{S}_\perp, \mathbf{S}_Q \rangle$ exists that satisfies properties 1 and 2 of the theorem. Let $n$ be the maximal number of assertions occurring in an ABox in $\mathbf{S}$. We next define two abstract reasoners $\mathsf{ans}_1$ and $\mathsf{ans}_2$; it is straightforward to check that both are sound, monotonic, strongly faithful, and first-order reproducible.

Given an arbitrary $\mathcal{FOL}$-TBox $\mathcal{T}_{in}$, abstract reasoner $\mathsf{ans}_1$ uses the datalog program $\mathcal{F}_{\mathcal{T}_{in}}^1$ defined as follows:

- If $\mathcal{T} \not\subseteq \mathcal{T}_{in}$, then $\mathcal{F}_{\mathcal{T}_{in}}^1 = \emptyset$.

- If $\mathcal{T} \subseteq \mathcal{T}_{in}$, then $\mathcal{F}_{\mathcal{T}_{in}}^1$ contains the following $n$ rules:

$$
\begin{aligned}
r_0 &= & B(x_0) \wedge A(x_0) &\rightarrow A(x_0) \\
r_1 &= & B(x_0) \wedge R(x_0, x_1) \wedge A(x_1) &\rightarrow A(x_0) \\
r_2 &= & B(x_0) \wedge R(x_0, x_1) \wedge R(x_1, x_2) \wedge A(x_3) &\rightarrow A(x_0) \\
& & \cdots & \\
r_n &= & B(x_0) \wedge R(x_0, x_1) \wedge \ldots \wedge R(x_{n-1}, x_n) \wedge A(x_n) &\rightarrow A(x_0)
\end{aligned}
$$

Given an arbitrary $\mathcal{FOL}$-TBox $\mathcal{T}_{in}$, abstract reasoner $\mathsf{ans}_2$ uses the datalog program $\mathcal{F}_{\mathcal{T}_{in}}^2$ defined as follows, where predicate $Z$ is private to $\mathcal{F}_{\mathcal{T}_{in}}^2$ (and hence it does not affect the soundness of the abstract reasoner):

- If $\mathcal{T} \not\subseteq \mathcal{T}_{in}$, then $\mathcal{F}^2_{\mathcal{T}_{in}} = \emptyset$.

- If $\mathcal{T} \subseteq \mathcal{T}_{in}$, then $\mathcal{F}^2_{\mathcal{T}_{in}}$ contains $\mathcal{F}^1_{\mathcal{T}_{in}}$ as well as the following rules:

$$
\begin{aligned}
r_{Z_1} = \quad & R(x_0, x_1) \wedge \ldots \wedge R(x_n, x_{n+1}) \wedge A(x_{n+1}) \to Z(x_0) \\
r_{Z_2} = \quad & R(x_0, x_1) \wedge Z(x_1) \to Z(x_0) \\
r_{Z_3} = \quad & Z(x) \wedge B(x) \to A(x)
\end{aligned}
$$

Now let $\mathcal{A}$ be an arbitrary ABox containing at most $n$ assertions. We next show that, for each assertion $\alpha$ not containing predicate $Z$, we have $\mathcal{F}^1_{\mathcal{T}_{in}} \cup \mathcal{A} \models \alpha$ if and only if $\mathcal{F}^2_{\mathcal{T}_{in}} \cup \mathcal{A} \models \alpha$. The ($\Rightarrow$) direction is trivial since $\mathcal{F}^1_{\mathcal{T}_{in}} \subseteq \mathcal{F}^2_{\mathcal{T}_{in}}$, so we consider the ($\Leftarrow$) direction. Furthermore, since $r_{Z_3}$ is the only rule in $\mathcal{F}^2_{\mathcal{T}_{in}} \setminus \mathcal{F}^1_{\mathcal{T}_{in}}$ that does not contain $Z$ in the head, the claim is nontrivial only if $\alpha$ is of the form $A(a_0)$ for some individual $a_0$ occurring in $\mathcal{A}$. Since the antecedent of $r_{Z_3}$ is satisfied for $a_0$, we have $B(a_0) \in \mathcal{A}$ and $\mathcal{F}^2_{\mathcal{T}_{in}} \cup \mathcal{A} \models Z(a_0)$. But then, for the latter to be implied by $r_{Z_1}$ and $r_{Z_2}$, individuals $a_0, a_1, \ldots, a_k$ with $0 \le k$ exist such that $R(a_i, a_{i+1}) \in \mathcal{A}$ for each $1 \le i < k$, and $A(a_k) \in \mathcal{A}$. Since $\mathcal{A}$ contains at most $n$ assertions, w.l.o.g. we can assume that $k \le n$. But then, since $\mathcal{F}^1_{\mathcal{T}_{in}}$ contains rule $r_k$, we have $\mathcal{F}^1_{\mathcal{T}_{in}} \cup \mathcal{A} \models A(a_0)$ as well, which proves our claim. As a consequence of this claim and the fact that all ABoxes in $\mathbf{S}$ contain at most $n$ assertions, we have $\mathsf{cert}(*, \mathcal{F}^1_{\mathcal{T}_{in}}, \mathcal{A}) = \mathsf{cert}(*, \mathcal{F}^2_{\mathcal{T}_{in}}, \mathcal{A})$ for each $\mathcal{A} \in \mathbf{S}_\perp$, and $\mathsf{cert}(\mathcal{Y}, \mathcal{F}^1_{\mathcal{T}_{in}}, \mathcal{A}) = \mathsf{cert}(\mathcal{Y}, \mathcal{F}^2_{\mathcal{T}_{in}}, \mathcal{A})$ for each $\langle \mathcal{A}, \mathcal{Y} \rangle \in \mathbf{S}_Q$.

Let $\mathcal{A} = \{B(a_0), R(a_0, a_1), \ldots, R(a_n, a_{n+1}), A(a_{n+1})\}$. Then $\mathsf{cert}(\mathcal{Q}, \mathcal{T}, \mathcal{A}) = \{a_0\}$ and $\mathsf{cert}(\mathcal{Q}, \mathcal{F}^1_{\mathcal{T}_{in}}, \mathcal{A}) = \emptyset$, so $\mathsf{ans}_1$ is not $(\mathcal{Q}, \mathcal{T})$-complete. Since $\mathbf{S}$ is exhaustive for $\mathcal{C}$ and $\mathcal{Q}$, abstract reasoner $\mathsf{ans}_1$ does not pass $\mathbf{S}$; by the claim from the previous paragraph, abstract reasoner $\mathsf{ans}_2$ does not pass $\mathbf{S}$ either. We next show that $\mathsf{ans}_2$ is $(\mathcal{Q}, \mathcal{T})$-complete, which contradicts the assumption that $\mathbf{S}$ satisfies property 2 and thus proves the claim of this theorem.

Consider an arbitrary ABox $\mathcal{A}$ containing $m$ assertions. Clearly, $a_0 \in \mathsf{cert}(\mathcal{Q}, \mathcal{T}, \mathcal{A})$ if and only if individuals $a_0, a_1, \ldots, a_k$ with $0 \le k \le m$ exist such that $B(a_0) \in \mathcal{A}$, $R(a_i, a_{i+1}) \in \mathcal{A}$ for each $1 \le i < k$, and $A(a_k) \in \mathcal{A}$. Now assume that $k \le n$; since $r_k \in \mathcal{F}^2_{\mathcal{T}_{in}}$, we have $\mathcal{F}^2_{\mathcal{T}_{in}} \cup \mathcal{A} \models A(a_0)$ and thus $a_0 \in \mathsf{cert}(\mathcal{Q}, \mathcal{F}^2_{\mathcal{T}_{in}}, \mathcal{A})$. In contrast, assume that $k > n$; since $r_{Z_1} \in \mathcal{F}^2_{\mathcal{T}_{in}}$, we have $\mathcal{F}^2_{\mathcal{T}_{in}} \cup \mathcal{A} \models Z(a_{k-n-1})$; since $r_{Z_2} \in \mathcal{F}^2_{\mathcal{T}_{in}}$, we have $\mathcal{F}^2_{\mathcal{T}_{in}} \cup \mathcal{A} \models Z(a_i)$ for each $0 \le i \le k-n-1$; finally, since $r_{Z_3} \in \mathcal{F}^2_{\mathcal{T}_{in}}$, we have $\mathcal{F}^2_{\mathcal{T}_{in}} \cup \mathcal{A} \models A(a_0)$; but then, $a_0 \in \mathsf{cert}(\mathcal{Q}, \mathcal{F}^2_{\mathcal{T}_{in}}, \mathcal{A})$, as required. $\qquad\square$

As a corollary to Theorem 3.38, we next show that testing abstract reasoners in $\mathcal{C}^{\mathcal{T}}_f$ cannot be done in general using $\mathcal{Q}$-simple test suites.

**Corollary 3.39.** *For $\mathcal{Q} = \{A(x) \wedge B(x) \to Q(x)\}$ and $\mathcal{T} = \{\exists R.A \sqsubseteq A\}$, no $\mathcal{Q}$-simple $\mathcal{T}$-test suite exists that is exhaustive for $\mathcal{Q}$ and the class of all sound, monotonic, strongly faithful, and first-order reproducible abstract reasoners applicable to $\mathcal{T}$.*

*Proof.* If $\mathbf{S}$ is a $\mathcal{Q}$-simple $\mathcal{T}$-test suite that is exhaustive for $\mathcal{Q}$ and the class mentioned in the Theorem, by Proposition 3.18 each abstract reasoner $\mathsf{ans}$ from the class that does not pass $\mathbf{S}$ is not $(\mathcal{Q}, \mathcal{T})$-complete, which contradicts Theorem 3.38. $\qquad\square$

Theorem 3.38 effectively says that, if an abstract reasoner $\mathsf{ans} \in \mathcal{C}_f^\mathcal{T}$ does not pass a $\mathcal{T}$-test suite **S**, we cannot conclude that $\mathsf{ans}$ is not $(\mathcal{Q}, \mathcal{T})$-complete. Please note that this holds only if $\mathsf{ans}$ fails a test of the form $\langle \mathcal{A}, \mathcal{Y} \rangle$ where $\mathcal{Q} \neq \mathcal{Y}$: if $\mathcal{Q} = \mathcal{Y}$, then $\mathcal{A}$ is a counterexample to $(\mathcal{Q}, \mathcal{T})$-completeness of $\mathsf{ans}$. Thus, **S** may show $\mathsf{ans}$ to be not $(\mathcal{Q}, \mathcal{T})$-complete, but it is not guaranteed to do so. This is illustrated by the following example.

**Example 3.40.** Let $\mathcal{Q} = \{A(x) \wedge B(x) \to Q(x)\}$ and let $\mathcal{T} = \{\exists R.A \sqsubseteq A, \exists R.C \sqsubseteq C\}$. Furthermore, let $\mathbf{S} = \langle \emptyset, \mathbf{S}_Q \rangle$ be the general test suite defined as follows:

$$\mathbf{S}_Q = \{ \quad \langle \, \{\, A(c)\,\}, \qquad\quad \{\, A(x) \wedge B(x) \to Q(x)\,\} \,\rangle,$$
$$\langle \, \{\, R(c,d), A(d)\,\}, \,\{\, A(c) \to Q'\,\} \,\rangle,$$
$$\langle \, \{\, R(c,d), C(d)\,\}, \,\{\, C(c) \to Q'\,\} \,\rangle \qquad\qquad \}$$

Let $\mathcal{R} = \langle \mathcal{R}_D, \emptyset, \mathcal{Q} \rangle$ where $\mathcal{R}_D = \{R(x,y) \wedge A(y) \to A(x), \ R(x,y) \wedge C(y) \to C(x)\}$; clearly, $\mathcal{R}$ is a rewriting $\mathcal{Q}$ w.r.t. $\mathcal{T}$. In Section 3.7.3 we show how to compute **S** from $\mathcal{R}$ using a variant of injective instantiation in a way that guarantees exhaustiveness for $\mathcal{C}_f^\mathcal{T}$ and $\mathcal{Q}$.

Now let $\mathsf{ans}_1 \in \mathcal{C}_f^\mathcal{T}$ be the abstract reasoner defined by $\mathcal{F}_\mathcal{T}^1 = \{R(x,y) \wedge A(y) \to A(x)\}$. The reasoner does not pass **S** since $\mathsf{cert}(\{C(c) \to Q'\}, \mathcal{F}_\mathcal{T}^1, \{R(c,d), C(d)\}) = \mathsf{f}$. Note, however, that the reasoner is $(\mathcal{Q}, \mathcal{T})$-complete. Thus, if a test suite is not $\mathcal{Q}$-simple, passing it is a sufficient, but not a necessary condition for $(\mathcal{Q}, \mathcal{T})$-completeness. In fact, note that $\mathcal{T}$ contains the TBox for Theorem 3.38, so by the theorem we cannot 'reduce' **S** so that it correctly identifies all reasoners in $\mathcal{C}_f^\mathcal{T}$ that are not $(\mathcal{Q}, \mathcal{T})$-complete.

In practice, however, one can try to mitigate this fundamental theoretical limitation by eliminating the irrelevant axioms from the rewriting $\mathcal{R}$ and thus increasing the likelihood of obtaining a $\mathcal{T}$-test suite that a $(\mathcal{Q}, \mathcal{T})$-complete abstract reasoner will pass. For example, using the techniques by Cuenca Grau, Horrocks, Kazakov, and Sattler (2008a) we can extract the module of $\mathcal{R}$ relevant to the query. In the example from the previous paragraph, this would remove the rule $R(x,y) \wedge C(y) \to C(x)$ from $\mathcal{R}$, and injective instantiation will produce the test suite $\mathbf{S}' = \langle \emptyset, \mathbf{S}'_Q \rangle$ where $\mathbf{S}'_Q$ is defined as follows:

$$\mathbf{S}'_Q = \{ \quad \langle \, \{\, A(c)\,\}, \qquad\qquad \{\, A(x) \wedge B(x) \to Q(x)\,\} \,\rangle,$$
$$\langle \, \{\, R(c,d), A(d)\,\}, \,\{\, A(c) \to Q'\,\} \,\rangle \qquad\qquad \}$$

Abstract reasoner $\mathsf{ans}_1$ from the previous paragraph now passes $\mathbf{S}'$ and is thus guaranteed to be $(\mathcal{Q}, \mathcal{T})$-complete.

Now let $\mathsf{ans}_2$ be the abstract reasoner defined by $\mathcal{F}_\mathcal{T}^2 = \{B(x) \wedge R(x,y) \wedge A(y) \to A(x)\}$. Clearly, abstract reasoner $\mathsf{ans}_2$ is not $(\mathcal{Q}, \mathcal{T})$-complete, so $\mathsf{ans}_2$ does not pass $\mathbf{S}'_Q$. From the latter, however, we cannot immediately conclude that $\mathbf{S}'$ is not $(\mathcal{Q}, \mathcal{T})$-complete: the test that fails does not involve the original query $\mathcal{Q}$. As a possible remedy, we can try to unfold $\mathcal{R}$ to a certain level and then injectively instantiate the result in hope of obtaining a $\mathcal{T}$-test suite that will identify $\mathsf{ans}_2$ as not being $(\mathcal{Q}, \mathcal{T})$-complete. In particular, the first unfolding of $\mathcal{R}$ produces the following query:

$$B(x) \wedge R(x,y) \wedge A(y) \to Q(x)$$

Instantiating this rewriting produces the following test suite, which does not prove that $\mathsf{ans}_2$ is not $(\mathcal{Q}, \mathcal{T})$-complete.

$$\mathbf{S}''_Q = \{ \quad \langle \, \{\, B(c), R(c,d), A(d)\,\}, \,\{\, A(x) \wedge B(x) \to Q(x)\,\} \,\rangle \quad \}$$

Another round of unfolding, however, produces the following query:

$$B(x) \wedge R(x, y) \wedge R(y, z) \wedge A(z) \rightarrow Q(x)$$

Instantiating this query produces the following test suite:

$$\mathbf{S}_Q''' = \{ \quad \langle \{ B(c), R(c, d), R(d, e), A(e) \}, \{ A(x) \wedge B(x) \rightarrow Q(x) \} \rangle \quad \}$$

Now $\mathsf{ans}_2$ does not pass $\mathbf{S}_Q'''$, so we can conclude that $\mathsf{ans}_2$ is not $(\mathcal{Q}, \mathcal{T})$-complete. $\diamond$

To better understand Example 3.40, consider a first-order reproducible abstract reasoner $\mathsf{ans}$, an arbitrary UCQ $\mathcal{Q}$, and a TBox $\mathcal{T}$ such that $\mathcal{R} = \langle \mathcal{R}_D, \emptyset, \mathcal{R}_Q \rangle$ is a datalog rewriting of $\mathcal{Q}$ w.r.t. $\mathcal{T}$. Datalog program $\mathcal{R}_D \cup \mathcal{R}_Q$ is equivalent to the (possibly infinite) UCQ $\mathcal{R}_Q^u$ obtained from $\mathcal{R}_D \cup \mathcal{R}_Q$ via exhaustive unfolding. We now have the following possibilities.

First, assume that $\mathsf{ans}$ is not $(\mathcal{Q}, \mathcal{T})$-complete. Since $\mathcal{R}_D \cup \mathcal{R}_Q$ is equivalent to $\mathcal{R}_Q^u$, each certain answer $\vec{a}$ to $\mathcal{Q}$ w.r.t. $\mathcal{T}$ and an arbitrary ABox $\mathcal{A}$ is 'produced' by some $r \in \mathcal{R}_Q^u$. But then, the injective instantiation $\mathcal{A}_\lambda^r$ of $r$ will provide us with the counterexample for the $(\mathcal{Q}, \mathcal{T})$-completeness of $\mathsf{ans}$. Thus, we can prove that $\mathsf{ans}$ is not $(\mathcal{Q}, \mathcal{T})$-complete by generating the elements of $\mathcal{R}_Q^u$ in a fair manner (i.e., without indefinitely delaying the generation of some element of $\mathcal{R}_Q^u$) and checking whether $\mathsf{cert}(\mathcal{Q}, \mathcal{T}, \mathcal{A}_\lambda^r) \subseteq \mathsf{ans}(\mathcal{Q}, \mathcal{T}, \mathcal{A}_\lambda^r)$; we are guaranteed to eventually encounter some $r \in \mathcal{R}_Q^u$ that invalidates this condition and thus proves that $\mathsf{ans}$ is not $(\mathcal{Q}, \mathcal{T})$-complete.

Second, assume that $\mathsf{ans}$ is $(\mathcal{Q}, \mathcal{T})$-complete. Using the above approach, we will determine that $\mathsf{cert}(\mathcal{Q}, \mathcal{T}, \mathcal{A}_\lambda^r) \subseteq \mathsf{ans}(\mathcal{Q}, \mathcal{T}, \mathcal{A}_\lambda^r)$ holds for each $r \in \mathcal{R}_Q^u$. Now if $\mathcal{R}_Q^u$ is finite (i.e., if the unfolding of $\mathcal{R}_D \cup \mathcal{R}_Q$ terminates), then $\mathcal{R}_Q^u$ is the UCQ rewriting of $\mathcal{Q}$ w.r.t. $\mathcal{T}$, so by the results from Section 3.6 we can conclude that $\mathsf{ans}$ is indeed $(\mathcal{Q}, \mathcal{T})$-complete. If, however, $\mathcal{R}_Q^u$ is infinite, then we will never obtain a sufficient assurance for the $(\mathcal{Q}, \mathcal{T})$-complete of $\mathsf{ans}$. In the following section we show a possible remedy to this problem.

### 3.7.3 TESTING FIRST-ORDER REPRODUCIBLE ABSTRACT REASONERS

In this section, we show how to compute a $\mathcal{T}$-test suite $\mathbf{S} = \langle \mathbf{S}_\perp, \mathbf{S}_Q \rangle$ exhaustive for $\mathcal{C}_f^{\mathcal{T}}$ and $\mathcal{Q}$ from a datalog$^{\pm, \vee}$ rewriting $\mathcal{R} = \langle \mathcal{R}_D, \mathcal{R}_\perp, \mathcal{R}_Q \rangle$ of $\mathcal{Q}$ w.r.t. $\mathcal{T}$. Since first-order reproducible abstract reasoners are strongly faithful, we need to consider only injective instantiations of $\mathcal{R}$. Thus, the rules in $\mathcal{R}_\perp$ and $\mathcal{R}_Q$ should be instantiated as in Section 3.6. A rule $r \in \mathcal{R}_D$, however, is instantiated into a pair $\langle \mathcal{A}, \mathcal{Y} \rangle \in \mathbf{S}_Q$ with $\mathcal{A}$ the ABox obtained by instantiating the body of $r$ and $\mathcal{Y}$ the Boolean UCQ obtained by instantiating the head of $r$. Intuitively, such tests allow us to check whether the (unknown) first-order theory $\mathcal{F}_\mathcal{T}$ that captures the behaviour of the abstract reasoner entails $r$.

**Definition 3.41.** *Let $\mathcal{Q}$ be a UCQ with query predicate $Q$, let $\mathcal{T}$ be an admissible TBox, let $\mathcal{R} = \langle \mathcal{R}_D, \mathcal{R}_\perp, \mathcal{R}_Q \rangle$ be a datalog$^{\pm, \vee}$ rewriting of $\mathcal{Q}$ w.r.t. $\mathcal{T}$, and let $\lambda$ be a substitution mapping each variable occurring in $\mathcal{R}$ into a distinct fresh individual. The* injective instantiation *of $\mathcal{R}$ w.r.t. $\lambda$ is the pair $\mathbf{I}^{\mathcal{R}, \lambda} = \langle \mathbf{I}_\perp^{\mathcal{R}, \lambda}, \mathbf{I}_Q^{\mathcal{R}, \lambda} \rangle$ where $\mathbf{I}_\perp^{\mathcal{R}, \lambda}$ is the smallest set of ABoxes and $\mathbf{I}_Q^{\mathcal{R}, \lambda}$ is the smallest set of pairs of an ABox and a UCQ such that*

- $\mathcal{A}_\lambda^r \in \mathbf{I}_\perp^{\mathcal{R}, \lambda}$ *for each $r \in \mathcal{R}_\perp$,*

- $\langle \mathcal{A}_\lambda^r, \mathcal{Q} \rangle \in \mathbf{I}_Q^{\mathcal{R},\lambda}$ for each $r \in \mathcal{R}_Q$ such that $\mathsf{cert}(*, \mathcal{R}_D \cup \mathcal{R}_\perp, \mathcal{A}_\lambda^r) = \mathsf{f}$, and

- $\langle \mathcal{A}_\lambda^r, \mathcal{Y} \rangle \in \mathbf{I}_Q^{\mathcal{R},\lambda}$ for each $r \in \mathcal{R}_D$ of the form (6) such that $\mathsf{cert}(*, \mathcal{R}_D \cup \mathcal{R}_\perp, \mathcal{A}_\lambda^r) = \mathsf{f}$, where $\mathcal{Y}$ is the UCQ $\mathcal{Y} = \{\varphi_i(\lambda(\vec{x}), \vec{y}_i) \to Q' \mid 1 \le i \le m\}$ with the propositional query predicate $Q'$.

$\mathbf{I}^{\mathcal{R},\lambda}$ is clearly unique up to the renaming of the fresh individuals in $\lambda$, so $\lambda$ is typically left implicit, and one talks of the injective instantiation $\mathbf{I}^{\mathcal{R}} = \langle \mathbf{I}_\perp^{\mathcal{R}}, \mathbf{I}_Q^{\mathcal{R}} \rangle$ of $\mathcal{R}$.

**Example 3.42.** Consider the query $\mathcal{Q} = \{A(x) \to Q(x)\}$ and the $\mathcal{EL}$-TBox $\mathcal{T}$ consisting of the following axioms, whose translation into first-order logic is shown after the $\rightsquigarrow$ symbol.

$$\begin{aligned}
\exists R.A \sqsubseteq B \quad &\rightsquigarrow \quad \forall x, y.[R(x,y) \wedge A(y) \to B(x)] \\
\exists R.C \sqsubseteq A \quad &\rightsquigarrow \quad \forall x, y.[R(x,y) \wedge C(y) \to A(x)] \\
B \sqsubseteq C \quad &\rightsquigarrow \quad \forall x.[B(x) \to C(x)] \\
C \sqsubseteq \exists R.D \quad &\rightsquigarrow \quad \forall x.[C(x) \to \exists y.[R(x,y) \wedge D(y)]] \\
A \sqcap D \sqsubseteq \perp \quad &\rightsquigarrow \quad \forall x.[A(x) \wedge D(x) \to \perp]
\end{aligned}$$

Then, $\mathcal{R} = \langle \mathcal{R}_D, \mathcal{R}_\perp, \mathcal{R}_Q \rangle$ as defined next is a datalog rewriting of $\mathcal{Q}$ w.r.t. $\mathcal{T}$.

$$\begin{aligned}
\mathcal{R}_D &= \{\ R(x,y) \wedge A(y) \to B(x),\ R(x,y) \wedge C(y) \to A(x),\ B(x) \to C(x)\ \} \\
\mathcal{R}_\perp &= \{\ A(x) \wedge D(x) \to \perp\ \} \\
\mathcal{R}_Q &= \{\ A(x) \to Q(x)\ \}
\end{aligned}$$

The injective instantiation $\mathbf{I}^{\mathcal{R}} = \langle \mathbf{I}_\perp^{\mathcal{R}}, \mathbf{I}_Q^{\mathcal{R}} \rangle$ of $\mathcal{R}$ is shown below.

$$\begin{aligned}
\mathbf{I}_\perp^{\mathcal{R}} = \{\ &\{A(c), D(c)\}\ \} \\
\mathbf{I}_Q^{\mathcal{R}} = \{\ &\langle\ \{A(c)\}, & \{A(x) \to Q(x)\}\ \rangle, \\
&\langle\ \{R(c,d), A(d)\}, \{B(c) \to Q'\}\ \rangle, \\
&\langle\ \{R(c,d), C(d)\}, \{A(c) \to Q'\}\ \rangle, \\
&\langle\ \{B(c)\}, & \{C(c) \to Q'\}\ \rangle \quad\ \}
\end{aligned}$$
$\diamond$

We now show that the injective instantiation of a datalog$^{\pm,\vee}$ rewriting of $\mathcal{Q}$ w.r.t. $\mathcal{T}$ is a $\mathcal{T}$-test suite exhaustive for $\mathcal{C}_f^{\mathcal{T}}$ and $\mathcal{Q}$.

**Theorem 3.43.** Let $\mathcal{Q}$ be a UCQ, let $\mathcal{T}$ be a TBox, let $\mathcal{R} = \langle \mathcal{R}_D, \mathcal{R}_\perp, \mathcal{R}_Q \rangle$ be a datalog$^{\pm,\vee}$ rewriting of $\mathcal{Q}$ w.r.t. $\mathcal{T}$, and let $\mathbf{I}^{\mathcal{R}} = \langle \mathbf{I}_\perp^{\mathcal{R}}, \mathbf{I}_Q^{\mathcal{R}} \rangle$ be the injective instantiation of $\mathcal{R}$. Then, $\mathbf{I}^{\mathcal{R}}$ is a $\mathcal{T}$-test suite that is exhaustive for $\mathcal{C}_f^{\mathcal{T}}$ and $\mathcal{Q}$.

*Proof.* Let $\lambda$ be the substitution that $\mathbf{I}^{\mathcal{R}}$ is obtained from. We first show that $\mathbf{I}^{\mathcal{R}}$ is a $\mathcal{T}$-test suite.

- Consider an arbitrary $\mathcal{A} \in \mathbf{I}_\perp^{\mathcal{R}}$. Then, a rule $r \in \mathcal{R}_\perp$ exist such that $\mathcal{A} = \mathcal{A}_\lambda^r$; clearly $\mathsf{cert}(*, \{r\}, \mathcal{A}) = \mathsf{t}$, so $\mathsf{cert}(*, \mathcal{R}_D \cup \mathcal{R}_\perp, \mathcal{A}) = \mathsf{t}$ as well; since $\mathcal{R}$ is a datalog$^{\pm,\vee}$ rewriting of $\mathcal{Q}$ w.r.t. $\mathcal{T}$, we have that $\mathcal{T} \cup \mathcal{A}$ is unsatisfiable, as required.

- Consider an arbitrary $\mathcal{A} \in \mathbf{I}_Q^{\mathcal{R}}$. Then, $\mathsf{cert}(*, \mathcal{R}_D \cup \mathcal{R}_\perp, \mathcal{A}) = \mathsf{f}$ by Definition 3.41; since $\mathcal{R}$ is a $\mathsf{datalog}^{\pm,\vee}$ rewriting of $\mathcal{Q}$ w.r.t. $\mathcal{T}$, we have that $\mathcal{T} \cup \mathcal{A}$ is satisfiable, as required.

To show that $\mathbf{I}^{\mathcal{R}}$ is exhaustive for $\mathcal{C}_f^{\mathcal{T}}$ and $\mathcal{Q}$, consider an arbitrary abstract reasoner $\mathsf{ans} \in \mathcal{C}_f^{\mathcal{T}}$ that passes $\mathbf{I}^{\mathcal{R}}$—that is, $\mathsf{ans}$ satisfies the following two properties:

(a) $\mathsf{ans}(*, \mathcal{T}, \mathcal{A}') = \mathsf{t}$ for each $\mathcal{A}' \in \mathbf{I}_\perp^{\mathcal{R}}$, and

(b) $\mathsf{ans}(*, \mathcal{T}, \mathcal{A}') = \mathsf{f}$ implies $\mathsf{cert}(\mathcal{Y}, \mathcal{T}, \mathcal{A}') \subseteq \mathsf{ans}(\mathcal{Y}, \mathcal{T}, \mathcal{A}')$ for each $\langle \mathcal{Y}, \mathcal{A}' \rangle \in \mathbf{I}_Q^{\mathcal{R}}$.

Since $\mathsf{ans}$ is first-order reproducible, a set of first-order sentences $\mathcal{F}_\mathcal{T}$ exists such that, for each ABox $\mathcal{A}$, we have

- $\mathsf{ans}(*, \mathcal{T}, \mathcal{A}) = \mathsf{cert}(*, \mathcal{F}_\mathcal{T}, \mathcal{A})$, and

- if $\mathsf{ans}(*, \mathcal{T}, \mathcal{A}) = \mathsf{f}$, then $\mathsf{ans}(\mathcal{Q}, \mathcal{T}, \mathcal{A}) = \mathsf{cert}(\mathcal{Q}, \mathcal{F}_\mathcal{T}, \mathcal{A})$.

By the assumption on $\mathcal{F}_\mathcal{T}$ in Definition 3.33 and the fact that $\lambda$ maps variables to fresh individuals, we have $\mathsf{rng}(\lambda) \cap \mathsf{ind}(\mathcal{F}_\mathcal{T}) = \emptyset$.

Let $\mathcal{R}_D^1$ and $\mathcal{R}_D^2$ be the smallest sets of rules satisfying the following conditions for each rule $r \in \mathcal{R}_D$:

- $\mathsf{cert}(*, \mathcal{F}_\mathcal{T}, \mathcal{A}_\lambda^r) = \mathsf{t}$ implies $r' \in \mathcal{R}_D^1$, where $r'$ is obtained from $r$ by replacing the head with $\perp$, and

- $\mathsf{cert}(*, \mathcal{F}_\mathcal{T}, \mathcal{A}_\lambda^r) = \mathsf{f}$ implies $r \in \mathcal{R}_D^2$.

Furthermore, let $\mathcal{R}_Q^1$ and $\mathcal{R}_Q^2$ be the sets of rules obtained from $\mathcal{R}_Q$ in an analogous way. Since $\mathcal{R}_D^1 \cup \mathcal{R}_D^2$ is obtained from $\mathcal{R}_D$ by replacing some head formulae with $\perp$, we clearly have $\mathcal{R}_D^1 \cup \mathcal{R}_D^2 \models \mathcal{R}_D$; analogously, we have $\mathcal{R}_Q^1 \cup \mathcal{R}_Q^2 \models \mathcal{R}_Q$.

We next show that $\mathcal{F}_\mathcal{T} \models \mathcal{R}_\perp$; the latter holds if and only if $\mathcal{F}_\mathcal{T} \models r$ for each rule $r \in \mathcal{R}_\perp$. Consider an arbitrary rule $r \in \mathcal{R}_\perp$; note that $\mathsf{head}(r) = \perp$. Then, by Definition 3.41 we have $\mathcal{A}_\lambda^r \in \mathbf{I}_\perp^{\mathcal{R}}$; by (a) we have $\mathsf{ans}(*, \mathcal{T}, \mathcal{A}_\lambda^r) = \mathsf{t}$; by Definition 3.33 we have $\mathsf{cert}(*, \mathcal{F}_\mathcal{T}, \mathcal{A}_\lambda^r) = \mathsf{t}$ and hence $\mathcal{F}_\mathcal{T} \cup \mathcal{A}_\lambda^r \models \perp$; finally, since $\mathsf{rng}(\lambda) \cap \mathsf{ind}(\mathcal{F}_\mathcal{T}) = \emptyset$, by Proposition 2.1 we have $\mathcal{F}_\mathcal{T} \models r$, as required.

We next show that $\mathcal{F}_\mathcal{T} \models \mathcal{R}_D^1$; the latter holds if and only if $\mathcal{F}_\mathcal{T} \models r$ for each rule $r \in \mathcal{R}_D^1$. Consider an arbitrary rule $r \in \mathcal{R}_D^1$; note that $\mathsf{head}(r) = \perp$. Then, by the definition of $\mathcal{R}_D^1$ we have $\mathsf{cert}(*, \mathcal{F}_\mathcal{T}, \mathcal{A}_\lambda^r) = \mathsf{t}$ and hence $\mathcal{F}_\mathcal{T} \cup \mathcal{A}_\lambda^r \models \perp$; finally, since $\mathsf{rng}(\lambda) \cap \mathsf{ind}(\mathcal{F}_\mathcal{T}) = \emptyset$, by Proposition 2.1 we have $\mathcal{F}_\mathcal{T} \models r$, as required.

In a completely analogous way as in the previous paragraph, it is possible to show that $\mathcal{F}_\mathcal{T} \models \mathcal{R}_Q^1$.

We next show that $\mathcal{F}_\mathcal{T} \models \mathcal{R}_D^2$; the latter holds if and only if $\mathcal{F}_\mathcal{T} \models r$ for each rule $r \in \mathcal{R}_D^2$. Consider an arbitrary rule $r \in \mathcal{R}_D^2$ of the form (6); by the definition of $\mathcal{R}_D^2$ we have $\mathsf{cert}(*, \mathcal{F}_\mathcal{T}, \mathcal{A}_\lambda^r) = \mathsf{f}$, so by Definition 3.33 we have $\mathsf{ans}(*, \mathcal{T}, \mathcal{A}_\lambda^r) = \mathsf{f}$. Then, by Definition 3.41 we have $\langle \mathcal{A}_\lambda^r, \mathcal{Y} \rangle \in \mathbf{I}_Q^{\mathcal{R}}$ where $\mathcal{Y}$ is the UCQ $\mathcal{Y} = \{\varphi_i(\lambda(\vec{x}), \vec{y_i}) \to Q' \mid 1 \leq i \leq m\}$. Note that $\mathcal{T} \models r$ by Definition 2.2, so by Proposition 2.1 we have that $\mathcal{T} \cup \mathcal{A}_\lambda^r \models \bigvee_{i=1}^m \varphi_i(\lambda(\vec{x}), \vec{y_i})$; by the definition of $\mathcal{Y}$ and the fact that $Q'$ does not occur in $\mathcal{T}$, we have $\mathcal{Y} \cup \mathcal{T} \cup \mathcal{A}_\lambda^r \models Q'$;

but then, $\mathsf{cert}(\mathcal{Y}, \mathcal{T}, \mathcal{A}_\lambda^r) = \mathsf{t}$. The latter observation, $\mathsf{ans}(*, \mathcal{T}, \mathcal{A}_\lambda^r) = \mathsf{f}$, and (b) then imply $\mathsf{ans}(\mathcal{Y}, \mathcal{T}, \mathcal{A}_\lambda^r) = \mathsf{t}$, so by Definition 3.33 we have $\mathsf{cert}(\mathcal{Y}, \mathcal{F}_\mathcal{T}, \mathcal{A}_\lambda^r) = \mathsf{t}$. Since $Q'$ occurs only in $\mathcal{Y}$ (note that each predicate occurring in $\mathcal{F}_\mathcal{T}$ but not in $\mathcal{T}$ is private to $\mathcal{F}_\mathcal{T}$, so $Q'$ cannot occur in $\mathcal{F}_\mathcal{T}$), we have $\mathcal{F}_\mathcal{T} \cup \mathcal{A}_\lambda^r \models \bigvee_{i=1}^m \varphi_i(\lambda(\vec{x}), \vec{y}_i)$. Finally, since $\mathsf{rng}(\lambda) \cap \mathsf{ind}(\mathcal{F}_\mathcal{T}) = \emptyset$, by Proposition 2.1 we have $\mathcal{F}_\mathcal{T} \models r$, as required.

We next show that $\mathcal{Q} \cup \mathcal{F}_\mathcal{T} \models \mathcal{R}_Q^2$; the latter holds if and only if $\mathcal{Q} \cup \mathcal{F}_\mathcal{T} \models r$ for each rule $r \in \mathcal{R}_Q^2$. Consider an arbitrary rule $r \in \mathcal{R}_Q^2$; note that $\mathsf{head}(r)$ is an atom with predicate $Q$, and that by the definition of $\mathcal{R}_Q^2$ we have $\mathsf{cert}(*, \mathcal{F}_\mathcal{T}, \mathcal{A}_\lambda^r) = \mathsf{f}$, so by Definition 3.33 we have $\mathsf{ans}(*, \mathcal{T}, \mathcal{A}_\lambda^r) = \mathsf{f}$. Furthermore, by Definition 3.41, we have $\mathsf{cert}(*, \mathcal{R}_D \cup \mathcal{R}_\perp, \mathcal{A}_\lambda^r) = \mathsf{f}$. Let $\vec{a}$ be the tuple of the arguments in $\lambda(\mathsf{head}(r))$. Then, by Definition 3.41 we have $\langle \mathcal{A}_\lambda^r, \mathcal{Q} \rangle \in \mathbf{I}_Q^\mathcal{R}$; clearly, $\vec{a} \in \mathsf{cert}(\{r\}, \emptyset, \mathcal{A}_\lambda^r)$, but then we have $\vec{a} \in \mathsf{cert}(\mathcal{R}_Q, \mathcal{R}_D \cup \mathcal{R}_\perp, \mathcal{A}_\lambda^r)$ by the monotonicity of first-order logic. Since $\mathcal{R}$ is a rewriting of $\mathcal{Q}$ w.r.t. $\mathcal{T}$, by Definition 2.2 we have $\vec{a} \in \mathsf{cert}(\mathcal{Q}, \mathcal{T}, \mathcal{A}_\lambda^r)$. The latter observation, $\mathsf{ans}(*, \mathcal{T}, \mathcal{A}_\lambda^r) = \mathsf{f}$, and (b) then imply $\vec{a} \in \mathsf{ans}(\mathcal{Q}, \mathcal{T}, \mathcal{A}_\lambda^r)$. By Definition 3.33 we have $\vec{a} \in \mathsf{cert}(\mathcal{Q}, \mathcal{F}_\mathcal{T}, \mathcal{A}_\lambda^r)$; hence, $\mathcal{F}_\mathcal{T} \cup \mathcal{A}_\lambda^r \models Q(\vec{a})$. Finally, since we have $\mathsf{rng}(\lambda) \cap \mathsf{ind}(\mathcal{F}_\mathcal{T}) = \emptyset$, by Proposition 2.1 we have $\mathcal{F}_\mathcal{T} \models r$, as required.

The following table summarises the entailment relationships between various first-order theories obtained thus far:

$$\mathcal{F}_\mathcal{T} \models \mathcal{R}_\perp \qquad \mathcal{F}_\mathcal{T} \models \mathcal{R}_D^1 \qquad \mathcal{F}_\mathcal{T} \models \mathcal{R}_D^2 \qquad \mathcal{F}_\mathcal{T} \models \mathcal{R}_Q^1$$
$$\mathcal{Q} \cup \mathcal{F}_\mathcal{T} \models \mathcal{R}_Q^2 \qquad \mathcal{R}_Q^1 \cup \mathcal{R}_Q^2 \models \mathcal{R}_Q \qquad \mathcal{R}_D^1 \cup \mathcal{R}_D^2 \models \mathcal{R}_D$$

Clearly, this implies the following entailments:

$$\mathcal{F}_\mathcal{T} \models \mathcal{R}_D \cup \mathcal{R}_\perp \qquad \mathcal{Q} \cup \mathcal{F}_\mathcal{T} \models \mathcal{R}_D \cup \mathcal{R}_\perp \cup \mathcal{R}_Q$$

We now complete the proof of this theorem and show that $\mathsf{ans}$ is $(\mathcal{Q}, \mathcal{T})$-complete. To this end, consider an arbitrary ABox $\mathcal{A}$; we have the following possibilities, depending on the satisfiability of $\mathcal{T} \cup \mathcal{A}$.

- Assume that $\mathcal{T} \cup \mathcal{A}$ is unsatisfiable. Then $\mathsf{cert}(*, \mathcal{R}_D \cup \mathcal{R}_\perp, \mathcal{A}) = \mathsf{t}$ by Definition 2.2; by the above mentioned entailments, we have $\mathsf{cert}(*, \mathcal{F}_\mathcal{T}, \mathcal{A}) = \mathsf{t}$; consequently, $\mathsf{ans}(*, \mathcal{T}, \mathcal{A}) = \mathsf{t}$ by Definition 3.33, as required.

- Assume that $\mathcal{T} \cup \mathcal{A}$ is satisfiable and $\mathsf{ans}(*, \mathcal{T}, \mathcal{A}) = \mathsf{f}$, and consider an arbitrary tuple $\vec{a} \in \mathsf{cert}(\mathcal{Q}, \mathcal{T}, \mathcal{A})$. Then, $\mathsf{cert}(*, \mathcal{R}_D \cup \mathcal{R}_\perp, \mathcal{A}) = \mathsf{f}$ and $\vec{a} \in \mathsf{cert}(\mathcal{R}_Q, \mathcal{R}_D \cup \mathcal{R}_\perp, \mathcal{A})$ by Definition 2.2. By the above mentioned entailments, we have $\vec{a} \in \mathsf{cert}(\mathcal{Q}, \mathcal{F}_\mathcal{T}, \mathcal{A})$; hence, $\vec{a} \in \mathsf{ans}(\mathcal{Q}, \mathcal{T}, \mathcal{A})$ by Definition 3.33, as required. □

Note that the size of the test suite obtained by Theorem 3.43 is linear in the size of the rewriting, which, we believe, makes our approach suitable for use in practice.

### 3.7.4 Testing Ground Queries

As shown in Section 3.7.2, if an abstract reasoner $\mathsf{ans} \in \mathcal{C}_f^\mathcal{T}$ does not pass a $\mathcal{T}$-test suite $\mathbf{S}$ that is not $\mathcal{Q}$-simple, we cannot always conclude that $\mathsf{ans}$ is not $(\mathcal{Q}, \mathcal{T})$-complete. From a practical point of view, it would be highly beneficial to identify situations where not passing $\mathbf{S}$ would show that $\mathsf{ans}$ is indeed incomplete for $\mathcal{Q}$ and $\mathcal{T}$. Furthermore, in applications where prototypical queries are not known at design time, we would like to design completeness tests that are query-independent—that is, which test an abstract reasoner for completeness w.r.t. $\mathcal{T}$ regardless of the input data and query. In this section, we show that we can achieve these two goals by focusing on ground queries. This restriction is not unreasonable in practice, since any SPARQL query can be equivalently expressed as a ground UCQ.

We first define a query-independent notion of exhaustiveness of a test suite.

**Definition 3.44.** *Let $\mathcal{T}$ be a TBox, let $\mathbf{S}$ be a $\mathcal{T}$-test suite, and let $\mathcal{C}$ be a class of abstract reasoners applicable to $\mathcal{T}$. Then, $\mathbf{S}$ is* exhaustive for $\mathcal{C}$ and all ground UCQs *if each ans $\in \mathcal{C}$ that passes $\mathbf{S}$ is $(\mathcal{Q}, \mathcal{T})$-complete for each ground UCQ $\mathcal{Q}$.*

Then, we define the notion of a *ground rewriting* of $\mathcal{T}$—a rewriting that captures all query answers w.r.t. $\mathcal{T}$, regardless of the input ground query and ABox—and we show how to instantiate such ground rewritings.

**Definition 3.45.** *A* ground rewriting *of a TBox $\mathcal{T}$ is a pair $\mathcal{R} = \langle \mathcal{R}_D, \mathcal{R}_\perp \rangle$ such that, for each ground UCQ $\mathcal{Q}$, the triple $\langle \mathcal{R}_D, \mathcal{R}_\perp, \mathcal{Q} \rangle$ is a datalog$^\vee$ rewriting of $\mathcal{T}$ w.r.t. $\mathcal{Q}$. An injective instantiation $\mathbf{I}^\mathcal{R}$ of such $\mathcal{R}$ is defined as $\mathbf{I}^\mathcal{R} = \mathbf{I}^{\mathcal{R}'}$ for $\mathcal{R}' = \langle \mathcal{R}_D, \mathcal{R}_\perp, \emptyset \rangle$.*

Note that Definition 3.45 implies that each variable occurring in the head of a rule in $\mathcal{R}$ also occurs in the rule body. Tools such as REQUIEM and KAON2 can easily be adapted to compute a ground rewriting of a TBox $\mathcal{T}$ in practice. We next show that injective instantiation of a ground rewriting of $\mathcal{T}$ yields a $\mathcal{T}$-test suite that provides us with sufficient and necessary check for completeness w.r.t. all ground UCQs.

**Theorem 3.46.** *Let $\mathcal{T}$ be a TBox, and let $\mathcal{R} = \langle \mathcal{R}_D, \mathcal{R}_\perp \rangle$ be a ground rewriting of $\mathcal{T}$. Then, the following two claims hold.*

1. *$\mathbf{I}^\mathcal{R}$ is exhaustive for $\mathcal{C}_f^\mathcal{T}$ and all ground UCQs.*

2. *Each abstract reasoner ans $\in \mathcal{C}_f^\mathcal{T}$ that does not pass $\mathbf{I}^\mathcal{R}$ is not $(\mathcal{Q}, \mathcal{T})$-complete for some ground UCQ $\mathcal{Q}$.*

*Proof.* (Property 1) Consider an arbitrary abstract reasoner ans $\in \mathcal{C}_f^\mathcal{T}$ that passes $\mathbf{I}^\mathcal{R}$. Let $\mathcal{F}_\mathcal{T}$ be the first-order theory that characterises the behaviour of ans; as in the proof of Theorem 3.43, the fact that ans passes $\mathbf{I}^\mathcal{R}$ implies $\mathcal{F}_\mathcal{T} \models \mathcal{R}_D \cup \mathcal{R}_\perp$. Furthermore, consider an arbitrary ground UCQ $\mathcal{Q}$ and an arbitrary ABox $\mathcal{A}$. That ans is $(\mathcal{Q}, \mathcal{T})$-complete can be shown as in the proof of Theorem 3.43, with the minor difference that $\vec{a} \in \text{cert}(\mathcal{Q}, \mathcal{T}, \mathcal{A})$ implies $\vec{a} \in \text{cert}(\mathcal{Q}, \mathcal{R}_D \cup \mathcal{R}_\perp, \mathcal{A})$ by Definition 3.45.

(Property 2) Note that, since $\mathcal{R}$ is a ground rewriting of $\mathcal{T}$, by Definition 3.41 all UCQs in $\mathbf{I}^\mathcal{R}$ are ground. Thus, if some abstract reasoner ans $\in \mathcal{C}_f^\mathcal{T}$ does not pass $\mathbf{I}^\mathcal{R}$, this clearly shows that ans is not $(\mathcal{Q}, \mathcal{T})$-complete for some ground UCQ $\mathcal{Q}$. $\qquad\square$

## 4. Comparing Incomplete Abstract Reasoners

In this section, we investigate techniques that, given a query $\mathcal{Q}$ and a TBox $\mathcal{T}$, allow us to determine whether an abstract reasoner $\text{ans}_2$ is 'more complete' than an abstract reasoner $\text{ans}_1$—that is, whether for all ABoxes $\mathcal{A}$, abstract reasoner $\text{ans}_2$ computes more answers to $\mathcal{Q}$ and $\mathcal{T}$ than abstract reasoner $\text{ans}_1$. This idea is formalised by the following definition.

**Definition 4.1.** *Let $\mathcal{Q}$ be a UCQ, let $\mathcal{T}$ be a TBox, and let $\mathsf{ans}_1$ and $\mathsf{ans}_2$ be abstract reasoners applicable to $\mathcal{T}$. Then, $\mathsf{ans}_1 \leq_{\mathcal{Q}, \mathcal{T}} \mathsf{ans}_2$ if the following conditions hold for each ABox $\mathcal{A}$:*

*1.* $\mathsf{cert}(*, \mathcal{T}, \mathcal{A}) = \mathsf{t}$ *and* $\mathsf{ans}_1(*, \mathcal{T}, \mathcal{A}) = \mathsf{t}$ *imply* $\mathsf{ans}_2(*, \mathcal{T}, \mathcal{A}) = \mathsf{t}$*; and*

*2.* $\mathsf{cert}(*, \mathcal{T}, \mathcal{A}) = \mathsf{f}$, $\mathsf{ans}_1(*, \mathcal{T}, \mathcal{A}) = \mathsf{f}$, *and* $\mathsf{ans}_2(*, \mathcal{T}, \mathcal{A}) = \mathsf{f}$ *imply*

$$\mathsf{ans}_1(\mathcal{Q}, \mathcal{T}, \mathcal{A}) \cap \mathsf{cert}(\mathcal{Q}, \mathcal{T}, \mathcal{A}) \subseteq \mathsf{ans}_2(\mathcal{Q}, \mathcal{T}, \mathcal{A}) \cap \mathsf{cert}(\mathcal{Q}, \mathcal{T}, \mathcal{A}).$$

*Furthermore, $\mathsf{ans}_1 <_{\mathcal{Q}, \mathcal{T}} \mathsf{ans}_2$ if $\mathsf{ans}_1 \leq_{\mathcal{Q}, \mathcal{T}} \mathsf{ans}_2$ and an ABox $\mathcal{A}$ exists such that at least one of the following two conditions holds:*

*3.* $\mathsf{cert}(*, \mathcal{T}, \mathcal{A}) = \mathsf{t}$, $\mathsf{ans}_1(*, \mathcal{T}, \mathcal{A}) = \mathsf{f}$, *and* $\mathsf{ans}_2(*, \mathcal{T}, \mathcal{A}) = \mathsf{t}$*; or*

*4.* $\mathsf{cert}(*, \mathcal{T}, \mathcal{A}) = \mathsf{f}$, $\mathsf{ans}_1(*, \mathcal{T}, \mathcal{A}) = \mathsf{f}$, $\mathsf{ans}_2(*, \mathcal{T}, \mathcal{A}) = \mathsf{f}$, *and*

$$\mathsf{ans}_1(\mathcal{Q}, \mathcal{T}, \mathcal{A}) \cap \mathsf{cert}(\mathcal{Q}, \mathcal{T}, \mathcal{A}) \subsetneq \mathsf{ans}_2(\mathcal{Q}, \mathcal{T}, \mathcal{A}) \cap \mathsf{cert}(\mathcal{Q}, \mathcal{T}, \mathcal{A}).$$

**Example 4.2.** Consider the abstract reasoners $\mathsf{rdf}$, $\mathsf{rdfs}$, $\mathsf{rl}$, and $\mathsf{classify}$ introduced in Example 3.3 and the query $\mathcal{Q}$ and TBox $\mathcal{T}$ from Example 3.14. We clearly have the following:

$$\mathsf{rdf} \leq_{\mathcal{Q}, \mathcal{T}} \mathsf{rdfs} \leq_{\mathcal{Q}, \mathcal{T}} \mathsf{rl} \leq_{\mathcal{Q}, \mathcal{T}} \mathsf{classify}$$

Furthermore, for any two of these abstract reasoners, an ABox exists that distinguishes the abstracts reasoners w.r.t. $\mathcal{Q}$ and $\mathcal{T}$; for example, for ABox $\mathcal{A}' = \{\mathsf{takesCo}(c, d), \mathsf{MathsCo}(d)\}$, we have $\mathsf{rdfs}(\mathcal{Q}, \mathcal{T}, \mathcal{A}') = \emptyset$ and $\mathsf{rl}(\mathcal{Q}, \mathcal{T}, \mathcal{A}') = \{c\}$. As a result, we also have the following:

$$\mathsf{rdf} <_{\mathcal{Q}, \mathcal{T}} \mathsf{rdfs} <_{\mathcal{Q}, \mathcal{T}} \mathsf{rl} <_{\mathcal{Q}, \mathcal{T}} \mathsf{classify} \qquad \qquad \diamond$$

We would like to check whether $\mathsf{ans}_1 \leq_{\mathcal{Q}, \mathcal{T}} \mathsf{ans}_2$ and $\mathsf{ans}_1 <_{\mathcal{Q}, \mathcal{T}} \mathsf{ans}_2$ for any given pair of abstract reasoners by subjecting the reasoners to a finite set of tests. Towards this goal, we next define the relations $\leq_{\mathcal{Q}, \mathcal{T}}^{\mathbf{R}}$ and $<_{\mathcal{Q}, \mathcal{T}}^{\mathbf{R}}$ that compare abstract reasoners w.r.t. a given finite set $\mathbf{R}$ of ABoxes. Ideally, given $\mathcal{Q}$ and $\mathcal{T}$, we would like to compute a finite $\mathbf{R}$ such that $\leq_{\mathcal{Q}, \mathcal{T}}^{\mathbf{R}}$ and $<_{\mathcal{Q}, \mathcal{T}}^{\mathbf{R}}$ coincide with $\leq_{\mathcal{Q}, \mathcal{T}}$ and $<_{\mathcal{Q}, \mathcal{T}}$ on all abstract reasoners from a class $\mathcal{C}$ of interest. These ideas are captured by the following definitions.

**Definition 4.3.** *Let $\mathcal{Q}$ be a UCQ, let $\mathcal{T}$ be a TBox, let $\mathbf{R}$ be a finite set of ABoxes, and let $\mathsf{ans}_1$ and $\mathsf{ans}_2$ be abstract reasoners applicable to $\mathcal{T}$.*

*Then, $\mathsf{ans}_1 \leq_{\mathcal{Q}, \mathcal{T}}^{\mathbf{R}} \mathsf{ans}_2$ if Conditions 1 and 2 from Definition 4.1 hold for each ABox $\mathcal{A} \in \mathbf{R}$. Furthermore, $\mathsf{ans}_1 <_{\mathcal{Q}, \mathcal{T}}^{\mathbf{R}} \mathsf{ans}_2$ if $\mathsf{ans}_1 \leq_{\mathcal{Q}, \mathcal{T}}^{\mathbf{R}} \mathsf{ans}_2$ and either Condition 3 or Condition 4 from Definition 4.1 holds for some ABox $\mathcal{A} \in \mathbf{R}$.*

**Definition 4.4.** *Let $\mathcal{Q}$ be a UCQ, let $\mathcal{T}$ be a TBox, and let $\mathcal{C}$ be a class of abstract reasoners applicable to $\mathcal{T}$. A finite set $\mathbf{R}$ of ABoxes is $(\mathcal{Q}, \mathcal{T})$-representative for $\mathcal{C}$ if the following conditions hold for all $\mathsf{ans}_1, \mathsf{ans}_2 \in \mathcal{C}$:*

*1.* $\mathsf{ans}_1 \leq_{\mathcal{Q}, \mathcal{T}}^{\mathbf{R}} \mathsf{ans}_2$ *if and only if* $\mathsf{ans}_1 \leq_{\mathcal{Q}, \mathcal{T}} \mathsf{ans}_2$*; and*

2. $\mathsf{ans}_1 <^{\mathbf{R}}_{\mathcal{Q},\mathcal{T}} \mathsf{ans}_2$ *if and only if* $\mathsf{ans}_1 <_{\mathcal{Q},\mathcal{T}} \mathsf{ans}_2$.

As we show next, to prove that $\mathbf{R}$ is $(\mathcal{Q},\mathcal{T})$-representative, it suffices to show the 'only if' implication in Condition 1 and the 'if' implication in Condition 2 from Definition 4.4.

**Proposition 4.5.** *Let $\mathcal{Q}$ be a UCQ, let $\mathcal{T}$ be a TBox, let $\mathcal{C}$ be a class of abstract reasoners applicable to $\mathcal{T}$, and let $\mathbf{R}$ be a finite set of ABoxes such that*

1. $\mathsf{ans}_1 \leq^{\mathbf{R}}_{\mathcal{Q},\mathcal{T}} \mathsf{ans}_2$ *implies* $\mathsf{ans}_1 \leq_{\mathcal{Q},\mathcal{T}} \mathsf{ans}_2$, *and*

2. $\mathsf{ans}_1 <_{\mathcal{Q},\mathcal{T}} \mathsf{ans}_2$ *implies* $\mathsf{ans}_1 <^{\mathbf{R}}_{\mathcal{Q},\mathcal{T}} \mathsf{ans}_2$.

*Then, $\mathbf{R}$ is $(\mathcal{Q},\mathcal{T})$-representative for $\mathcal{C}$.*

*Proof.* Note that $\mathsf{ans}_1 \leq_{\mathcal{Q},\mathcal{T}} \mathsf{ans}_2$ trivially implies $\mathsf{ans}_1 \leq^{\mathbf{R}}_{\mathcal{Q},\mathcal{T}} \mathsf{ans}_2$; thus, Condition 1 of this proposition clearly implies Condition 1 of Definition 4.4. Furthermore, if some ABox $\mathcal{A} \in \mathbf{R}$ satisfies Condition 3 or 4 of Definition 4.1, Condition 1 or 2 of Definition 4.1 holds as well; consequently, Conditions 1 and 2 of this proposition imply Condition 2 of Definition 4.4.  □

An obvious question is whether a $\mathcal{Q}$-simple $\mathcal{T}$-test suite that is exhaustive for a class $\mathcal{C}$ and $\mathcal{Q}$ is also $(\mathcal{Q},\mathcal{T})$-representative for $\mathcal{C}$. The following example shows that this is not necessarily the case.

**Example 4.6.** Let $\mathcal{Q}$ and $\mathcal{T}$ be as specified in Example 3.14, and let $\mathbf{R} = \{\mathcal{A}_1, \ldots, \mathcal{A}_6\}$ for the ABoxes as specified in Example 3.16. As shown in Section 3, the $\mathcal{Q}$-simple $\mathcal{T}$-test suite $\mathbf{S} = \langle \mathbf{S}_\perp, \mathbf{S}_\mathcal{Q} \rangle$ with $\mathbf{S}_\perp = \{\mathcal{A}_6\}$ and $\mathbf{S}_\mathcal{Q} = \{\mathcal{A}_1, \ldots, \mathcal{A}_5\}$ is exhaustive for $\mathcal{C}^{\mathcal{Q},\mathcal{T}}_w$ and $\mathcal{Q}$.

Let $\mathsf{trivial}$ be the abstract reasoner that returns the empty set on each input, and consider also the RDF-based abstract reasoner $\mathsf{rdf}$ from Example 3.3, which ignores the TBox and evaluates the query directly against the ABox. Clearly, $\mathsf{trivial} \leq_{\mathcal{Q},\mathcal{T}} \mathsf{rdf}$; furthermore, $\mathsf{trivial} <_{\mathcal{Q},\mathcal{T}} \mathsf{rdf}$ since for $\mathcal{A} = \{\mathsf{St}(c), \mathsf{takesCo}(c,d), \mathsf{MathCo}(d)\}$ we have $\mathsf{rdf}(\mathcal{Q},\mathcal{T},\mathcal{A}) = \{c\}$ whereas $\mathsf{trivial}(\mathcal{Q},\mathcal{T},\mathcal{A}) = \emptyset$. Both abstract reasoners, however, return the empty set of answers for all ABoxes in $\mathbf{R}$ and thus $\mathsf{rdf} \leq^{\mathbf{R}}_{\mathcal{Q},\mathcal{T}} \mathsf{trivial}$. Hence, by using $\mathbf{R}$ we cannot differentiate the two abstract reasoners. ◇

## 4.1 Negative Result

The following strong result shows that, for numerous TBoxes $\mathcal{T}$, no finite set of ABoxes exists that can differentiate two arbitrary abstract reasoners from the class of all sound, first-order reproducible, monotonic, and strongly faithful reasoners. Note that this result is stronger than the negative result in Theorem 3.21, as it applies to a smaller class of abstract reasoners and all TBoxes that imply at least one concept subsumption.

**Theorem 4.7.** *Let $\mathcal{T}$ be an arbitrary TBox mentioning an atomic role $R$ and atomic concepts $A$ and $B$ such that $\mathcal{T} \models A \sqsubseteq B$, and let $\mathcal{Q} = \{B(x) \rightarrow Q(x)\}$. Then, no finite set of ABoxes exists that is $(\mathcal{Q},\mathcal{T})$-representative for the class of all sound, monotonic, strongly faithful, and first-order reproducible abstract reasoners applicable to $\mathcal{T}$.*

*Proof.* Assume that a finite set of ABoxes $\mathbf{R}$ exists that is $(\mathcal{Q}, \mathcal{T})$-representative for the class of all sound, monotonic, strongly faithful, and first-order reproducible abstract reasoners applicable to $\mathcal{T}$. Let $n$ be the maximum number of assertions in an ABox in $\mathbf{R}$.

For an arbitrary integer $k \geq 1$, let $\mathsf{ans}_k$ be the first-order reproducible abstract reasoner that, given an $\mathcal{FOL}$-TBox $\mathcal{T}_{in}$, uses the following datalog program $\mathcal{F}_{\mathcal{T}_{in}}^k$:

$$\mathcal{F}_{\mathcal{T}_{in}}^k = \begin{cases} \emptyset & \text{if } \mathcal{T}_{in} \not\models A \sqsubseteq B \\ A(x_0) \wedge R(x_0, x_1) \wedge \ldots \wedge R(x_{k-1}, x_k) \rightarrow B(x_0) & \text{if } \mathcal{T}_{in} \models A \sqsubseteq B \end{cases}$$

Clearly, each $\mathsf{ans}_k$ is sound, monotonic, and strongly faithful; furthermore, $\mathsf{ans}_k(*, \mathcal{T}, \mathcal{A}) = \mathsf{f}$ for each ABox $\mathcal{A}$. We next show that $\mathsf{ans}_{n+1}(\mathcal{Q}, \mathcal{T}, \mathcal{A}) \subseteq \mathsf{ans}_{n+2}(\mathcal{Q}, \mathcal{T}, \mathcal{A})$ for each ABox $\mathcal{A} \in \mathbf{R}$. Consider an arbitrary $a_0 \in \mathsf{ans}_{n+1}(\mathcal{Q}, \mathcal{T}, \mathcal{A})$; then, individuals $a_0, a_1, \ldots, a_{n+1}$ exist such that $R(a_{\ell-1}, a_\ell) \in \mathcal{A}$ for each $1 \leq \ell \leq n+1$. Since $\mathcal{A}$ contains at most $n$ assertions but the rule in $\mathcal{F}_{\mathcal{T}}^{n+1}$ contains $n + 1$ body atoms, we have $a_i = a_j$ for some $i \neq j$—that is, $\mathcal{A}$ contains an $R$-cycle. But then, the rule in $\mathcal{F}_{\mathcal{T}}^{n+2}$ can be matched to $\mathcal{A}$ by mapping $x_0$ to $a_0$, so $a_0 \in \mathsf{ans}_{n+2}(\mathcal{Q}, \mathcal{T}, \mathcal{A})$. Therefore, we have $\mathsf{ans}_{n+1} \leq_{\mathcal{Q}, \mathcal{T}}^{\mathbf{R}} \mathsf{ans}_{n+2}$.

For $\mathcal{A} = \{A(a_0), R(a_0, a_1), \ldots, R(a_n, a_{n+1})\}$, however, we have $a_0 \in \mathsf{ans}_{n+1}(\mathcal{Q}, \mathcal{T}, \mathcal{A})$ and $\mathsf{ans}_{n+2}(\mathcal{Q}, \mathcal{T}, \mathcal{A}) = \emptyset$; thus, $\mathsf{ans}_{n+1} \leq_{\mathcal{Q}, \mathcal{T}} \mathsf{ans}_{n+2}$ does not hold, which contradicts our assumption that $\mathbf{R}$ is exhaustive for the class of abstract reasoners from this theorem. □

## 4.2 Compact Abstract Reasoners

Theorem 4.7 suggests that we need to make additional assumptions on the abstract reasoners that we wish to compare using a finite set of ABoxes. In this section, we show that representative sets of ABoxes can be computed in practice if we further restrict ourselves to abstract reasoners that we call $(\mathcal{Q}, \mathcal{T})$-*compact*. Intuitively, such an abstract reasoner processes $\mathcal{Q}$, $\mathcal{T}$, and $\mathcal{A}$ by computing all certain answers of $\mathcal{Q}$, $\mathcal{A}$, and some subset $\mathcal{T}'$ of $\mathcal{T}$, where the subset depends only on $\mathcal{T}$ and $\mathcal{Q}$. In other words, the behaviour of compact abstract reasoners can be simulated by the following process: select the subset of axioms in the input TBox that can be processed, and then compute all certain answers w.r.t. the selected fragment of the TBox. The class of $(\mathcal{Q}, \mathcal{T})$-compact abstract reasoners thus captures the properties of concrete reasoners such as Jena or Oracle's Semantic Data Store that discard axioms from the input TBox that fall outside a certain fragment (e.g., existential restrictions on the right-hand of implications) and then encode the remaining axioms into a suitable set of rules.

**Definition 4.8.** *Let $\mathcal{Q}$ be a UCQ, and let $\mathcal{T}$ be a TBox. An abstract reasoner $\mathsf{ans}$ applicable to $\mathcal{T}$ is $(\mathcal{Q}, \mathcal{T})$-compact if a TBox $\mathcal{T}' \subseteq \mathcal{T}$ exists such that the following properties hold for each ABox $\mathcal{A}$:*

1. $\mathsf{cert}(*, \mathcal{T}', \mathcal{A}) = \mathsf{t}$ *implies* $\mathsf{ans}(*, \mathcal{T}, \mathcal{A}) = \mathsf{t}$;

2. $\mathsf{cert}(*, \mathcal{T}', \mathcal{A}) = \mathsf{f}$ *implies* $\mathsf{ans}(*, \mathcal{T}, \mathcal{A}) = \mathsf{f}$ *and* $\mathsf{ans}(\mathcal{Q}, \mathcal{T}, \mathcal{A}) = \mathsf{cert}(\mathcal{Q}, \mathcal{T}', \mathcal{A})$.

*Abstract reasoner $\mathsf{ans}$ is compact if it is $(\mathcal{Q}, \mathcal{T})$-compact for each UCQ $\mathcal{Q}$ and each TBox $\mathcal{T}$ to which $\mathsf{ans}$ is applicable. Finally, $\mathcal{C}_c^{\mathcal{Q}, \mathcal{T}}$ is the class of all $(\mathcal{Q}, \mathcal{T})$-compact and strongly $(\mathcal{Q}, \mathcal{T})$-faithful abstract reasoners applicable to $\mathcal{T}$.*

**Example 4.9.** All abstract reasoners defined in Example 3.3 are $(\mathcal{Q}, \mathcal{T})$-compact for the query $\mathcal{Q}$ and $\mathcal{EL}$-TBox $\mathcal{T}$ from Example 3.14. Indeed, for abstract reasoner rdf the subset $\mathcal{T}'$ of $\mathcal{T}$ is given by $\mathcal{T}' = \emptyset$; for abstract reasoner rdfs it is $\mathcal{T}' = \{(8)\}$; for abstract reasoner rl it is $\mathcal{T}' = \{(8), (9), (10)\}$; and for abstract reasoner classify it is $\mathcal{T}' = \mathcal{T}$. $\qquad \diamond$

The abstract reasoners $\mathsf{ans}_k$ defined in the proof of of Theorem 4.7 are not $(\mathcal{Q}, \mathcal{T})$-compact for the query and the TBoxes to which Theorem 4.7 applies.

**Proposition 4.10.** *Let $\mathcal{Q} = \{B(x) \to Q(x)\}$ and let $\mathcal{T} = \{A \sqsubseteq B, \; C \sqsubseteq \exists R.\top\}$. Then, for each $k \geq 1$, abstract reasoner $\mathsf{ans}_k$ from the proof of Theorem 4.7 is not $(\mathcal{Q}, \mathcal{T})$-compact.*

*Proof.* Let $\mathcal{Q}$ and $\mathcal{T}$ be as stated in the theorem and consider an arbitrary $k \geq 1$. Let $\mathcal{A}_1$ and $\mathcal{A}_2$ be ABoxes defined as follows:

$$\mathcal{A}_1 = \{A(a_0)\} \qquad \mathcal{A}_2 = \{A(a_0), R(a_0, a_1), \ldots, R(a_{k-1}, a_k)\}$$

Clearly, we have the following:

$$\mathsf{ans}_k(\mathcal{Q}, \mathcal{T}, \mathcal{A}_1) = \emptyset \qquad \mathsf{ans}_k(\mathcal{Q}, \mathcal{T}, \mathcal{A}_2) = \{a_0\}$$

One can straightforwardly check, however, that the following holds for each $\mathcal{T}'$ with $\mathcal{T}' \subseteq \mathcal{T}$:

$$\mathsf{cert}(\mathcal{Q}, \mathcal{T}', \mathcal{A}_1) = \mathsf{cert}(\mathcal{Q}, \mathcal{T}', \mathcal{A}_2)$$

Thus, $\mathsf{ans}_k$ is not $(\mathcal{Q}, \mathcal{T})$-compact. $\qquad \square$

Thus, the negative result from Theorem 4.7 does not immediately apply to a class containing only compact abstract reasoners.

### 4.3 Comparing Compact Abstract Reasoners

In this section, we show that a set of ABoxes that is $(\mathcal{Q}, \mathcal{T})$-representative for $\mathcal{C}_c^{\mathcal{Q}, \mathcal{T}}$ can be obtained by computing, for each subset $\mathcal{T}'$ of $\mathcal{T}$, a $\mathcal{Q}$-simple $\mathcal{T}'$-test suite that is exhaustive for $\mathcal{C}_s^{\mathcal{Q}, \mathcal{T}'}$. A minor complication arises due to the fact that $\mathcal{T}'$ can contain fewer individuals than $\mathcal{T}$. To deal with such cases correctly, the ABoxes in $\mathbf{S}_\perp^{\mathcal{T}'}$ are not allowed to contain individuals occurring in $\mathcal{T}$ but not in $\mathcal{T}'$, and the ABoxes in $\mathbf{S}_\mathcal{Q}^{\mathcal{T}'}$ are not allowed to contain individuals occurring in $\mathcal{T}$ but not in $\mathcal{Q} \cup \mathcal{T}'$. This assumption is without loss of generality: given a $(\mathcal{Q}, \mathcal{T}')$-test suite $\mathbf{S}^{\mathcal{T}'}$, one can replace all individuals in $\mathcal{T}$ but not in $\mathcal{Q} \cup \mathcal{T}'$ with fresh individuals; the result of such replacement is a $(\mathcal{Q}, \mathcal{T}')$-test suite exhaustive for $\mathcal{C}_s^{\mathcal{Q}, \mathcal{T}'}$.

**Theorem 4.11.** *Let $\mathcal{Q}$ be a UCQ, and let $\mathcal{T}$ be a TBox. Furthermore, for each $\mathcal{T}' \subseteq \mathcal{T}$, let $\mathbf{S}^{\mathcal{T}'} = \langle \mathbf{S}_\perp^{\mathcal{T}'}, \mathbf{S}_\mathcal{Q}^{\mathcal{T}'} \rangle$ be a $\mathcal{Q}$-simple $\mathcal{T}'$-test suite that is exhaustive for $\mathcal{C}_s^{\mathcal{Q}, \mathcal{T}'}$ and $\mathcal{Q}$ such that no ABox in $\mathbf{S}_\perp^{\mathcal{T}'}$ contains an individual from $\mathsf{ind}(\mathcal{T}) \setminus \mathsf{ind}(\mathcal{T}')$ and no ABox in $\mathbf{S}_\mathcal{Q}^{\mathcal{T}'}$ contains an individual from $\mathsf{ind}(\mathcal{T}) \setminus \mathsf{ind}(\mathcal{Q} \cup \mathcal{T}')$. Then, the set $\mathbf{R}$ of ABoxes defined by*

$$\mathbf{R} = \bigcup_{\mathcal{T}' \subseteq \mathcal{T}} \mathbf{S}_\perp^{\mathcal{T}'} \cup \mathbf{S}_\mathcal{Q}^{\mathcal{T}'}$$

*is $(\mathcal{Q}, \mathcal{T})$-representative for $\mathcal{C}_c^{\mathcal{Q}, \mathcal{T}}$.*

*Proof.* Assume that **R** satisfies the conditions of the theorem, and let $\mathsf{ans}_1$ and $\mathsf{ans}_2$ be arbitrary abstract reasoners in $\mathcal{C}_c^{\mathcal{Q},\mathcal{T}}$. We next show that $\mathsf{ans}_1$ and $\mathsf{ans}_2$ satisfy the two properties in Proposition 4.5.

- Property 1 of Proposition 4.5:  $\mathsf{ans}_1 \leq_{\mathcal{Q},\mathcal{T}}^{\mathbf{R}} \mathsf{ans}_2$ implies $\mathsf{ans}_1 \leq_{\mathcal{Q},\mathcal{T}} \mathsf{ans}_2$

Since $\mathsf{ans}_1$ is $(\mathcal{Q},\mathcal{T})$-compact, a TBox $\mathcal{T}' \subseteq \mathcal{T}$ exists that satisfies the conditions of Definition 4.8. Assume that $\mathsf{ans}_1 \leq_{\mathcal{Q},\mathcal{T}}^{\mathbf{R}} \mathsf{ans}_2$; we next show that Conditions 1 and 2 of Definition 4.1 are satisfied for an arbitrary ABox $\mathcal{A}$.

(Condition 1) Assume that $\mathsf{cert}(*,\mathcal{T},\mathcal{A}) = \mathsf{t}$ and $\mathsf{ans}_1(*,\mathcal{T},\mathcal{A}) = \mathsf{t}$. By the contrapositive of property 2 of Definition 4.8, then $\mathsf{cert}(*,\mathcal{T}',\mathcal{A}) = \mathsf{t}$. Since **R** contains all the ABoxes of some $\mathcal{Q}$-simple $\mathcal{T}'$-test suite that is exhaustive for $\mathcal{C}_s^{\mathcal{Q},\mathcal{T}'}$ and $\mathcal{Q}$, by Theorem 3.30 there exist an ABox $\mathcal{A}' \in \mathbf{R}$ and a $\mathcal{T}'$-stable renaming $\mu$ such that $\mathsf{dom}(\mu) = \mathsf{ind}(\mathcal{T}' \cup \mathcal{A}')$ and $\mu(\mathcal{A}') \subseteq \mathcal{A}$; since $\mathcal{A}'$ does not contain individuals from $\mathsf{ind}(\mathcal{T}) \setminus \mathsf{ind}(\mathcal{T}')$, renaming $\mu$ is also $\mathcal{T}$-stable. By the definition of a $\mathcal{T}'$-test suite, $\mathsf{cert}(*,\mathcal{T}',\mathcal{A}') = \mathsf{t}$; furthermore, by property 1 of Definition 4.8 we have $\mathsf{ans}_1(*,\mathcal{T},\mathcal{A}') = \mathsf{t}$. Since $\mathsf{ans}_1 \leq_{\mathcal{Q},\mathcal{T}}^{\mathbf{R}} \mathsf{ans}_2$ we have $\mathsf{ans}_2(*,\mathcal{T},\mathcal{A}') = \mathsf{t}$. Since $\mathsf{ans}_2$ is strongly $(\mathcal{Q},\mathcal{T})$-faithful and $\mu$ is $\mathcal{T}$-stable, we have $\mathsf{ans}_2(*,\mathcal{T},\mu(\mathcal{A}')) = \mathsf{t}$. Finally, since $\mu(\mathcal{A}') \subseteq \mathcal{A}$ and $\mathsf{ans}_2$ is $(\mathcal{Q},\mathcal{T})$-monotonic, we have $\mathsf{ans}_2(*,\mathcal{T},\mathcal{A}) = \mathsf{t}$, as required.

(Condition 2) Assume that $\mathsf{cert}(*,\mathcal{T},\mathcal{A}) = \mathsf{f}$, $\mathsf{ans}_1(*,\mathcal{T},\mathcal{A}) = \mathsf{f}$, and $\mathsf{ans}_2(*,\mathcal{T},\mathcal{A}) = \mathsf{f}$, and consider an arbitrary tuple $\vec{a} \in \mathsf{ans}(\mathcal{Q},\mathcal{T},\mathcal{A}) \cap \mathsf{cert}(\mathcal{Q},\mathcal{T},\mathcal{A})$. By the contrapositive of property 1 of Definition 4.8, then $\mathsf{cert}(*,\mathcal{T}',\mathcal{A}) = \mathsf{f}$; but then, by property 2 of Definition 4.8, we have $\vec{a} \in \mathsf{cert}(\mathcal{Q},\mathcal{T}',\mathcal{A})$. Since **R** contains all the ABoxes of some $\mathcal{Q}$-simple $\mathcal{T}'$-test suite that is exhaustive for $\mathcal{C}_s^{\mathcal{Q},\mathcal{T}'}$ and $\mathcal{Q}$, by Theorem 3.30 there exist an ABox $\mathcal{A}' \in \mathbf{R}$, a tuple $\vec{b} \in \mathsf{cert}(\mathcal{Q},\mathcal{T}',\mathcal{A}')$, and a $(\mathcal{Q},\mathcal{T}')$-stable renaming $\mu$ such that $\mathsf{dom}(\mu) = \mathsf{ind}(\mathcal{Q} \cup \mathcal{T}' \cup \mathcal{A}')$, $\mu(\mathcal{A}') \subseteq \mathcal{A}$, and $\mu(\vec{b}) = \vec{a}$; since $\mathcal{A}'$ does not contain individuals from $\mathsf{ind}(\mathcal{T}) \setminus \mathsf{ind}(\mathcal{Q} \cup \mathcal{T}')$, renaming $\mu$ is also $(\mathcal{Q},\mathcal{T})$-stable. By the definition of a $(\mathcal{Q},\mathcal{T}')$-test suite, $\mathsf{cert}(*,\mathcal{T}',\mathcal{A}') = \mathsf{f}$; furthermore, by property 2 of Definition 4.8 we have $\vec{b} \in \mathsf{ans}_1(\mathcal{Q},\mathcal{T},\mathcal{A}')$. Since $\mathsf{ans}_1 \leq_{\mathcal{Q},\mathcal{T}}^{\mathbf{R}} \mathsf{ans}_2$ we have $\vec{b} \in \mathsf{ans}_2(\mathcal{Q},\mathcal{T},\mathcal{A}')$. Since $\mathsf{ans}_2$ is strongly $(\mathcal{Q},\mathcal{T})$-faithful and $\mu$ is $(\mathcal{Q},\mathcal{T})$-stable, we have that $\vec{a} \in \mathsf{ans}_2(\mathcal{Q},\mathcal{T},\mu(\mathcal{A}'))$. Finally, since $\mu(\mathcal{A}') \subseteq \mathcal{A}$ and $\mathsf{ans}_2$ is $(\mathcal{Q},\mathcal{T})$-monotonic, we have $\vec{a} \in \mathsf{ans}_2(\mathcal{Q},\mathcal{T},\mathcal{A})$, as required.

- Property 2 of Proposition 4.5:  $\mathsf{ans}_1 <_{\mathcal{Q},\mathcal{T}} \mathsf{ans}_2$ implies $\mathsf{ans}_1 <_{\mathcal{Q},\mathcal{T}}^{\mathbf{R}} \mathsf{ans}_2$

Assume that $\mathsf{ans}_1 <_{\mathcal{Q},\mathcal{T}} \mathsf{ans}_2$. By Definition 4.1, then $\mathsf{ans}_1 \leq_{\mathcal{Q},\mathcal{T}} \mathsf{ans}_2$ and an ABox $\mathcal{A}$ exists satisfying Conditions 3 and 4 of Definition 4.1. Clearly, $\mathsf{ans}_1 \leq_{\mathcal{Q},\mathcal{T}}^{\mathbf{R}} \mathsf{ans}_2$; hence, what remains to be shown is that **R** contains an ABox that satisfies Conditions 3 and 4 of Definition 4.1. Since $\mathsf{ans}_1$ is $(\mathcal{Q},\mathcal{T})$-compact, a TBox $\mathcal{T}' \subseteq \mathcal{T}$ exists that satisfies the conditions of Definition 4.8.

(Condition 3) Assume that $\mathsf{cert}(*,\mathcal{T},\mathcal{A}) = \mathsf{t}$, and assume also that $\mathsf{ans}_1(*,\mathcal{T},\mathcal{A}) = \mathsf{t}$ and $\mathsf{ans}_2(*,\mathcal{T},\mathcal{A}) = \mathsf{f}$. As in the proof of Condition 1, we can identify an ABox $\mathcal{A}' \in \mathbf{R}$ and a $\mathcal{T}$-stable renaming $\mu$ such that $\mathsf{ans}_1(*,\mathcal{T},\mathcal{A}') = \mathsf{t}$ and $\mu(\mathcal{A}') \subseteq \mathcal{A}$. Since $\mathsf{ans}_2$ is $(\mathcal{Q},\mathcal{T})$-monotonic and $\mathsf{ans}_2(*,\mathcal{T},\mathcal{A}) = \mathsf{f}$, we have $\mathsf{ans}_2(*,\mathcal{T},\mu(\mathcal{A}')) = \mathsf{f}$; furthermore, since $\mathsf{ans}_2$ is strongly $(\mathcal{Q},\mathcal{T})$-faithful and $\mu$ is $\mathcal{T}$-stable, we also have $\mathsf{ans}_2(*,\mathcal{T},\mathcal{A}') = \mathsf{f}$. But then, Condition 3 of Definition 4.1 is satisfied for $\mathcal{A}' \in \mathbf{R}$.

(Condition 4) Assume that $\mathsf{cert}(*,\mathcal{T},\mathcal{A}) = \mathsf{f}$ and $\mathsf{ans}_1(*,\mathcal{T},\mathcal{A}) = \mathsf{ans}_2(*,\mathcal{T},\mathcal{A}) = \mathsf{f}$, and consider an arbitrary tuple $\vec{a} \in [\mathsf{ans}_1(\mathcal{Q},\mathcal{T},\mathcal{A}) \cap \mathsf{cert}(\mathcal{Q},\mathcal{T},\mathcal{A})] \setminus \mathsf{ans}_2(\mathcal{Q},\mathcal{T},\mathcal{A})$. As in the proof of Condition 2, we can identify an ABox $\mathcal{A}' \in \mathbf{R}$, a $(\mathcal{Q},\mathcal{T})$-stable renaming $\mu$, and a tuple $\vec{b} \in \mathsf{cert}(\mathcal{Q}, \mathcal{T}', \mathcal{A}')$ such that $\mu(\mathcal{A}') \subseteq \mathcal{A}$, $\mu(\vec{b}) = \vec{a}$, and $\vec{b} \in \mathsf{ans}_1(\mathcal{Q}, \mathcal{T}, \mathcal{A}')$. Since $\mathsf{ans}_2$ is $(\mathcal{Q}, \mathcal{T})$-monotonic and $\vec{a} \notin \mathsf{ans}_2(\mathcal{Q}, \mathcal{T}, \mathcal{A})$, we have $\vec{a} \notin \mathsf{ans}_2(\mathcal{Q}, \mathcal{T}, \mu(\mathcal{A}'))$; furthermore, since $\mathsf{ans}_2$ is strongly $(\mathcal{Q}, \mathcal{T})$-faithful and $\mu$ is $(\mathcal{Q}, \mathcal{T})$-stable, we also have $\vec{b} \notin \mathsf{ans}_2(\mathcal{Q}, \mathcal{T}, \mathcal{A}')$. But then, Condition 4 of Definition 4.1 is satisfied for $\mathcal{A}' \in \mathbf{R}$. □

Theorems 3.32 and 4.11 immediately suggest an approach for computing a set of ABoxes that is a $(\mathcal{Q}, \mathcal{T})$-representative for $\mathcal{C}_c^{\mathcal{Q}, \mathcal{T}}$. First, we compute a UCQ rewriting of $\mathcal{Q}$ w.r.t. each subset of $\mathcal{T}$; then, we instantiate each rule in each such rewriting using an injective instantiation mapping; finally, we compute $\mathbf{R}$ as a union of all ABoxes in all test suites. Such a naïve procedure, however, is not practical since it requires computing an exponential number of UCQ rewritings. We next present a more practical approach to computing a set of ABoxes that is $(\mathcal{Q}, \mathcal{T})$-representative for $\mathcal{C}_c^{\mathcal{Q}, \mathcal{T}}$. Intuitively, instead of computing exponentially many rewritings, one can compute a single UCQ rewriting of $\mathcal{Q}$ w.r.t. $\mathcal{T}$ that is *subset-closed*—that is, which contains a rewriting for each subset of $\mathcal{T}$.

**Definition 4.12.** *A UCQ rewriting* $\mathcal{R} = \langle \mathcal{R}_\perp, \mathcal{R}_Q \rangle$ *of* $\mathcal{Q}$ *w.r.t.* $\mathcal{T}$ *is subset-closed if for each* $\mathcal{T}' \subseteq \mathcal{T}$ *a tuple* $\mathcal{R}' = \langle \mathcal{R}'_\perp, \mathcal{R}'_Q \rangle$ *exists such that* $\mathcal{R}'_\perp \subseteq \mathcal{R}_\perp$, $\mathcal{R}'_Q \subseteq \mathcal{R}_Q$ *and* $\mathcal{R}'$ *is a UCQ rewriting of* $\mathcal{Q}$ *w.r.t.* $\mathcal{T}'$.

The following corollary is an immediate consequence of Theorems 3.27, 3.32, and 4.11.

**Corollary 4.13.** *Let* $\mathcal{Q}$ *be a UCQ, let* $\mathcal{T}$ *be a TBox, let* $\mathcal{R}$ *be a subset-closed UCQ rewriting of* $\mathcal{Q}$ *w.r.t.* $\mathcal{T}$, *and let* $\mathbf{I}^{\mathcal{R}} = \langle \mathbf{I}_\perp^{\mathcal{R}}, \mathbf{I}_Q^{\mathcal{R}} \rangle$ *be the injective instantiation of* $\mathcal{R}$. *Then, the set of ABoxes* $\mathbf{R} = \mathbf{I}_\perp^{\mathcal{R}} \cup \mathbf{I}_Q^{\mathcal{R}}$ *is* $(\mathcal{Q}, \mathcal{T})$-representative for $\mathcal{C}_c^{\mathcal{Q}, \mathcal{T}}$.

Practical query rewriting systems such as REQUIEM are optimised to produce as small a UCQ rewriting as possible, so their output is typically not subset-closed. Therefore, our technique requires the modification of UCQ rewriting algorithms implemented in existing systems. As illustrated by the following example, the required modification typically involves disabling (at least partially) subsumption-based optimisations.

**Example 4.14.** Let $\mathcal{Q}$ and $\mathcal{T}$ be as specified in Example 3.14, and let $\mathbf{S} = \langle \mathbf{S}_\perp, \mathbf{S}_Q \rangle$ be the $\mathcal{T}$-test suite from Example 3.16. A system such as REQUIEM can compute such $\mathcal{R}$ for the given $\mathcal{Q}$ and $\mathcal{T}$. Note, however, that $\mathcal{R}$ is not subset-closed; for example, a UCQ rewriting of $\mathcal{Q}$ w.r.t. $\mathcal{T}' = \emptyset$ is $\mathcal{Q}$, and it is not a subset of $\mathcal{R}_Q$. The rewriting can be made subset-closed by extending $\mathcal{R}_Q$ with the following rules:

$$\mathsf{St}(x) \wedge \mathsf{takesCo}(x, y) \wedge \mathsf{MathCo}(x, y) \rightarrow Q(x)$$
$$\mathsf{St}(x) \wedge \mathsf{takesCo}(x, y) \wedge \mathsf{CalcCo}(x, y) \rightarrow Q(x)$$
$$\mathsf{MathSt}(x) \wedge \mathsf{St}(x) \rightarrow Q(x)$$

Systems such as REQUIEM, however, typically discard such rules by applying subsumption optimisations described in Section 3.5.3. ◇

As the following example shows, a subset-closed UCQ rewriting of $\mathcal{Q}$ w.r.t. $\mathcal{T}$ can, in the worst case, be exponentially larger than the 'minimal' UCQ rewritings of $\mathcal{Q}$ w.r.t. $\mathcal{T}$.

**Example 4.15.** Let $\mathcal{Q} = \{C(x) \rightarrow Q(x)\}$, and let $\mathcal{T}$ be the following TBox:

$$\mathcal{T} = \{B \sqsubseteq A_i \mid 1 \leq i \leq n\} \cup \{A_1 \sqcap \ldots \sqcap A_n \sqsubseteq C\}$$

Furthermore, let $\mathcal{R} = \langle \mathcal{R}_\perp, \mathcal{R}_Q \rangle$ be such that $\mathcal{R}_\perp = \emptyset$ and $\mathcal{R}_Q$ contains the following rules:

$$C(x) \to Q(x)$$
$$B(x) \to Q(x)$$
$$A_1(x) \wedge \ldots \wedge A_n(x) \to Q(x)$$

Clearly, $\mathcal{R}$ is a UCQ rewriting of $\mathcal{Q}$ w.r.t. $\mathcal{T}$; however, the number of rules in a subset-closed UCQ rewriting of $\mathcal{Q}$ w.r.t. $\mathcal{T}$ is exponential in $n$. ◇

## 5. Evaluation

We implemented our techniques for computing exhaustive test suites and for comparing incomplete concrete reasoners in a prototype tool called SyGENiA.[1] Our tool uses REQUIEM for computing UCQ and datalog rewritings.[2]

We considered two evaluation scenarios. The first one uses the well-known Lehigh University Benchmark (LUBM) (Guo et al., 2005), which consists of a relatively small TBox about an academic domain, 14 test queries, and a data generator. The second one uses a small version of GALEN (Rector & Rogers, 2006)—a complex ontology commonly used in medical applications.

We evaluated the following concrete reasoners: Sesame v2.3-prl,[3] DLE-Jena v2.0,[4] OWLim v2.9.1,[5] Minerva v1.5,[6] and Jena v2.6.3[7] in all of its three variants (Micro, Mini, and Max).

### 5.1 Computing Exhaustive Test Suites

Given a UCQ $\mathcal{Q}$ and a TBox $\mathcal{T}$, our tool uses REQUIEM to compute a datalog rewriting $\mathcal{R}$ for $\mathcal{Q}$ and $\mathcal{T}$. If $\mathcal{R}$ is a UCQ rewriting, then our tool computes a simple test suite by either full or injective instantiation (see Sections 3.5 and 3.6, respectively); otherwise, the tool computes a non-simple test suite by instantiating $\mathcal{R}$ as described in Section 3.7.3.

#### 5.1.1 Simple Test Suites

In the case of the LUBM benchmark, each of the 14 test queries leads to a UCQ rewriting w.r.t. the TBox.[8] Therefore, we computed a UCQ rewriting for each query $\mathcal{Q}$ in the benchmark using REQUIEM and instantiated it, both fully and injectively, thus obtaining $\mathcal{Q}$-simple $\mathcal{T}$-test suites that are exhaustive for $\mathcal{Q}$ and $\mathcal{C}_w^{\mathcal{Q},\mathcal{T}}$ and $\mathcal{C}_s^{\mathcal{Q},\mathcal{T}}$, respectively. The times needed to compute the test suites and the size of each test suite are shown in Table 3, where ♯**S** denotes the total number of ABoxes in the corresponding test suites.

---

1. `http://code.google.com/p/sygenia/`
2. `http://www.cs.ox.ac.uk/projects/requiem/home.html`
3. `http://www.openrdf.org/`
4. `http://lpis.csd.auth.gr/systems/DLE-Jena/`
5. `http://www.ontotext.com/owlim/`
6. `http://www.alphaworks.ibm.com/tech/semanticstk`
7. `http://jena.sourceforge.net/`
8. Since REQUIEM does not currently support individuals in the queries, we replaced the individuals in queries by distinguished variables.

|  |  | $\mathcal{Q}_1$ | $\mathcal{Q}_2$ | $\mathcal{Q}_3$ | $\mathcal{Q}_4$ | $\mathcal{Q}_5$ | $\mathcal{Q}_6$ | $\mathcal{Q}_7$ | $\mathcal{Q}_8$ | $\mathcal{Q}_9$ | $\mathcal{Q}_{10}$ | $\mathcal{Q}_{11}$ | $\mathcal{Q}_{12}$ | $\mathcal{Q}_{13}$ | $\mathcal{Q}_{14}$ |
|---|---|---|---|---|---|---|---|---|---|---|---|---|---|---|---|
| $\mathcal{C}_w^{\mathcal{Q},\mathcal{T}}$ | Time | 1.2 | 0.7 | 0.2 | 6.7 | 0.2 | 2.1 | 0.7 | 7 | 2.4 | 7.4 | 0.07 | 0.2 | 0.2 | 0.05 |
|  | ♯**S** | 2 | 20 | 2 | 2 352 | 8 | 1207 | 345 | 3 092 | 5 | 3 919 | 7 | 25 | 10 | 1 |
| $\mathcal{C}_s^{\mathcal{Q},\mathcal{T}}$ | Time | 1.2 | 0.6 | 0.2 | 0.6 | 0.2 | 1.2 | 0.4 | 1.5 | 2.5 | 0.6 | 0.05 | 0.1 | 0.1 | 0.08 |
|  | ♯**S** | 1 | 4 | 1 | 22 | 4 | 169 | 37 | 36 | 1 | 169 | 2 | 3 | 5 | 1 |

Table 3: Computation of simple test suites for LUBM. Times are given in seconds.

|  |  | $\mathcal{Q}'_1$ | $\mathcal{Q}'_2$ | $\mathcal{Q}'_3$ | $\mathcal{Q}'_4$ |
|---|---|---|---|---|---|
| $\mathcal{C}_w^{\mathcal{Q},\mathcal{T}}$ | Time | 14 | 34 | 67 | 4.6 |
|  | ♯**S** | 6 049 | 12 085 | 12 085 | 79 |
| $\mathcal{C}_s^{\mathcal{Q},\mathcal{T}}$ | Time | 2.2 | 6 | 40 | 1.7 |
|  | ♯**S** | 79 | 151 | 151 | 25 |

Table 4: Computation of simple test suites for GALEN. Times are given in seconds.

As shown in the table, simple test suites could be computed in times ranging from 0.05 to 7 seconds, both for $\mathcal{C}_w^{\mathcal{Q},\mathcal{T}}$ and $\mathcal{C}_s^{\mathcal{Q},\mathcal{T}}$. The optimisations implemented in REQUIEM ensure that the UCQ rewritings are relatively small, so the resulting test suites also consist of a relatively small number of ABoxes. Notice, however, the significant difference between the numbers of ABoxes in test suites obtained via injective instantiation (which range from 1 to 169 with an average of 32), and those obtained via full instantiation (which range from 1 to 3, 919 with an average of 702). Furthermore, each rule in a rewriting contains at most 6 atoms, therefore each ABox in a test suite also contains at most 6 assertions.

In the case of GALEN, we used the following sample queries, for which REQUIEM can compute a UCQ rewriting:

$$\mathcal{Q}'_1: \quad \mathsf{HaemoglobinConcentrationProcedure}(x) \to Q(x)$$
$$\mathcal{Q}'_2: \quad \mathsf{PlateletCountProcedure}(x) \to Q(x)$$
$$\mathcal{Q}'_3: \quad \mathsf{LymphocyteCountProcedure}(x) \to Q(x)$$
$$\mathcal{Q}'_4: \quad \mathsf{HollowStructure}(x) \to Q(x)$$

We instantiated each UCQ rewriting both fully and injectively. The times needed to compute the test suites and the size of each test suite are shown in Table 4.

As shown in the table, simple test suites for GALEN can be computed in times ranging from 1.7 to 67 seconds with an average of 33 seconds. Thus, computing test suites for GALEN is more time consuming than for LUBM. This is unsurprising since the TBox of GALEN is significantly more complex than that of LUBM. The number of ABoxes in the test suites ranged from 25 to 151 in the case of injective instantiations and from 79 to over 12, 000 in the case of full instantiations; again, note the significant difference between the sizes of the two kinds of test suites. In all cases, however, each individual ABox was very small, with the largest one containing only 11 assertions.

### 5.1.2 Non-Simple Test Suites

We also computed non-simple test suites for cases where no UCQ rewriting exists. As already mentioned, all LUBM queries are UCQ-rewritable. Therefore, we manually added the following query, for which REQUIEM computes a recursive datalog rewriting.

| | | **LUBM** | **GALEN** | | | |
|---|---|---|---|---|---|---|
| | | $\mathcal{Q}_{15}$ | $\mathcal{Q}_5'$ | $\mathcal{Q}_6'$ | $\mathcal{Q}_7'$ | $\mathcal{Q}_8'$ |
| $\mathcal{C}_f^{\mathcal{T}}$ | Time (s) | 1.4 | 5.2 | 1.3 | 2.7 | 1.6 |
| | $\sharp\mathbf{S}$ | 22 | 41 | 23 | 31 | 12 |

Table 5: General test suites computed from datalog rewritings for LUBM and GALEN.

| **System** | **Completeness Guarantee** | **Completeness w.r.t. LUBM data set** |
|---|---|---|
| JenaMax/DLE-Jena | $\mathcal{Q}_1$–$\mathcal{Q}_{14}$ | $\mathcal{Q}_1$–$\mathcal{Q}_{14}$ |
| OWLim | $\mathcal{Q}_1$–$\mathcal{Q}_5$, $\mathcal{Q}_7$, $\mathcal{Q}_9$, $\mathcal{Q}_{11}$–$\mathcal{Q}_{14}$ | $\mathcal{Q}_1$–$\mathcal{Q}_{14}$ |
| Jena Mini/Micro | $\mathcal{Q}_1$–$\mathcal{Q}_5$, $\mathcal{Q}_7$, $\mathcal{Q}_9$, $\mathcal{Q}_{11}$–$\mathcal{Q}_{14}$ | $\mathcal{Q}_1$–$\mathcal{Q}_{14}$ |
| Minerva | $\mathcal{Q}_1$–$\mathcal{Q}_4$, $\mathcal{Q}_9$, $\mathcal{Q}_{14}$ | $\mathcal{Q}_1$–$\mathcal{Q}_{14}$ |
| Sesame | $\mathcal{Q}_1$, $\mathcal{Q}_3$, $\mathcal{Q}_{11}$, $\mathcal{Q}_{14}$ | $\mathcal{Q}_1$–$\mathcal{Q}_5$, $\mathcal{Q}_{11}$, $\mathcal{Q}_{14}$ |

Table 6: Completeness guarantees for UCQ-rewritable queries in LUBM

$$\mathcal{Q}_{15}: \quad \mathsf{Organization}(x) \rightarrow Q(x)$$

Due to the complex structure of the GALEN TBox, test queries that are not UCQ rewritable can be easily identified. We have evaluated the following four.

$$\mathcal{Q}_5': \quad \mathsf{WestergrenESRProcedure}(x) \rightarrow Q(x)$$
$$\mathcal{Q}_6': \quad \mathsf{ArthroscopicProcedure}(x) \rightarrow Q(x)$$
$$\mathcal{Q}_7': \quad \mathsf{TrueCavity}(x) \rightarrow Q(x)$$
$$\mathcal{Q}_8': \quad \mathsf{BacterialCellWall}(x) \rightarrow Q(x)$$

Times needed to compute test suites and the size of each test suite are shown in Table 5.

## 5.2 Completeness Guarantees

As already discussed, existing concrete reasoners are captured by strongly $(\mathcal{Q}, \mathcal{T})$-faithful abstract reasoners. Hence, in order to establish completeness guarantees for such concrete reasoners, we restricted our tests to test suites computed using injective instantiations.

### 5.2.1 Results for Simple Test Suites

Our results for the original queries of the LUBM benchmark are shown in Table 6. For each concrete reasoner, the first column of the table shows the queries for which we were able to prove completeness using our techniques (i.e., the queries that are complete for an arbitrary data set), and the second column of the table shows the queries on which the concrete reasoner computes all answers on the canonical LUBM data set with one university. Our results clearly show that completeness w.r.t. the data set in the LUBM benchmark is no guarantee of completeness for arbitrary data sets; for example, OWLim, Minerva, and Jena Mini/Micro are complete for all queries w.r.t. the LUBM data set (and some of these systems are even complete for the more expressive UOBM benchmark); however, for certain queries, these systems were found to be incomplete for a data set in our test suites.

Jena Max and DLE-Jena are the only systems that are guaranteed to be complete for all 14 LUBM queries regardless of the data set—that is, these systems behave exactly like a complete OWL reasoner for all LUBM queries and the LUBM TBox. According to Jena's documentation, Jena Max supports all types of axioms used in the LUBM TBox, hence it is expected to be complete for the LUBM TBox and queries. Interestingly, when tested with some of the LUBM data sets, Jena Max could not compute the answers to many of the queries, so we used smaller LUBM data sets instead. This demonstrates an additional advantage of our approach: it does not require reasoning w.r.t. very large data sets, since the ABoxes in test suites typically contain only a small number of assertions. Regarding DLE-Jena, according to its technical description (Meditskos & Bassiliades, 2008), the system uses a complete DL reasoner to materialise certain subsumptions in a preprocessing step and then uses Jena to saturate the ABox, much like the abstract reasoner classify from Example 3.3. Hence, DLE-Jena is at least as complete as Jena Mini and, in addition, it is able to draw the inferences that Jena Mini is missing (see below).

OWLim is complete for all LUBM queries that do not involve reasoning with existential quantifiers in the consequent of implications. It is well known that the latter is not supported by the system. Jena Mini and Micro exhibited exactly the same behaviour as OWLim, despite the fact that Jena Mini can handle a larger fragment of OWL than OWLim. Clearly, the LUBM TBox and queries are not sufficiently complex to reveal the differences between OWLim, and Jena Mini/Micro.

Minerva is guaranteed to be complete for only six queries. Like DLE-Jena, it uses a DL reasoner to materialise entailed subsumptions between atomic concepts, but it uses a custom method for saturating the ABox. After investigating several ABoxes from the test suites we concluded that Minerva cannot correctly handle (at-least) inverse role axioms; for example, it cannot find the entailment $\{\ R \sqsubseteq R^-,\ R(a,b)\ \} \models R(b,a)$.

Finally, Sesame is complete for only four queries. This is unsurprising since Sesame is an RDFS reasoner and is thus complete only for a small fragment of OWL 2 DL.

We next discuss the results of tests based on the GALEN ontology and test queries $\mathcal{Q}_1'$–$\mathcal{Q}_4'$. We could not run Jena Max since GALEN heavily uses existential restrictions, which (according to Jena's documentation) might cause problems. Minerva was the only system that provided completeness guarantee for at least one query ($\mathcal{Q}_4'$); this is because Minerva precomputes subsumption relationships between atomic concepts that depend on existential restrictions on the right hand side of TBox axioms, which most other systems do not handle. Also, unlike LUBM, the version of GALEN that we used does not contain inverse roles, so Minerva performed much better on this ontology. All other systems were identified as being incomplete for all test queries.

### 5.2.2 RESULTS FOR NON-SIMPLE TEST SUITES

Results for test queries that are not UCQ-rewritable are summarised in Table 7. Symbol '✓' indicates that the concrete reasoner was found complete for the given query. Furthermore, whenever a concrete reasoner failed a test suite, we tried to prove the reasoner to be incomplete as discussed in the examples in Section 3.7.2; in all cases we were successful, so symbol '×' indicates that the concrete reasoner was identified as being incomplete for a given query. Finally, symbol '−' indicates that the concrete reasoner ran out of memory.

In the case of LUBM, we were able to establish completeness guarantees w.r.t. query $\mathcal{Q}_{15}$ for OWLim, Jena Micro, DLE-Jena, and Jena Max. Note that all these systems can handle recursive TBox statements, so completeness for $\mathcal{Q}_{15}$ is not surprising. RDFS, however, cannot express recursive TBox statements involving roles, so Sesame—an RDFS reasoner—fails to compute certain answers to some tests.

| | LUBM | GALEN | | | |
|---|---|---|---|---|---|
| | $\mathcal{Q}_{15}$ | $\mathcal{Q}'_5$ | $\mathcal{Q}'_6$ | $\mathcal{Q}'_7$ | $\mathcal{Q}'_8$ |
| OWLim | ✓ | × | × | × | ✓ |
| Jena Max | ✓ | - | - | - | - |
| Jena Micro | ✓ | × | × | × | ✓ |
| DLE-Jena | ✓ | × | × | × | ✓ |
| Minerva | × | × | ✓ | × | ✓ |
| Sesame | × | × | × | × | × |

Table 7: Completeness guarantees for datalog-rewritable queries

| | | $\mathcal{Q}_1$ | $\mathcal{Q}_2$ | $\mathcal{Q}_3$ | $\mathcal{Q}_4$ | $\mathcal{Q}_5$ | $\mathcal{Q}_6$ | $\mathcal{Q}_7$ | $\mathcal{Q}_8$ | $\mathcal{Q}_9$ | $\mathcal{Q}_{10}$ | $\mathcal{Q}_{11}$ | $\mathcal{Q}_{12}$ | $\mathcal{Q}_{13}$ | $\mathcal{Q}_{14}$ |
|---|---|---|---|---|---|---|---|---|---|---|---|---|---|---|---|
| $\mathcal{C}_s^{\mathcal{Q},\mathcal{T}}$ | Time | 1.4 | 1.1 | 0.2 | 1.8 | 0.8 | 1.2 | 9.5 | 7.8 | - | 0.9 | 0.05 | 0.5 | 0.6 | 0.04 |
| | ♯**R** | 1 | 24 | 17 | 130 | 136 | 219 | 925 | 777 | - | 219 | 2 | 74 | 185 | 1 |

Table 8: Representative sets of ABoxes for LUBM. Times are given in seconds.

In the case of GALEN, completeness is guaranteed on query $\mathcal{Q}'_8$ for OWLim, Jena Micro, DLE-Jena, and Minerva, and additionally on query $\mathcal{Q}'_6$ for Minerva. As already mentioned, answers to queries on GALEN depend on positive occurrences of existential restrictions in axioms, which most systems cannot handle. We could not run Jena Max on GALEN.

## 5.3 Comparing Incomplete Concrete Reasoners

We also implemented the techniques for comparing reasoners from Section 4.3. To this end, we modified REQUIEM to compute subset-closed rewritings, which are then injectively instantiated to obtain a $(\mathcal{Q},\mathcal{T})$-representative sets of ABoxes **R**.

### 5.3.1 Tests with LUBM

As shown in Table 8, representative sets of ABoxes could be computed in just a few seconds for most LUBM queries. The only exception was $\mathcal{Q}_9$, for which REQUIEM did not terminate after disabling rule subsumption optimisations. The size of the representative sets ranged between 1 and 777 ABoxes. As expected, representative sets contain more ABoxes than the exhaustive test suites for the same query and TBox (see Table 3).

All combinations of system and query for which the tests in Section 5.2 identified the system as incomplete are shown in Table 9. The table shows the proportion of certain answers that each system returned when applied to the LUBM data set, the ABoxes in **R**, and the ABoxes in the test suite **S** used in Section 5.2 to check the system's completeness. As shown in the table, OWLim and Jena Micro exhibited the same behaviour and were 'almost' complete. In contrast, Sesame was the least complete for all queries. Furthermore, please note the difference between the values obtained for **R** and those for **S**; in particular, Sesame did not compute any certain answer for $\mathcal{Q}_5$ on **S**, whereas the system is able to compute certain answers for $\mathcal{Q}_5$ on some ABoxes (e.g., on the LUBM data set). This is because the ABoxes in **S** cannot distinguish Sesame from a trivial reasoner that always returns the empty set of answers; however, the set **R** can make such a distinction.

| | **Minerva** | | | | | | | **OWLim & JMicro** | | |
|---|---|---|---|---|---|---|---|---|---|---|
| | $\mathcal{Q}_5$ | $\mathcal{Q}_6$ | $\mathcal{Q}_7$ | $\mathcal{Q}_8$ | $\mathcal{Q}_{10}$ | $\mathcal{Q}_{12}$ | $\mathcal{Q}_{13}$ | $\mathcal{Q}_6$ | $\mathcal{Q}_8$ | $\mathcal{Q}_{10}$ |
| LUBM | 1 | 1 | 1 | 1 | 1 | 1 | 1 | 1 | 1 | 1 |
| **S** | 0.25 | 0.86 | 0.86 | 0.98 | 0.86 | 0.25 | 0.2 | 0.99 | 0.98 | 0.99 |
| **R** | 0.8 | 0.86 | 0.86 | 0.81 | 0.84 | 0.92 | 0.23 | 0.96 | 0.98 | 0.97 |

| | **Sesame** | | | | | | | | | |
|---|---|---|---|---|---|---|---|---|---|---|
| | $\mathcal{Q}_2$ | $\mathcal{Q}_4$ | $\mathcal{Q}_5$ | $\mathcal{Q}_6$ | $\mathcal{Q}_7$ | $\mathcal{Q}_8$ | $\mathcal{Q}_9$ | $\mathcal{Q}_{10}$ | $\mathcal{Q}_{12}$ | $\mathcal{Q}_{13}$ |
| LUBM | 1 | 1 | 1 | 0.87 | 0.83 | 0.64 | 0.83 | 0 | 0 | |
| **S** | 0.75 | 0.68 | 0 | 0.003 | 0.04 | 0.04 | 0 | 0.001 | 0.25 | 0.2 |
| **R** | 0.75 | 0.06 | 0.36 | 0.033 | 0.01 | 0.004 | - | 0.028 | 0.017 | 0.23 |

Table 9: Reasoner comparison for LUBM

| | | $\mathcal{Q}_1'$ | $\mathcal{Q}_2'$ | $\mathcal{Q}_3'$ | $\mathcal{Q}_4'$ |
|---|---|---|---|---|---|
| $\mathcal{C}_s^{\mathcal{Q},\mathcal{T}}$ | Time | 15 | 46 | 70 | 2 |
| | $\sharp\mathbf{R}$ | 140 | 266 | 266 | 127 |

Table 10: Representative sets of ABoxes for GALEN

### 5.3.2 Tests with GALEN

As shown in Table 10, representative sets of ABoxes for GALEN could be computed in times ranging from 2 to 70 seconds, and each set contains only a small number of ABoxes.

For each system and query, Table 11 shows the proportion of certain answers returned by the system on $\mathbf{R}$ and the test suite $\mathbf{S}$ from Section 5.2. Minerva was the most complete system. Jena Micro was better than DLE-Jena (apart from query $\mathcal{Q}_4'$), while DLE-Jena and OWLim behaved in almost the same way (again apart from query $\mathcal{Q}_4'$). As expected, Sesame was the least complete system.

The discrepancies between OWLim, Jena Micro, DLE-Jena and Minerva are rather surprising. OWLim and Jena theoretically support the same features of OWL; furthermore, DLE-Jena is an extension of Jena (Meditskos & Bassiliades, 2008) so DLE-Jena should be at least as complete as Jena, as in the case of LUBM. In order to explain these discrepancies, we analysed the test suites for queries $\mathcal{Q}_1'$–$\mathcal{Q}_4'$. More precisely, we selected ABoxes on which OWLim fails to return all certain answers but on which Jena Micro is complete, and then we identified the minimal set of TBox axioms that entail all certain answers. Our analysis revealed that, for query $\mathcal{Q}_4'$, OWLim fails to find the entailment

$$\mathcal{T} \cup \{\mathsf{Device}(a), \mathsf{HollowTopology}(b), \mathsf{hasTopology}(a,b)\} \models \mathsf{HollowStructure}(a),$$

which follows from the following GALEN axioms:

$$\mathsf{HollowTopology} \equiv \mathsf{Topology} \sqcap \exists\mathsf{hasState}.\mathsf{Hollow}$$

$$\mathsf{Device} \sqsubseteq \mathsf{SolidStructure}$$

$$\mathsf{HollowStructure} \equiv \mathsf{SolidStructure} \sqcap \exists\mathsf{hasTopology}.(\mathsf{Topology} \sqcap \exists\mathsf{hasState}.\mathsf{Hollow})$$

Although existential restrictions appear in several axioms, we can observe that no reasoning over existential variables is actually required, as the first and third axioms imply (by a simple structural transformation) the following axiom:

$$\mathsf{SolidStructure} \sqcap \exists\mathsf{hasTopology}.\mathsf{HollowTopology} \sqsubseteq \mathsf{HollowStructure}$$

| | $\mathcal{Q}'_1$ | | $\mathcal{Q}'_2$ | | $\mathcal{Q}'_3$ | | $\mathcal{Q}'_4$ | |
|---|---|---|---|---|---|---|---|---|
| | **S** | **R** | **S** | **R** | **S** | **R** | **S** | **R** |
| Sesame | 0.01 | 0.18 | ∼0 | 0.16 | ∼0 | 0.16 | 0.04 | 0.10 |
| OWLim | 0.54 | 0.65 | 0.52 | 0.63 | 0.52 | 0.63 | 0.52 | 0.48 |
| DLE-Jena | 0.54 | 0.65 | 0.52 | 0.63 | 0.52 | 0.63 | 0.76 | 0.9 |
| JMicro | 0.69 | 0.82 | 0.68 | 0.81 | 0.68 | 0.81 | 0.76 | 0.67 |
| Minerva | 0.84 | 0.91 | 0.84 | 0.90 | 0.84 | 0.90 | 1 | 1 |

Table 11: Reasoner comparison on GALEN

This axiom entails the required answer, and both systems can deal with axioms of this form; however, unlike Jena Micro, OWLim appears to be incapable of dealing with such cases.

Regarding DLE-Jena, according to its technical description (Meditskos & Bassiliades, 2008), the system has replaced several inference rules of Jena with queries to the DL reasoner, so it does not strictly extend Jena. Our investigation of the exhaustive test suite for query $\mathcal{Q}'_4$ revealed that DLE-Jena returns many answers that are based on existential restrictions on the right hand side of TBox axioms which Jena misses; however, the investigation also revealed that DLE-Jena misses several inferences that Jena's TBox reasoner can capture, which is probably due to the replacement of Jena's inference rules. This also explains why DLE-Jena performs worse than Minerva on GALEN.

These results clearly show that the behaviour of systems greatly depends on the given application scenario. For example, DLE-Jena was complete for all LUBM queries, but it did not perform equally well on GALEN. In contrast, Minerva did not perform well on LUBM, but it was the most complete system for GALEN. Our results thus allow application developers to conduct a thorough comparison of reasoning systems for a given application.

## 6. Conclusion

In this paper we have proposed a theoretical framework and practical techniques for establishing formally provable and algorithmically verifiable completeness guarantees for incomplete ontology reasoners. Our approach radically departs from ad hoc evaluation based on well-known benchmarks, and it provides a solid foundation for striking the balance between scalability and completeness in practical applications.

Our approach also opens up numerous and exciting possibilities for future research. For example, our work opens the door to the design of ontology-based information systems that are optimised for a class of ontologies, queries, and data relevant to a particular application. Such information systems could maximise scalability of reasoning while still ensuring completeness of query answers, even for rich ontologies and sophisticated queries.

## Acknowledgments

This is an extended version of the paper 'How Incomplete is your Semantic Web Reasoner?' by Giorgos Stoilos, Bernardo Cuenca Grau, and Ian Horrocks published at AAAI 2010 and the paper 'Completeness Guarantees for Incomplete Reasoners' by the same authors published at ISWC 2010.

This research has been supported by the EU project SEALS (FP7-ICT-238975), and by the EPSRC projects ExODA (EP/H051511/1) and HermiT (EP/F065841/1). B. Cuenca Grau is supported by a Royal Society University Research Fellowship.